\documentclass{article}

\usepackage{microtype}
\usepackage{graphicx}
\usepackage{subfigure}
\usepackage{booktabs} 

\usepackage{hyperref}

\usepackage[accepted]{icml2026}

\usepackage{amsmath,amssymb,mathtools,amsthm}

\usepackage[capitalize,noabbrev]{cleveref}

\theoremstyle{plain}
\newtheorem{theorem}{Theorem}[section]

\newtheorem{lemma}[theorem]{Lemma}

\theoremstyle{definition}
\newtheorem{definition}[theorem]{Definition}

\theoremstyle{remark}
\newtheorem{remark}[theorem]{Remark}

\usepackage[textsize=tiny]{todonotes}

\usepackage{nicefrac}             
\usepackage{xcolor}
\usepackage[normalem]{ulem}
\usepackage{longtable}            
\usepackage{afterpage}
\usepackage{multirow}
\usepackage{bbm}
\usepackage{tikz}
\usepackage{tabto}
\usepackage{enumitem}

\usepackage{siunitx}       
\usepackage{threeparttable}
\usepackage{array}
\usepackage[utf8]{inputenc}

\usepackage{algorithm}

\usepackage[noend]{algpseudocode} 

\algrenewcommand\algorithmiccomment[1]{\(\triangleright\)~#1}

\newcommand{\algsmall}{\small}

\definecolor{ForestGreen}{RGB}{34,139,34}
\definecolor{MossGreen}{RGB}{134,150,97}

\definecolor{RoyalBlue}{RGB}{0,85,255}
\definecolor{CrimsonRed}{RGB}{217,0,43}

\begin{document}

\twocolumn[
\icmltitle{Position: Many generalization measures for deep learning are fragile 
}



\icmlsetsymbol{equal}{*}

\begin{icmlauthorlist}
\icmlauthor{Shuofeng Zhang}{tp}
\icmlauthor{Ard A. Louis}{tp}

\end{icmlauthorlist}

\icmlaffiliation{tp}{Rudolf Peierls Centre for Theoretical Physics, University of Oxford, United Kingdom}

\icmlcorrespondingauthor{Shuofeng Zhang}{shuofeng.zhang@physics.ox.ac.uk}

\icmlkeywords{Machine Learning, ICML}

\vskip 0.3in
]



\printAffiliationsAndNotice{}  

\begin{abstract}

In this position paper, we argue that many post-mortem generalization measures---those computed on trained networks---are \textbf{fragile}:  small training modifications that barely affect the performance of the 
underlying deep neural network can substantially change a measure's value, trend, or scaling behavior. For example,  minor hyperparameter changes, such as learning rate adjustments or switching between SGD variants, can reverse the slope of a learning curve in widely used generalization measures such as the path norm.  We also identify subtler forms of fragility. For instance, the PAC-Bayes origin measure is regarded as one of the most reliable, and is indeed less sensitive to hyperparameter tweaks than many other measures. However, it completely fails to capture differences in data complexity across learning curves.  This data fragility contrasts with the function-based marginal-likelihood PAC-Bayes bound, which does capture differences in data-complexity, including scaling behavior, in learning curves, but which is not a post-mortem measure. 
Beyond demonstrating that many post-mortem bounds are fragile, this position paper also argues that developers of new measures should explicitly audit them for fragility.
\end{abstract}

\section{Introduction to the position statement}
Classical generalization theories—based on VC dimension, Rademacher complexity  and related tools—have a long history in machine learning.  They were developed with two intertwined aims: to furnish tight mathematical guarantees and to explain generalization. For deep networks the same aspiration has driven a profusion of bounds.  While tight bounds have proven elusive, with some exceptions~\citep{dziugaite2017computing,prezortiz2020tighter,valleprez2020generalization,lotfi2022pacb}, the expectation has been that these measures can generate qualitative insight into the generalization behavior of deep neural networks (DNNs).   These bounds have typically focused on weights after training,  which is why they are sometimes called post-mortem bounds.  
 Large-scale comparisons have compared the behavior of these measures, albeit often only at the sign level, e.g.\ does changing a setting produce the same change in the underlying DNN and a generalization measure with the same sign~\citep{jiang2019fantastic,dziugaite2020in}.   

A notable property of DNNs is that to first order, generalization performance is robust to modest changes in network size, hyperparameters, optimizers, or stopping criteria. 
This raises a natural question: do generalization measures share this robustness? If a generalization measure captures the basic underlying factors that drive DNN performance, one would expect it to exhibit a similar degree of robustness.
The \textbf{main position} this paper argues for is that:  
\textbf{\noindent Many of the post-mortem bounds studied in the literature are \textit{fragile}: small training tweaks that barely affect the underlying DNN performance  can drastically change a measure’s behaviour.  Investigators should carefully audit their generalization
measures for fragility, using the kinds of tests we propose
in this paper.  }

To test these fragilities systematically, we develop a fragility audit that evaluates measures under controlled perturbations that hold data and architecture fixed while nudging a single training knob. Three stressors structure the audit in the main text: learning‑curve behavior as sample size increases (\S\ref{sec:training-hyperparameter-fragility}); temporal behavior, including when the model has interpolated (\S\ref{sec:temporal-fragility}); and responses to data complexity via label noise and dataset swaps (\S\ref{sec:data_complexity_fragility}). \S\ref{sec:compression_fragility} reveals that the tightest existing bound fails across both its bias and complexity components.
In the supplementary material, we test a much wider set of measures and develop an expanded fragility audit.


In contrast to the post-mortem bounds, we show that a function-based marginal-likelihood PAC-Bayes bound successfully tracks dataset complexity and learning-curve scaling~\citep{valleprez2020generalization}. This bound is based on a simple measure of inductive bias, the probability that a random set of weights produces a function that fits the training data set. Because it operates in function space and does not depend explicitly on training dynamics, this bound is inherently insensitive to the choice of optimizer or most training hyperparameters. This insensitivity is both a strength, making the bound robust rather than fragile, and a limitation, as it prevents the bound from reflecting how training procedures influence generalization, including mechanisms such as feature learning. Nonetheless, the strong performance of this relatively simple approach suggests that developing more sophisticated function-space generalization measures could be a fruitful direction for future research.


 Finally, we sketch a few potential causes of fragility, in the hope they can lead to new insights.  We show that some of the qualitative failures of norm-based bounds with sample size appear to track a recent prediction of qualitative changes in how norms scale with data set size in the simpler setting of linear regression~\cite{zhang2025closed}.   
 We also exploit a symmetry in scale‑invariant networks and prove a non‑asymptotic equivalence between fixed learning‑rate with fixed weight decay and a matched run with exponentially increasing learning rate and time‑varying weight decay; the two training procedures compute the same predictor at every iterate (\S\ref{sec:exp++lr}; cf.\ \citealp{li2019an}). Under this invariance lever, magnitude‑sensitive post‑mortem measures can inflate by orders of magnitude while test error remains flat.  This suggests that generalization measures should carefully track key invariances present in the underlying DNNs. 

We issue a call to action: Further work to understand the causes of fragility is essential. Otherwise it is hard to assess whether a generalization measures reflects a true causal relationship or not.


  

\textit{Our contribution:} We (1) formulate a stress-test protocol for hyperparameter stability, temporal drift, and data complexity; (2) demonstrate that popular post-mortem measures are fragile, often reversing trends or failing to track data difficulty  under benign tweaks that leave the underlying predictor unchanged.; (3) introduce a quantitative score for measure instability; and (4) prove a non-asymptotic equivalence between fixed and exponential LR schedules for scale-invariant nets, exposing flaws in magnitude-sensitive measures.

\section{Related work}
\label{sec:related}

We use \emph{generalization bound} and \emph{generalization measure} interchangeably: both seek to predict out-of-sample performance, differing mainly in whether they arrive with formal guarantees. In modern deep networks these quantities are best read as \emph{measures}—diagnostics to compare across training/data regimes—rather than practically tight certificates.

\textbf{From classical capacity to modern practice.}
Early theory framed capacity via VC and Rademacher analyses \citep{vapnik1998statistical,anthony1999neural,bartlett2001rademacher,koltchinskii2001rademacher}. As overparameterization became the norm, evidence accumulated that uniform-convergence explanations often miss the mark: networks can fit random labels yet generalize on natural data \citep{zhang2016understandingc}, some bounds fail to track risk \citep{nagarajan2019uniform}, and benign overfitting can arise even at interpolation \citep{bartlett2020benign}. The field responded with diagnostics that foreground invariances, algorithmic dependence, and data interactions.

\textbf{Capacity-oriented diagnostics: norms, margins, distance from initialization.}
Early work  translated classical notions into deep settings. Spectrally-normalized margin bounds tied test error to Lipschitz-like control and margins, making scale explicit \citep{bartlett2017spectrallyb}. Path- and norm-based views clarified how depth and weight scales shape effective complexity \citep{neyshabur2015norm,neyshabur2019theb}. Empirically, modeling the \emph{distribution} of margins, not just the minimum, improves predictiveness across trained families \citep{jiang2018predicting,novak2018sensitivity}, and sample complexity can depend on norms rather than width \citep{golowich2018size}. In practice, enforcing Lipschitz continuity often improves out-of-sample performance \citep{gouk2020regularisation,yoshida2017spectral}. Distance-from-initialization and movement-from-pretraining offer reference-aware, width-robust complexity surrogates with both empirical support and bounds \citep{li2018on,zhou2021understanding,neyshabur2017exploring}, while optimization dynamics link large margins to implicit bias in separable regimes \citep{soudry2018the}.

\textbf{Geometry-oriented diagnostics: flatness and sharpness.}
A parallel line approached generalization through loss-landscape geometry. The “flat minima generalize better” intuition predates deep learning \citep{hochreiter1997flatb} and resurfaced when large-batch training was observed to converge to sharper minima with worse test error \citep{keskar2016on}. Because raw sharpness is parameterization-dependent, normalized and magnitude-aware definitions were proposed and shown to correlate more robustly with test error \citep{dinh2017sharpb,tsuzuku2020normalized,kim2022scale,jang2022reparametrization}. These ideas also shaped algorithms: SAM explicitly optimizes a local worst-case neighborhood and typically improves accuracy; adaptive variants probe when and why it helps \citep{foret2020sharpnessb,kwon2021asam,zela2022towards}. Independent probes—weight averaging, landscape visualizations, and training-dynamics analyses—add evidence that broader valleys often accompany better generalization \citep{izmailov2018averagingb,li2017visualizing,cohen2021gradient}. Stochastic optimization theory offers a mechanism: noise scale and heavy-tailed gradient noise can bias learning toward flatter regions \citep{smith2017donb,jastrzbski2017three,simsekli2019tail,mccandlish2018an}.

\textbf{Algorithm-aware certificates: PAC-Bayes as operational measures.}
Partly in response to limits of uniform convergence, PAC-Bayesian analysis made algorithm dependence explicit by trading empirical risk against a posterior–prior divergence \citep{mcallester1999someb,seeger2002pac,catoni2007pac,langford2002pac}. The framework became operational when non-vacuous deep-network certificates were obtained by optimizing stochastic posteriors \citep{dziugaite2017computing}, or simply working with the GP prior~\citep{valleprez2018deepb,zhang2021why}. Subsequent work tightened certificates with data-aware priors \citep{dziugaite2018data,prezortiz2020tighter}, and with compression-flavored posteriors that reduce effective description length \citep{arora2018strongerc,lotfi2022pacb}. Treating PAC-Bayes itself as a training objective yields models that are both accurate and tightly certified, while extensions broaden the scope to adversarial risk and fast/mixed-rate regimes \citep{rodrguezglvez2024mixed}. However, one worry is that while these optimised measures may provide quantitatively tighter bounds, they do so at the expense of qualitative insight~\citep{arora2018strongerc}. 

\textbf{Other perspectives.}
Complementary lenses provide boundary conditions any credible measure should respect. Algorithmic stability ties generalization to insensitivity under data perturbations \citep{bousquet2002stability,hardt2016train}; information-theoretic analyses bound excess risk by the information a learning rule extracts from the sample \citep{russo2016controlling,xu2017information}. Description-length and compression viewpoints link compressibility to generalization \citep{blier2018the,suzuki2020compression}. Linearized regimes (NTK/GP) delineate when very wide nets behave like their kernel limits \citep{jacot2018neuralb,lee2019wide}. Finally, double-descent phenomena and scaling laws offer external checks on diagnostics \citep{belkin2019reconciling,kaplan2020scaling}.

\textbf{Meta-evaluations and synthesis.}
Large-scale comparisons underscore that no single candidate explains generalization under all interventions~\citep{jiang2019fantastic,dziugaite2020in}, motivating a measures-as-diagnostics mindset. A recent critique shifts the lens from \emph{predictiveness} to \emph{tightness}, showing that uniformly tight bounds are out of reach in overparameterized settings and clarifying what such tightness could mean \citep{gastpar2023fantastic}. Our focus is complementary: rather than tightness per se, we study \emph{fragility}—how otherwise informative measures can fail under innocuous hyperparameter changes that leave underlying DNN performance essentially unchanged.

\section{Training-hyperparameter fragility}
\label{sec:training-hyperparameter-fragility}

The path norm (definition in App.~\ref{app:all-measures}) is often regarded as the strongest norm-based proxy for modern ReLU networks: it respects layerwise rescalings and has supported some of the sharpest norm-style capacity statements and practical diagnostics \citep{neyshabur2015norm,neyshabur2017a,gonon2023a}. Precisely because it is a strong candidate, it is a natural place to look for fragility.

\begin{figure*}[t]
\centering
\includegraphics[width=.98\linewidth]{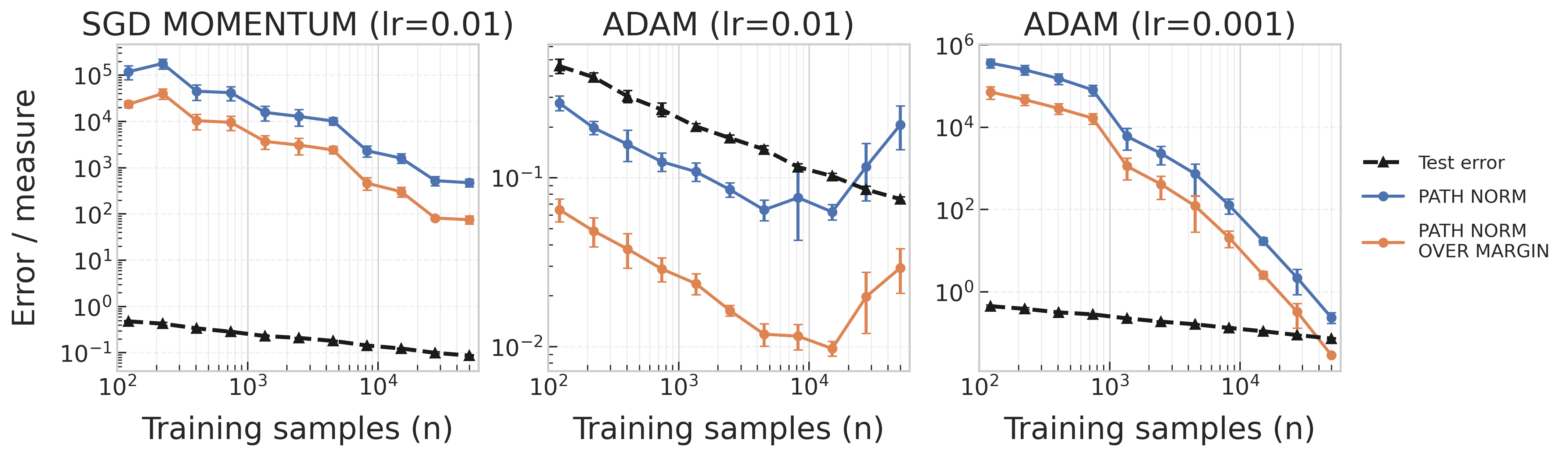}
\caption{\textbf{Path‑norm learning curves under small training‑pipeline changes (no early stopping).} \emph{Left:} \textsc{SGD} with momentum, LR $=0.01$. \emph{Middle:} \textsc{Adam}, LR $=0.01$. \emph{Right:} \textsc{Adam}, LR $=0.001$. Curves show \textsc{ResNet‑50} test error (black), path norm (blue), and path norm over margin (orange); axes are logarithmic and error bars denote seed variability. Training set sizes $n$ range from $10^2$ to the full FashionMNIST training set ($5\times 10^4$ samples).}
\label{fig:lcfrag}
\end{figure*}

In Fig.~\ref{fig:lcfrag} we show a triptych for \textsc{ResNet‑50} on \emph{FashionMNIST}, nudging exactly one hyperparameter at a time, either the optimizer or the learning rate (no early stopping), while keeping the data, model, and pipeline fixed.   Each curve shows test error (black), the raw path norm (blue), and the path norm divided by the empirical margin (orange). In the \textbf{left} plot, we run \textsc{SGD} with momentum at $\mathrm{LR}=0.01$: both path‑norm curves start at a value of roughly ($\sim 10^{5}$) and decrease more or less log‑linearly with increasing sample size, with the margin‑normalized curve tracking about a decade lower. Both drop faster than the underlying \textsc{ResNet‑50} test error.  Changing the optimizer to \textsc{Adam} but keeping $\mathrm{LR}=0.01$ (\textbf{middle} panel)  results in much smaller path norms (now near $10^{-1}$) and striking non-monotonic behavior. A minor adjustment of  \textsc{Adam}’s learning rate to $0.001$ (\textbf{right} panel)—and the metrics jump back up to $\sim 10^{5}\text{--}10^{6}$ but now collapse by four orders of magnitude as data size grows. These widely different behaviors arise purely from modest hyperparameter tweaks that barely affect the underlying \textsc{ResNet‑50} test error.

The contrasting behavior of the path-norm measure appears to mirror the simpler, analytically tractable case for linear regression. In that setting, for a fixed $\ell_r$ weight norm,  qualitatively different scaling of the norm with sample size obtains depending on the $\ell_p$‑minimizer. More precisely,  with $p \in (1,2]$, the $\ell_r$ weight norm is monotonic in $n$  if $r \leq 2(p-1)$ and plateaus if $r > 2(p-1)$~\citep{zhang2025closed}. 
This implies that fixing \emph{which} norm you measure does not fix its learning‑curve scaling, because the solution selected by optimization—through implicit (or explicit) bias, e.g., favoring different $\ell_p$‑minimizers—can induce different sample‑size scalings of that same norm. This change in qualitative behavior was also illustrated in diagonal linear networks where the implicit bias can be tuned by varying the learning dynamics~\cite{zhang2025closed}.   For the more complex setup of path-norm for \textsc{ResNet‑50} on FashionMNIST in Fig.~\ref{fig:lcfrag},  the optimizer and learning‑rate choices may analogously shift the deep model’s implicit bias, toggling the path‑norm scaling between monotone and U‑shaped, just as was seen for linear regression and deep linear networks.  

See App.~\ref{app:all-measures} for precise definitions for the measures we consider as well as further experiments, including the same stress test across Frobenius, margin, spectral, PAC‑Bayes, and VC‑style proxies\footnote{All experiments in this work are fully reproducible with minimum hardware requirements.
The anonymized code can be found at \url{https://anonymous.4open.science/r/Prior_bound-8A6C/}.}.

\section{Temporal behavior fragility}
\label{sec:temporal-fragility}

Temporal traces ask a simple question that post‑mortem snapshots cannot: once the classifier stops changing, does the measure stop moving? Let \(T_{\mathrm{int}}\) denote the first epoch at which training accuracy reaches \(100\%\). A stable diagnostic should largely settle for \(t>T_{\mathrm{int}}\); continued change indicates optimizer‑driven drift rather than functional change.

\begin{figure*}[t]
    \centering
    \includegraphics[width=.98\linewidth]{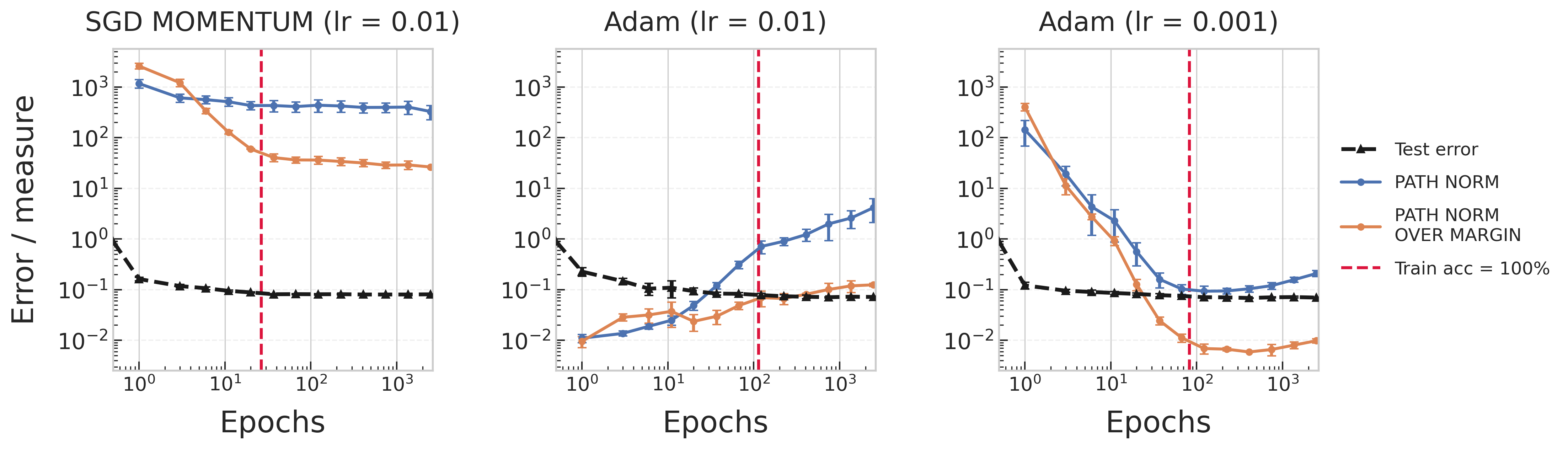}
    \caption{ResNet‑50 on FashionMNIST as a function of training epochs.   \emph{Left} (\textsc{SGD}+momentum, LR \(=0.01\)). 
      \emph{Middle} (\textsc{Adam}, LR \(=0.01\)). 
     \emph{Right} (\textsc{Adam}, LR \(=0.001\)). 
    While the path norm and path norm over margin measures show  qualitative different behavior upon small hyperparameter tweaks, the underlying generalization  is similar across the three panels.}
    \label{fig:temporal-fragility-norms}
\end{figure*}

A  set of three experiments on FashionMNIST illustrates the point. We fix the architecture and data, then vary one knob at a time across the three panels of Figure~\ref{fig:temporal-fragility-norms}. Holding the learning rate fixed at \(0.01\) and swapping the optimizer from \textsc{SGD} with momentum (left) to \textsc{Adam} (middle) flips the post‑interpolation regime: with \textsc{SGD}, the path norm sits near \(5\times10^{2}\) and then slowly declines and the ratio‑over‑margin trends downward (\(T_{\mathrm{int}}\!\approx\!27\)), whereas with \textsc{Adam} the path norm climbs from \(\approx 10^{-2}\) to \(10^{0}\) and the ratio rises in step, continuing past interpolation (\(T_{\mathrm{int}}\!\approx\!114\)). Holding the optimizer fixed (\textsc{Adam}) and reducing the learning rate from \(0.01\) to \(0.001\) (right) again changes the temporal behavior: both traces drop rapidly and then increase moderately after the dashed line. In all cases, the generalization curve remains on a comparable scale. 

A plausible mechanism underlies this split. After interpolation under cross‑entropy, gradients can continue to increase logit scale along near‑flat directions. \textsc{Adam} tends to amplify this scale drift, while \textsc{SGD}+momentum tempers it, yielding opposite signs for the post‑interpolation slope. The lesson is that weight-norm-based measures---and margin-normalized variants that co-move with them---are not monotone indicators of optimization progress: an upward trend can be optimizer‑driven scale inflation; a flat or gently declining trend need not signal stagnation.
We refer the readers to Appendix~\ref{app:temporal-behavior} for experimental results on the temporal behaviors of more generalization measures.

\section{Data-Complexity Fragility: Label Noise \& Dataset Difficulty}
\label{sec:data_complexity_fragility}

How DNN performance varies with data complexity or sample size (learning curves) offers key insights into the mechanisms of generalization. Therefore, reporting how a generalization measure behaves under changes in data complexity or sample size  should be a standard part of its evaluation.

\begin{figure*}[t]
\centering
\includegraphics[width=.98\linewidth]{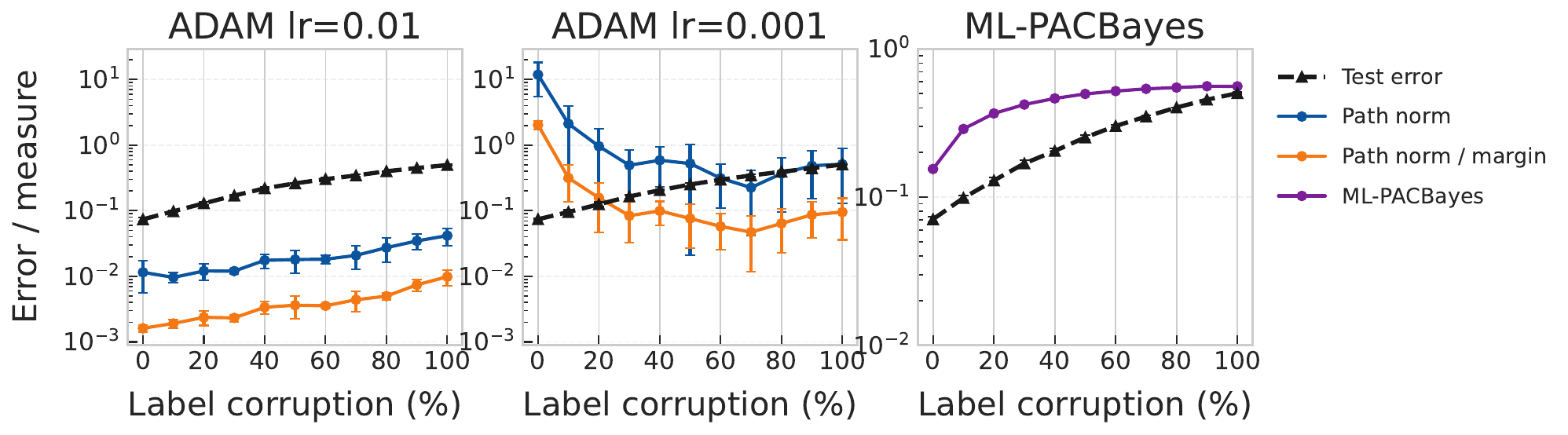}
\caption{\textbf{Label‑corruption sweeps with post‑mortem norms vs.\ function‑space ML‑PACBayes on RESNET-50.} 
\emph{Left:} \textsc{Adam}, $\mathrm{LR}{=}0.01$. \emph{Middle:} \textsc{Adam}, $\mathrm{LR}{=}0.001$. \emph{Right:} ML‑PACBayes; purple,  
Dashed black: test error; blue: path norm; orange: path norm over margin. 
A ten‑fold LR change reverses the path‑norm trend (left vs.\ middle), while ML‑PACBayes remains a smooth, monotone function of corruption (right) and is, by construction, agnostic to optimizer/LR. 
\emph{All panels use 10{,}000 training samples on binarized FashionMNIST.}}
\label{fig:pathfrag-adam}
\end{figure*}

The first two panels in Figure~\ref{fig:pathfrag-adam} 
illustrate how merely changing \textsc{Adam}’s learning rate flips the qualitative trend of the path‑norm curves (left vs.\ middle)  with data complexity (for  \textsc{SGD} with momentum on this data, see Appendix~\ref{app:label-corruption}). This illustrates the kind of fragility our framework urges authors to investigate and report when treating bounds as \emph{measures}. 

Finally, the last panel in figure~\ref{fig:pathfrag-adam}   illustrates how the function‑space ML‑PACBayes measure provides a non-vacuous bound that tracks data-complexity (see also 
Appendix~\ref{sec:post-mortem_vs_mar_lik}).
This performance is not mirrored by  several PAC–Bayes \emph{parameter‑space} surrogates, which on the one hand are much less sensitive to training hyperparameters than the path norm measures are, but on the other hand exhibit a different more subtle kind of fragility; they are typically \emph{insensitive} to label corruption in standard setups (see Appendix~\ref{app:label-corruption}). 
We call this behavior a fragility because it reflects a serious failing of these measures.

\begin{figure*}[t]
\centering
\includegraphics[width=.98\linewidth]{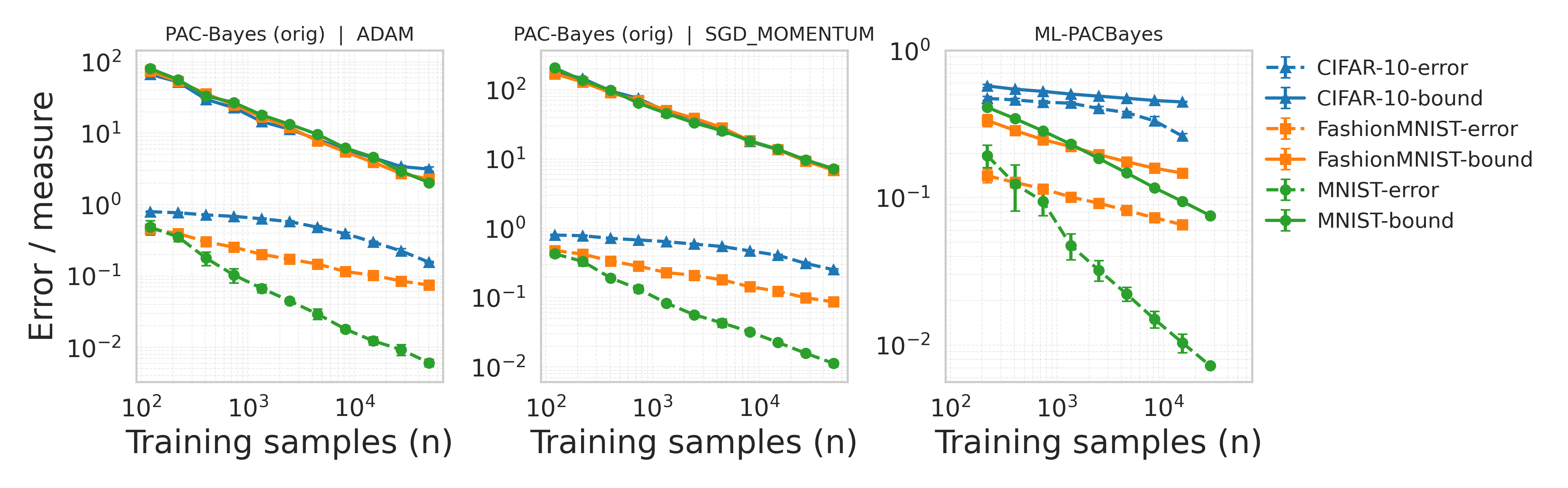}
\caption{\textbf{Fixed architecture, varying dataset.}. The model is RESNET-50.
\emph{Left:} \texttt{PACBAYES\_ORIG} with \textsc{Adam}. 
\emph{Middle:} \texttt{PACBAYES\_ORIG} with \textsc{SGD} (momentum). Changing from ADAM to SGD with momentum leads to a small increase in the overall scale of the \texttt{PACBAYES\_ORIG} learning curves (solid lines). However, for both optimizers, these learning curves don't distinguish  \textsc{MNIST}, \textsc{FashionMNIST}, and \textsc{CIFAR10},  missing \textit{both} the data‑complexity ordering and the changing slopes visible in the underlying learning curves (dashed lines). 
\emph{Right:} ML‑PACBayes (marginal‑likelihood PAC–Bayes) is tight and preserves the correct dataset ordering and reproduces the trends in the slope; for technical reasons, this panel is computed on the \emph{binarized} versions of the three datasets.}
\label{fig:pacorig-datasets}
\end{figure*}

To illustrate data-complexity fragility, we  chose the \texttt{PACBAYES\_ORIG} measure (definition in App.~\ref{app:all-measures}) because it is thought to be one of the best in its class.  We hold the \emph{architecture and hyperparameters fixed} and use the datasets \textsc{MNIST}, \textsc{FashionMNIST}, and \textsc{CIFAR10}.  As illustrated in \autoref{fig:pacorig-datasets}, 
 the dashed generalization–error curves separate cleanly and their \emph{slopes} decrease with dataset difficulty (MNIST steepest decline, CIFAR‑10 flattest).   However, the first two panels of \autoref{fig:pacorig-datasets} show that for a given set of hyperparameters, the  \texttt{PACBAYES\_ORIG} measure gives nearly identical numerical values for different datasets and therefore \emph{fails to recognize changes in dataset complexity}: its learning‑curve slopes are nearly indistinguishable across datasets and reflect changes in the test‑error slopes. 
By contrast, the third panel (ML‑PACBayes) is both \emph{tight} (close to the corresponding error curves) and \emph{data‑aware}: it preserves the MNIST\,$<$\,FashionMNIST\,$<$\,CIFAR‑10 ordering and reflects the different slopes. 
See Appendix~\ref{sec:post-mortem_vs_mar_lik} for the bound’s definition and a fuller discussion of why this function‑space quantity captures dataset difficulty while post-mortem parameter‑space  PAC-Bayes surrogates often do not.

There has been substantial recent interest in the study of scaling laws for deep neural networks (DNNs). 
These laws describe how generalization performance varies as a function of model size, training sample size, or computational cost 
(see, e.g.,~\citep{hestness2017deep, kaplan2020scaling, spigler2019asymptotic, hoffmann2022training, valleprez2020generalization, bahri2024explaining, nam2024visualising, goring2025feature} and references therein).   Here, we observe that the ML-PACBayes bound does quite well at capturing the scaling behavior with the training sample size.  However, it should be kept in mind that because this bound is evaluated with a Gaussian process (GP) method, it cannot capture feature-learning, which can play a crucial role in learning-curve scaling for certain datasets~\citep{nam2024visualising,goring2025feature}. 

In App.~\ref{sec:data-complexity}, we move 
beyond label corruption and dataset swaps, and also probe symmetry‑preserving vs.\ signal‑destroying data transforms via pixel permutations. Again, we find that the ML-PACBayes function‑space predictor tracks the expected invariances while several post‑mortem diagnostics do not. 

\section{Fragility of tightness: the case of compression bounds}
\label{sec:compression_fragility}

Recent work by \citet{lotfi2022pacb} proposes PAC-Bayes compression bounds that achieve state-of-the-art tightness. They argue that this tightness allows them to ``explain'' generalization.  While this approach is theoretically stimulating, our audit suggests that numerical tightness does not necessarily imply a correct causal explanation of the generalization phenomenon. We identify two specific fragilities.

\begin{figure*}[t]
\centering
\subfigure[CIFAR-10 (ID 3500) Bounds]{
    \includegraphics[width=0.31\linewidth]{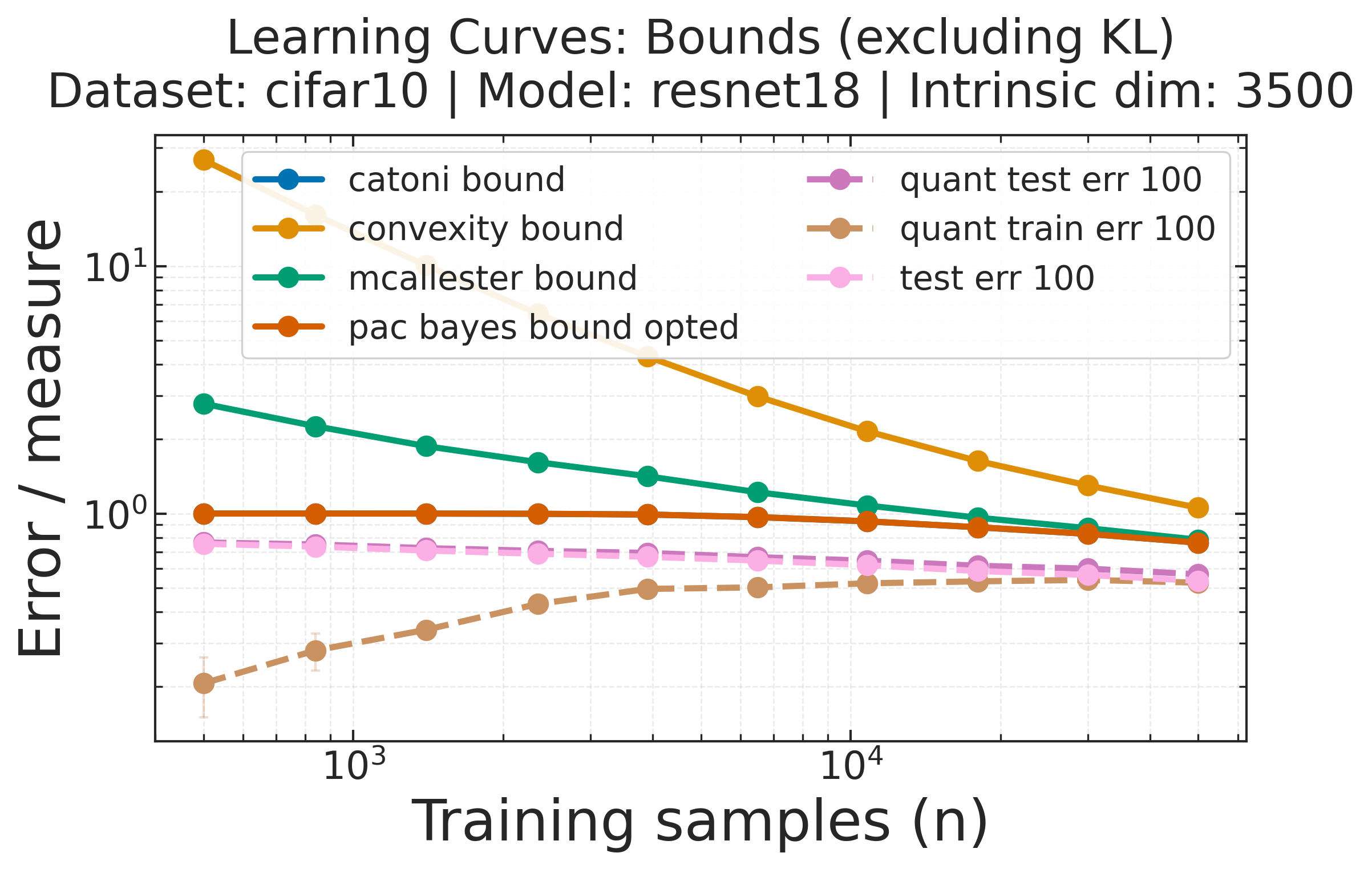}
    \label{fig:comp_c10_3500_met}
}
\hfill
\subfigure[MNIST (ID 1000) Bounds]{
    \includegraphics[width=0.31\linewidth]{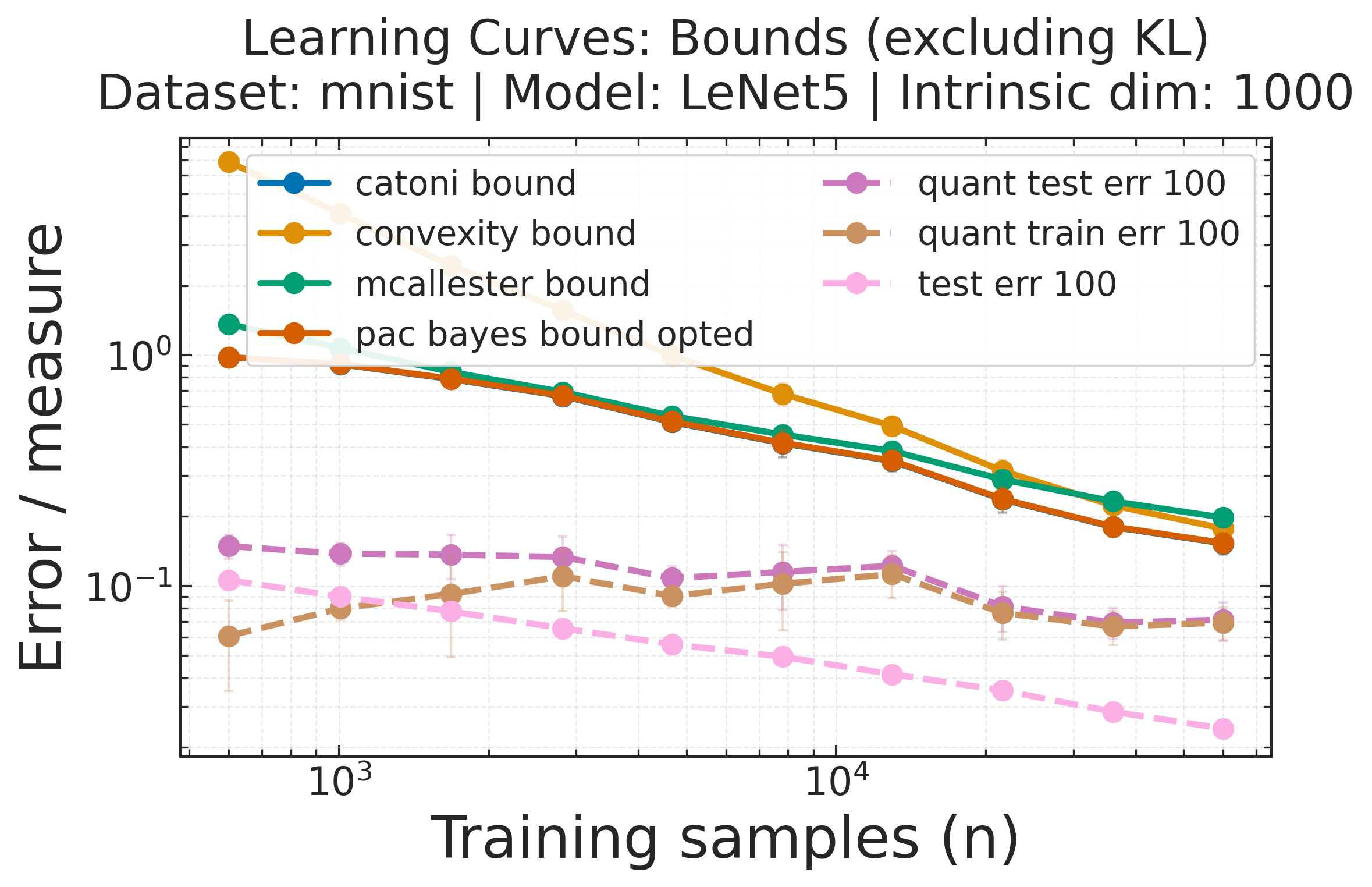}
    \label{fig:comp_mnist_met}
}
\hfill
\subfigure[KL Divergence (CIFAR vs MNIST)]{
    \includegraphics[width=0.31\linewidth]{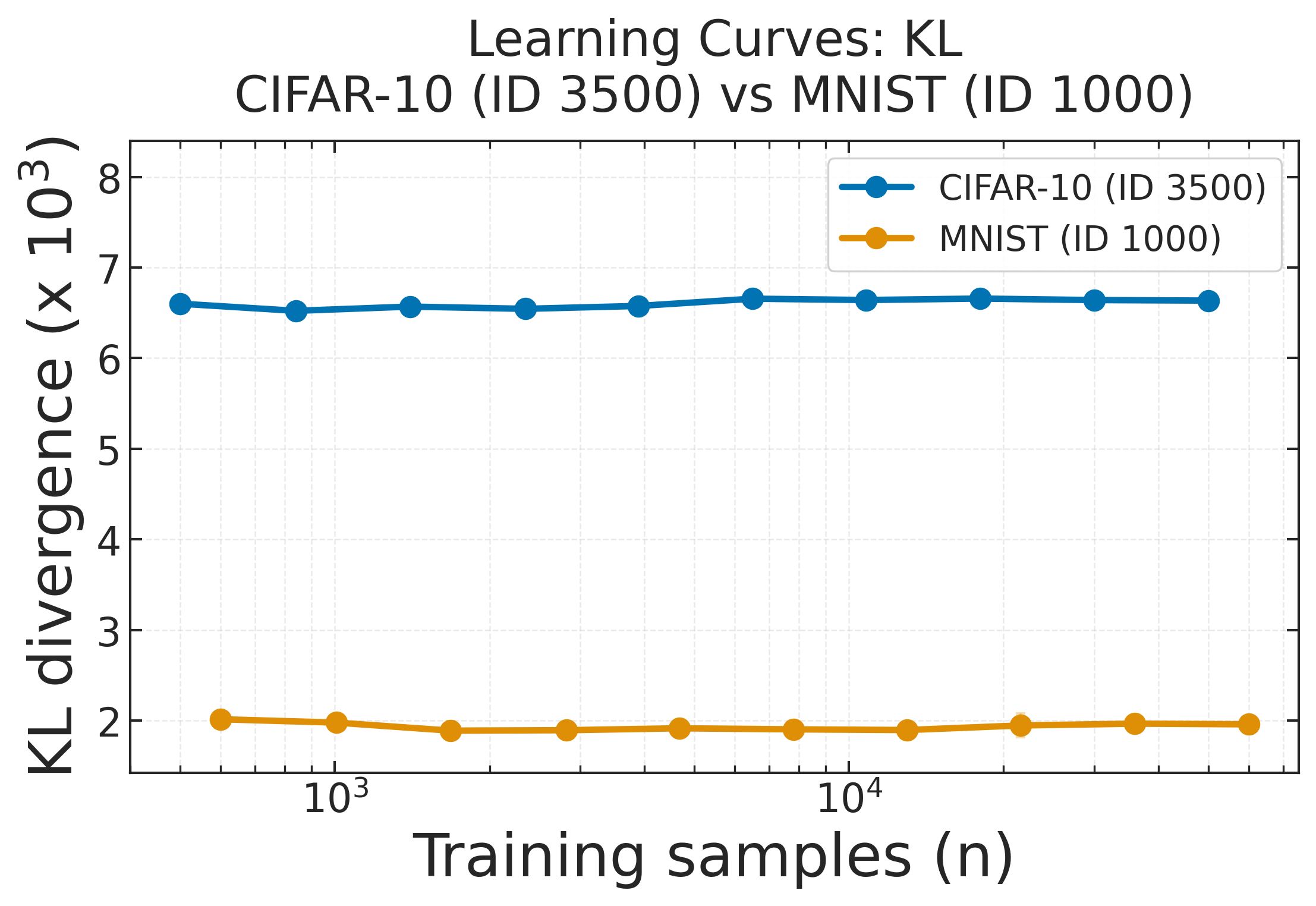}
    \label{fig:comp_merged_kl}
}
\caption{\textbf{Decomposing PAC-Bayes Compression Bounds.} \emph{Left \& Middle:} Learning curves for bounds and error rates. The ``quantized training error'' (dashed brown) is large and tracks the test error. \emph{Right:} The Raw KL divergence (complexity term) remains essentially flat for both datasets across sample sizes ($n$). Additional results for higher intrinsic dimensions are provided in Appendix~\ref{app:additional_compression}.}
\label{fig:compression_fragility}
\end{figure*}

\textbf{Bias dominance over complexity.} 
In standard learning theory, bounds typically trade off a small training error against a complexity penalty. However, as shown in the \textbf{left and middle panels} of Figure~\ref{fig:compression_fragility}, the ``quantized training error'' (dashed brown line)—which serves as the empirical risk term in the bound—is notably large. Crucially, as the sample size $n$ increases, this quantized training error does not vanish but instead grows to track the quantized test error. This suggests that the final bound's tightness is driven primarily by this large bias term (the quantization error) rather than by the  KL-divergence complexity term measuring the neural network's capacity. In practice, modern DNNs often achieve near-zero training error; a bound that relies on a large proxy training error to achieve tightness is arguably measuring the quality of the compression scheme rather than the generalization mechanism of the original network.

\textbf{Insensitivity of the complexity term.} 
A generalization measure should ideally reflect how the difficulty of the learning problem changes with data availability. However, the \textbf{right panel} of Figure~\ref{fig:compression_fragility} demonstrates that the KL divergence—the core measure of complexity in the PAC-Bayes formulation—remains almost perfectly flat with respect to training set size $n$. This flatness persists across different datasets (CIFAR-10 vs.\ MNIST) and intrinsic dimensions (see Appendix~\ref{app:additional_compression} for results with $\text{ID}=10000$). This implies that the complexity term fails to capture the scaling of learning difficulty, rendering the bound insensitive to the actual information gain from increasing data.

\section{Scale-invariant network and exponential learning rate schedule}
\label{sec:exp++lr}

Normalization layers render many modern networks effectively scale-invariant, motivating the exponentially increasing learning-rate scheme of \citet{li2019an}. In \textbf{Appendix~\ref{sec:proof_of_theorem_exp+wd}}, we prove a non-asymptotic equivalence: training with fixed learning rate (LR) and weight decay (WD) is exactly matched in function space by an exponentially increasing LR with time-varying WD. 

We refer to this multiplicative control parameter $\alpha$ as the \emph{Exp++ factor}. This theoretical result provides a precise ``invariance lever'': it allows us to arbitrarily inflate parameter norms while leaving the learned function $f(\theta)$—and thus the test error—mathematically identical at every step. This mechanism serves as a crucial control for our audit: if a generalization measure is sensitive to parameter magnitude (like path norm or standard PAC-Bayes), it will explode under this transformation, even though generalization performance remains identical.

The formal definitions, the statement of Theorem~\ref{theorem:exp+wd}, and its proof are provided in \textbf{Appendix~\ref{sec:proof_of_theorem_exp+wd}}. Empirical results utilizing this lever are detailed in Appendix~\ref{app:exp-lr}.

\section{Quantitative fragility: a scale‑free instability score}
\label{sec:quantitative_fragility}

The heuristic stress tests in Sections~\ref{sec:training-hyperparameter-fragility}--\ref{sec:data_complexity_fragility} show that several popular post‑mortem measures are fragile. To facilitate large-scale comparisons, we developed a complementary \emph{quantitative framework} that assigns a specific instability score to a measure. We introduce two statistics: \textbf{Conditional Measure Spread (CMS)}, which captures the fluctuation of a measure across runs with nearly identical test error, and \textbf{Excess CMS (eCMS)}, which isolates the component of that fluctuation driven specifically by hyperparameter changes (net of random seed variance).

\textbf{Summary of results.} As detailed in Appendix~\ref{app:quant_fragility}, we computed these scores for a wide range of measures. We found that parameter-count proxies are stable (eCMS $\approx 0$), while path norms, spectral distances, and standard PAC-Bayes surrogates exhibit high instability (large eCMS), quantitatively confirming the qualitative fragility observed in our stress tests. See Appendix~\ref{app:quant_fragility} for the formal definitions (Definitions~\ref{def:cms} and \ref{def:ecms}) and the full tabulated results.

\section{Alternative views}
\label{sec:altviews}

Reasonable researchers can disagree with our emphasis on parameter-space fragility. 
We highlight several viable counterpositions and address them in turn:


\textbf{Optimizer sensitivity as a feature.}
It is widely thought that optimizer, schedule, and batch size are first‑order determinants of generalization that \emph{should} be reflected in generalization measures. This view points to well‑documented implicit‑bias phenomena—e.g., margin growth under separable losses and optimizer‑dependent convergence paths—that influence which classifier is ultimately selected \citep{soudry2018the,smith2017donb,simsekli2019tail}. From this perspective, the very invariances we prize risk washing out real, practically actionable differences between training recipes. 

\emph{Response.} We agree that optimizer choice can change the learned function and that a diagnostic should detect \emph{functionally meaningful} changes. Our claim is narrower: diagnostics that swing under \emph{pure scale drift} or minor hyperparameter nudges while test error and predictions remain essentially fixed can mislead day‑to‑day comparisons. This is why our audit always pairs any optimizer comparison with either (i) matched‑prediction checkpoints (e.g., at the first $100\%$‑accuracy crossing) or (ii) a symmetry lever that alters parameter scale without changing the predictor (\S\ref{sec:exp++lr}). When pipeline changes \emph{do} alter the predictor, a stable measure should move in tandem with test error; when they do not, a robust measure should be indifferent.

\textbf{Stabilizing post‑mortem diagnostics.}
A second objection holds that many post‑training measures already address scale and parameterization issues. Normalized or reparameterization‑aware sharpness correlates more reliably with generalization than raw curvature \citep{dinh2017sharpb,tsuzuku2020normalized,kim2022scale,jang2022reparametrization}; modeling the \emph{distribution} of margins rather than the minimum improves predictiveness across runs \citep{jiang2018predicting}; reference‑aware surrogates such as distance from initialization or movement from pretraining reduce width and scale artifacts and come with supporting analyses \citep{li2018on,neyshabur2017exploring,zhou2021understanding}. On this view, fragility largely reflects naive implementations, not intrinsic flaws.

\emph{Response.} We see these developments as complementary and encouraging. Our results target precisely such “best‑effort” variants (normalized, margin‑aware, reference‑aware) and still uncover qualitative flips under mild pipeline changes when predictions are stable. We do not argue that post‑mortems are without merit;  rather, we argue they remain \emph{fragile enough} that authors should routinely report their stability under the kinds of stressors we outline (\S\ref{sec:training-hyperparameter-fragility}–\ref{sec:data_complexity_fragility}), and prefer versions that (i) are explicitly invariant to layerwise rescalings, (ii) condition on matched prediction milestones, and (iii) are benchmarked against data‑difficulty shifts and temporal drift after interpolation. In short, our audit is a bar to clear, not a dismissal of the enterprise.

\textbf{Limits of function–space GP references.}
A third objection is that using a  marginal‑likelihood PAC–Bayes predictor as a baseline (Appendix~\ref{sec:post-mortem_vs_mar_lik}) overstates its practical authority.  Since a GP is used to approximate the key marginal-likelihood term in the bound, the ML-PACBayes predictor could be seen as a measure of the performance of the underlying GP for any given architecture-data combination.  While GPs are known to closely match DNN performance on the datasets we test in this paper~\citep{lee2019wide},  it is hard to see how one could untangle the effects of hyperparameters or optimizer choice in this approach, in other words, the kinds of questions that post-mortem bounds are trying to address. More generally, 
modern deep-learning models are finite‑width, heavily augmented, and often trained far from the GP regime; priors that ignore augmentation, architectural quirks, or fine‑tuned tokenization may be badly mis‑specified. In addition, the bound controls the error of a stochastic Gibbs classifier drawn from a posterior over functions, not necessarily the deterministic network one actually deploys. Therefore, the GP route’s stability could stem from \emph{omitting} factors that matter in practice \citep{lee2019wide,valleprez2020generalization}.

\emph{Response.} We agree that one of the most important questions that needs explaining in DNNs is how, when, and why finite‑width networks outperform their infinite-width GP limits, in particular for settings where feature-learning is important (see e.g.~\cite{goring2025feature} and references therein).  We use the GP based marginal‑likelihood bound as a \emph{calibration} tool, not an oracle: it naturally encodes desirable invariances (insensitivity to parameter scale and optimizer path once the dataset is fixed) and for the cases we study, consistently tracks data difficulty and learning‑curve scaling across families. Those properties make it a valuable foil for stress‑testing post‑mortem bounds. Where the GP prior is clearly misspecified (e.g., heavy augmentation or domains where feature-learning is key), our recommendation is empirical: rerun the same fragility audit. If the GP predictor ceases to track error while a reparameterization‑invariant post‑mortem does, that is evidence \emph{for} the importance of a post‑mortem analysis in that regime.


\section{Call to action: a better standard for evaluation}
\label{sec:call_to_action}

Our position is not merely that current measures are fragile, but that the community must adopt new standards to diagnose and mitigate this fragility. We propose three concrete actions for authors, reviewers, and benchmark designers.

\textbf{1. Mandate the Fragility Audit.}
Just as error bars are expected for experimental results, a ``stability check'' should accompany any proposed generalization measure. Authors introducing a new bound or proxy should report its behavior under the \textbf{Minimum Fragility Stress-Test}:
\begin{itemize}[leftmargin=*, noitemsep, topsep=0pt]
    \item \textbf{The Hyperparameter Sweep:} Plot the measure against learning rate and optimizer changes. If the measure shifts by orders of magnitude while test error remains constant (as seen in Fig.~\ref{fig:lcfrag}), this sensitivity must be reported.
    \item \textbf{The Data-Complexity Check:}
    Demonstrate that the measure correctly ranks datasets by difficulty (e.g., MNIST vs. CIFAR-10) or tracks label noise (Fig. 3).
    \item \textbf{The Temporal Check:} Report the measure's post-interpolation slope. A valid generalization measure should not drift indefinitely after the training error has hit zero.
\end{itemize}

\textbf{2. Prioritize Invariance in Design.}
The community should pivot from deriving tighter worst-case bounds to designing \emph{invariant} measures. We recommend developing post-mortem diagnostics that are explicitly invariant to layerwise reparameterizations and scale transformations (such as the Exp++ mechanism in \S\ref{sec:exp++lr}). Measures based on raw parameter magnitudes should be avoided in favor of relative or normalized metrics, unless scale dependence is explicitly justified. 

\textbf{3. Develop Function-Space Surrogates.}
Our results with the ML-PACBayes bound suggest that function-space quantities naturally avoid parameter-space fragility. A major open challenge is to develop computable approximations of function-space complexity—such as approximate marginal likelihood or function-space entropy—that scale to modern deep networks. We call for a research program dedicated to estimating these quantities efficiently, potentially bridging the gap between the robustness of GP limits and the feature-learning capabilities of finite-width networks.

\textbf{Conclusion.} In summary, we have argued that generalization measures in deep learning must be judged not only by their tightness but also by their robustness. Through systematic auditing, we demonstrated that many popular post-mortem bounds—including path norms, spectral measures, and standard PAC-Bayes surrogates—are fragile: they can be made to vary wildly, trend in reverse, or collapse entirely under benign training changes that leave the underlying network's performance effectively unchanged.  More subtle, but important, fragilities include failing to capture basic generalisation trends with data size or data complexity. 


This fragility is likely more pronounced for measures that probe parameter space rather than function space. By contrast, a simple function-based marginal-likelihood PAC-Bayes bound, which captures the inductive bias of the architecture, tracks data complexity and learning-curve scaling, though it cannot distinguish optimizer effects. Its performance provides a natural benchmark for future, more sophisticated measures that also  capture post-training effects.

 We contend that the current practice of evaluating post-training diagnostics---often limited to mild variations in training pipelines---is insufficient to expose true fragility. By adopting the more rigorous audits and invariance principles proposed in this work, the field can move toward measures that genuinely reflect and, more importantly, help us \textit{understand} the mechanisms of generalization.



\bibliography{citations}
\bibliographystyle{icml2026}

\newpage
\appendix
\onecolumn

\renewcommand{\theHsection}{A\arabic{section}}
\renewcommand{\thefigure}{S\arabic{figure}}
\setcounter{figure}{0}

\section{Training-hyperparameter fragility for all measures}
\label{app:all-measures}

In this section we give concise definitions of the measures we study (aligning notation with prior large-scale evaluations) and provide additional experimental evidence beyond the main text~\citep{dziugaite2020in,jiang2019fantastic}. We then present figure-backed comparisons where small learning-rate or optimizer tweaks trigger qualitatively different curve shapes for the surrogate—plateaus, rebounds, late spikes, and crossings—while the accompanying accuracy curves remain comparatively calm.

\subsection{Measures considered}
We adopt the categories and normalizations used by prior experimental studies of generalization measures \citep{jiang2019fantastic,dziugaite2020in}; for exact constants and implementation choices, see App.~C.6 of \citet{dziugaite2020in}. Let a $d$-layer network have layer weights $\{W_i\}_{i=1}^d$ with initialization $\{W_i^0\}$; write $\|\cdot\|_F$ and $\|\cdot\|_2$ for Frobenius and spectral norms; let $n$ denote training-set size and $\gamma$ a robust (e.g., 10th-percentile) training margin.

\begin{itemize}
  \item \textbf{Frobenius distances.} Layerwise distances from initialization and norm aggregates, e.g.
  \[
    C_{\mathrm{FrobDist}}
    \;=\;
    \sqrt{\sum_{i=1}^d \|W_i - W_i^0\|_F^2 \;/\; n}\,,
    \qquad
    C_{\mathrm{param}}
    \;=\;
    \sqrt{\sum_{i=1}^d \|W_i\|_F^2 \;/\; n}\,.
  \]
  \item \textbf{Inverse margin.} A margin-based surrogate,
  \[
    C_{\mathrm{inv\text{-}margin}} \;\propto\; \frac{\sqrt{n}}{\gamma}\,.
  \]
  \item \textbf{Spectral metrics.} Products/means and distances in spectral norm, e.g.
  \[
    C_{\prod \mathrm{spec}}
    \;=\;
    \sqrt{\prod_{i=1}^d \|W_i\|_2^2 \;/\; n}\,,
    \qquad
    C_{\mathrm{DistSpecInit}}
    \;=\;
    \sqrt{\sum_{i=1}^d \|W_i - W_i^0\|_2^2 \;/\; n}\,.
  \]
\item \textbf{Combined spectral–Frobenius ratio.}
  We follow the combined ratio used in prior large‑scale studies (App.~C.6 of \citealp{dziugaite2020in}), denoted \texttt{FRO\_OVER\_SPEC}, which normalizes Frobenius quantities by spectral ones to reduce raw scale effects. We report it alongside its constituents in our plots.

  \item \textbf{PAC-Bayes families and flatness proxies.} Bounds/proxies parameterized by posterior radii $\sigma$ (and magnitude-aware $\sigma_0$), e.g.
  \[
    C_{\mathrm{PACBAYES\text{-}ORIG}}
    \;=\;
    \frac{1}{\sqrt{n}}\sqrt{\frac{\|w\|_2^2}{4\sigma^2}+\log\!\Big(\frac{n}{\delta}\Big)+10},
    \quad
    C_{\mathrm{Flatness}}
    \;=\; \frac{1}{\sigma\sqrt{n}},
    \quad
    C_{\mathrm{MAG\text{-}Flatness}}
    \;=\; \frac{1}{\sigma_0\sqrt{n}}.
  \]
  \item \textbf{Path norms.} With $w=\mathrm{vec}(W_1,\ldots,W_d)$ and $f_{w^2}(\mathbf{1})[i]$ the $i$th logit when all weights are squared elementwise and the input is all ones,
  \[
    C_{\mathrm{path.norm}}
    \;=\;
    \sqrt{\sum_i f_{w^2}(\mathbf{1})[i] \;/\; n},
    \qquad
    C_{\mathrm{path.norm\text{-}over\text{-}margin}}
    \;=\;
    \sqrt{\sum_i f_{w^2}(\mathbf{1})[i] \;/\; (\gamma^2 n)}.
  \]
  \item \textbf{VC-dimension proxy (parameter count).} A coarse parameter-count surrogate,
  \[
    C_{\mathrm{params}}
    \;=\;
    \sqrt{\sum_{i=1}^d k_i^2\,c_{i-1}(c_i+1) \;/\; n}\,,
  \]
  with kernel sizes $k_i$ and channel counts $c_i$.
\end{itemize}

\subsection{Frobenius distance}
\begin{figure}[t]
  \centering
  \includegraphics[width=0.48\linewidth]{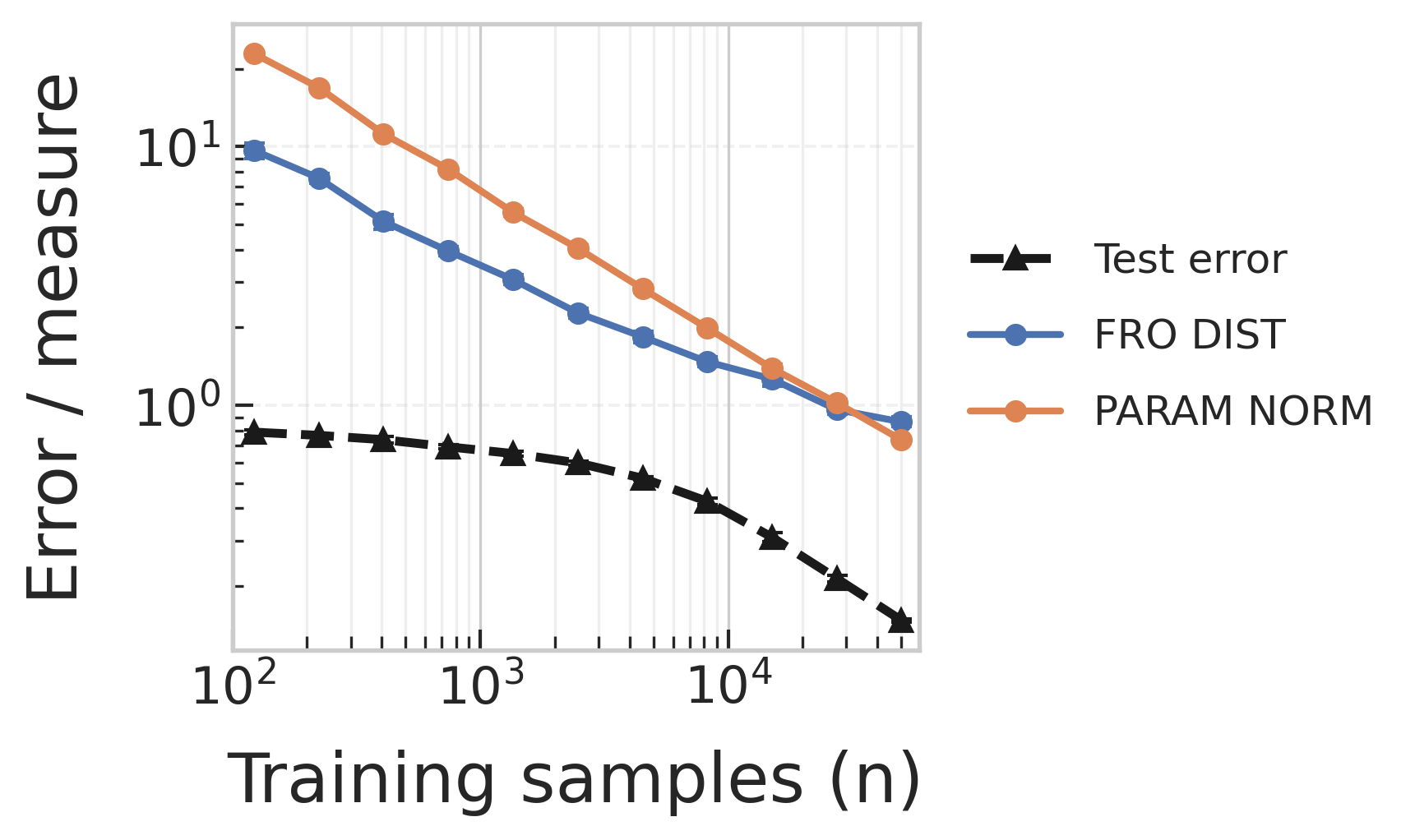}\hfill
  \includegraphics[width=0.48\linewidth]{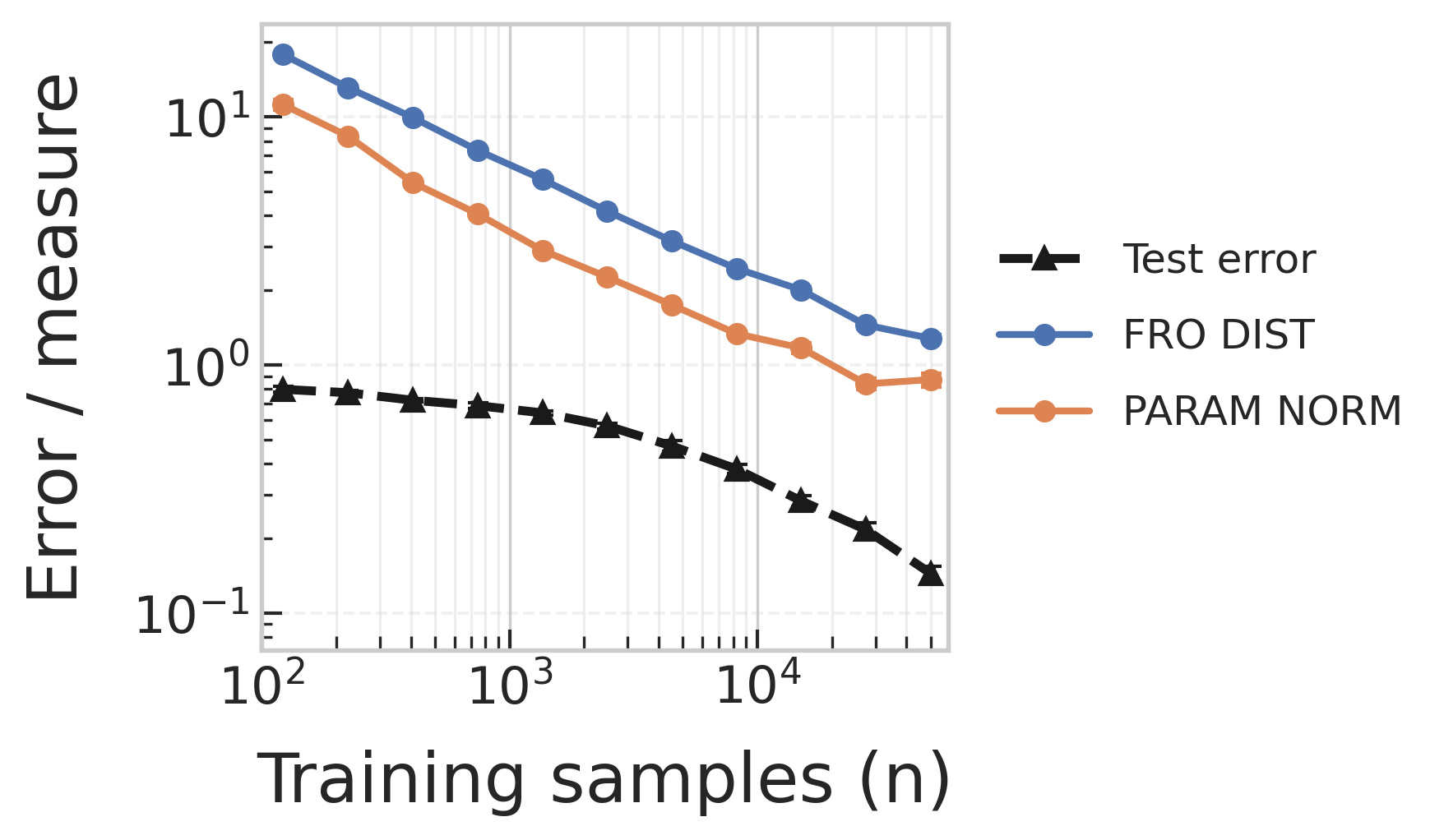}
  \caption{\textbf{Frobenius distance, ResNet-50 on CIFAR-10 (Adam).} Left: $\eta{=}10^{-3}$ shows a plateau and late collapse; right: $\eta{=}10^{-2}$ decays smoothly while accuracy tracks similarly.}
  \label{fig:frob-resnet-cifar-lr-shape}
\end{figure}

\begin{figure}[t]
  \centering
  \includegraphics[width=0.48\linewidth]{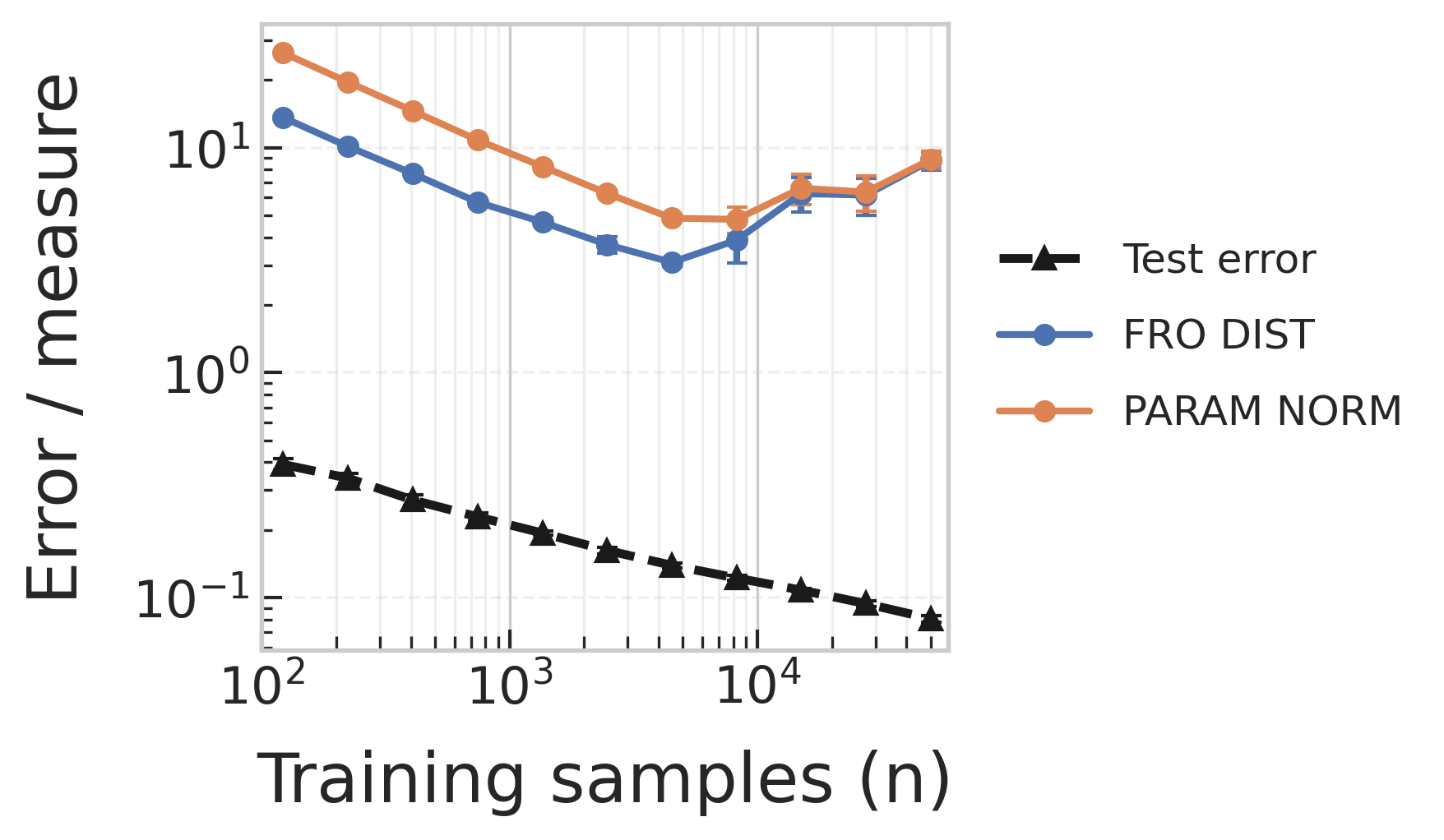}\hfill
  \includegraphics[width=0.48\linewidth]{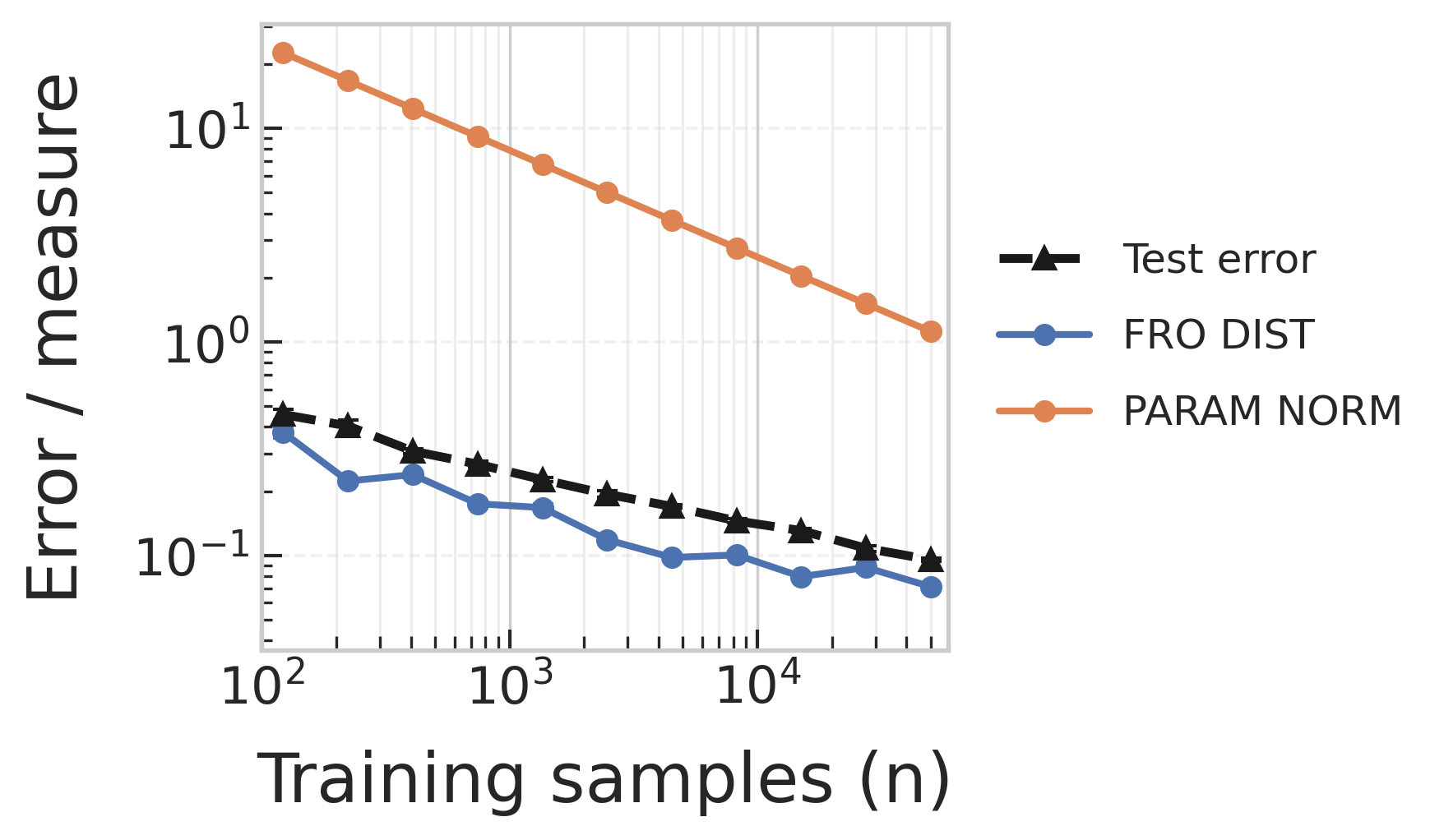}
  \caption{\textbf{Frobenius distance, DenseNet-121 on FashionMNIST ($\eta{=}10^{-2}$).} Left: \textsc{Adam} yields a U-shape; right: SGD+mom declines nearly log-linearly; in both cases test accuracy evolves similarly.}
  \label{fig:frob-densenet-fmnist-opt-shape}
\end{figure}

Two qualitatively different shapes appear when only the learning rate or optimizer changes, and they are visible in Fig.~\ref{fig:frob-resnet-cifar-lr-shape} and Fig.~\ref{fig:frob-densenet-fmnist-opt-shape}. On \textsc{ResNet-50}/CIFAR-10 with Adam, $\eta{=}10^{-3}$ produces a broad plateau followed by a late collapse, whereas $\eta{=}10^{-2}$ decays smoothly throughout (compare left vs.\ right in Fig.~\ref{fig:frob-resnet-cifar-lr-shape}). On \textsc{DenseNet-121}/FashionMNIST at fixed $\eta{=}10^{-2}$, Adam’s curve is U-shaped (drop then rebound), while momentum SGD traces a near log-linear decline (left vs.\ right in Fig.~\ref{fig:frob-densenet-fmnist-opt-shape}); the accuracy curves remain closely aligned in both pairs.

\subsection{Inverse margin}
\begin{figure}[t]
  \centering
  \includegraphics[width=0.48\linewidth]{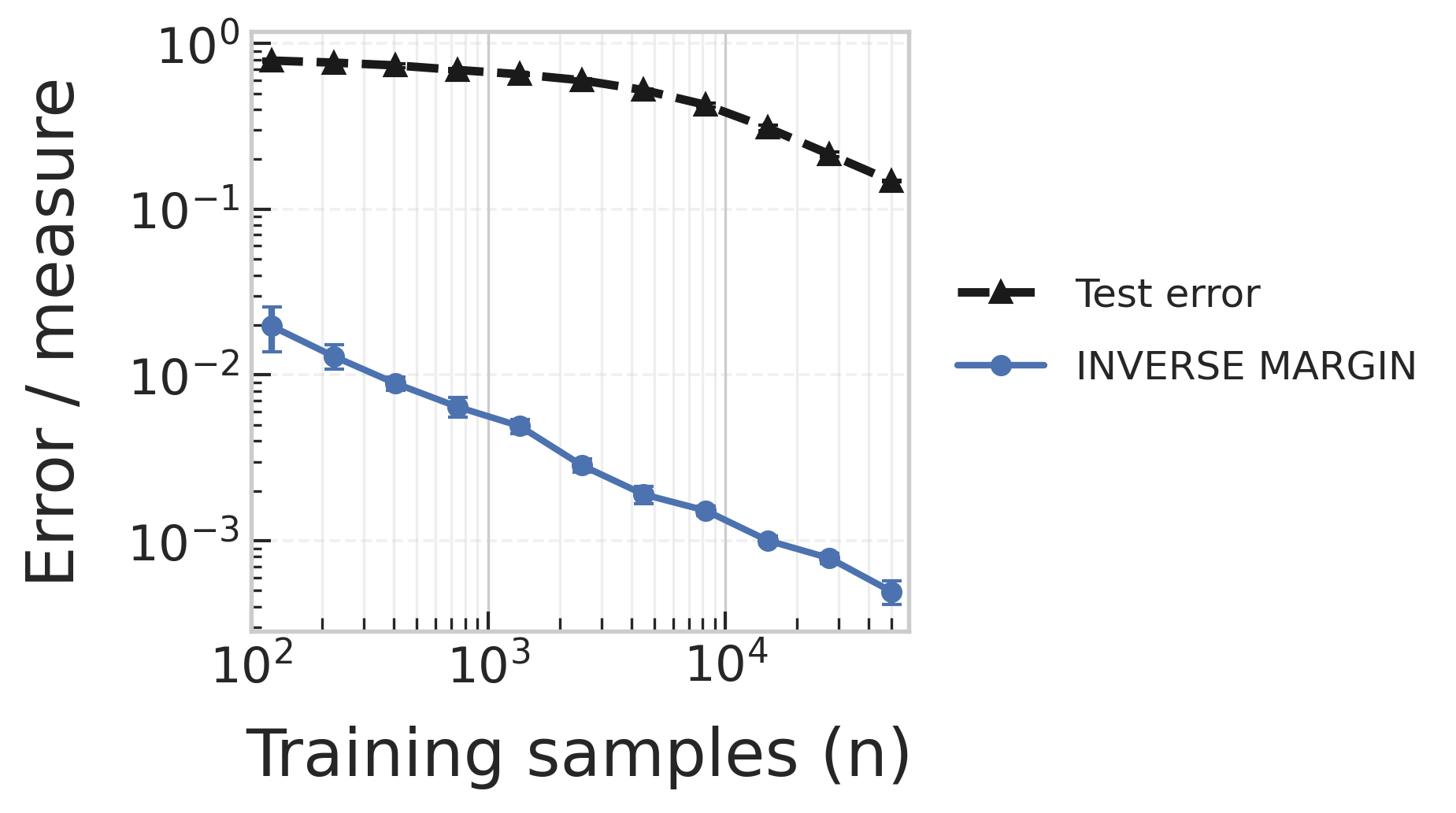}\hfill
  \includegraphics[width=0.48\linewidth]{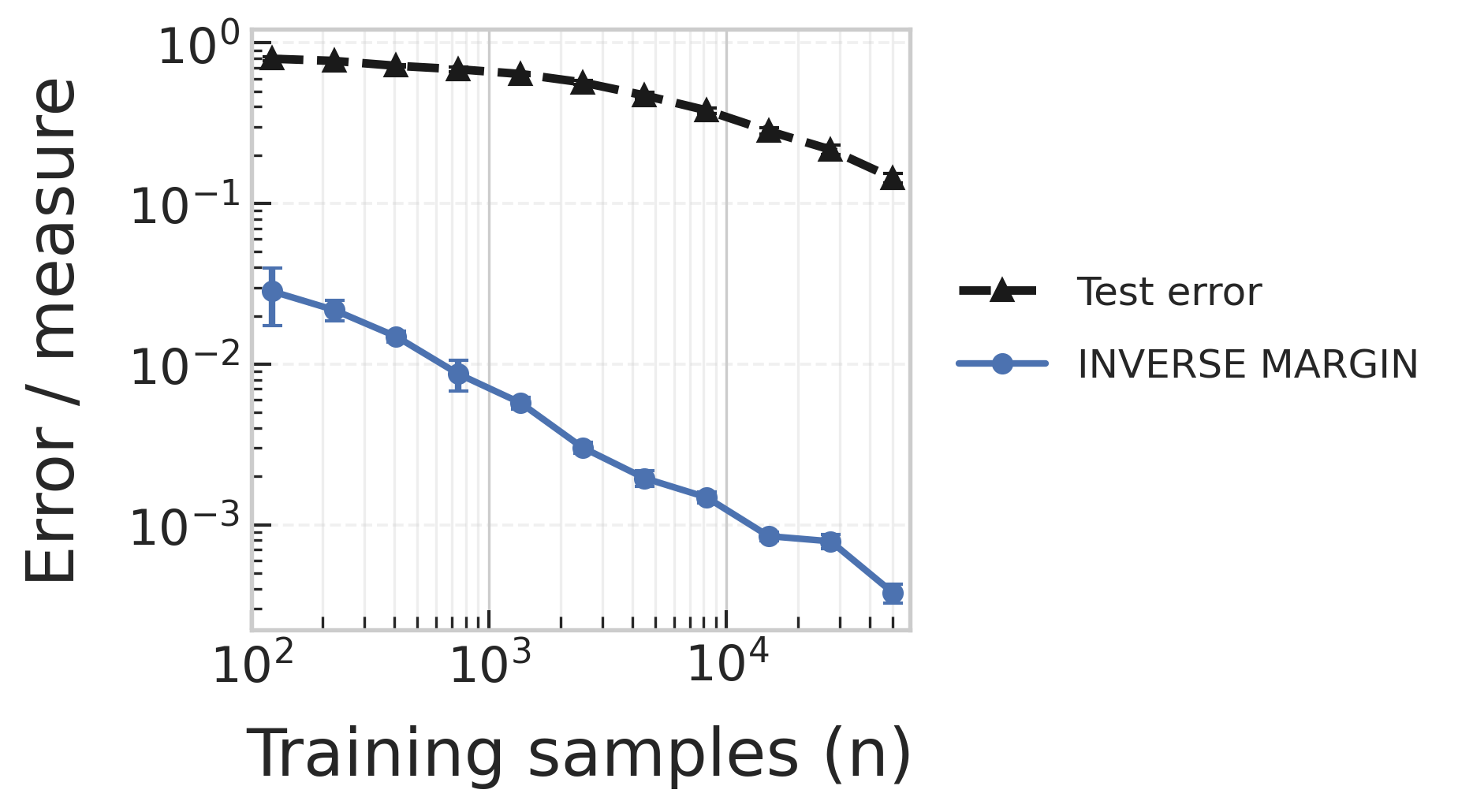}
  \caption{\textbf{Inverse margin, ResNet-50 on CIFAR-10 (Adam).} Left: $\eta{=}10^{-3}$ decays smoothly; right: $\eta{=}10^{-2}$ stalls then resumes, a kink absent from the accuracy curve.}
  \label{fig:margin-resnet-cifar-lr-shape}
\end{figure}

\begin{figure}[t]
  \centering
  \includegraphics[width=0.48\linewidth]{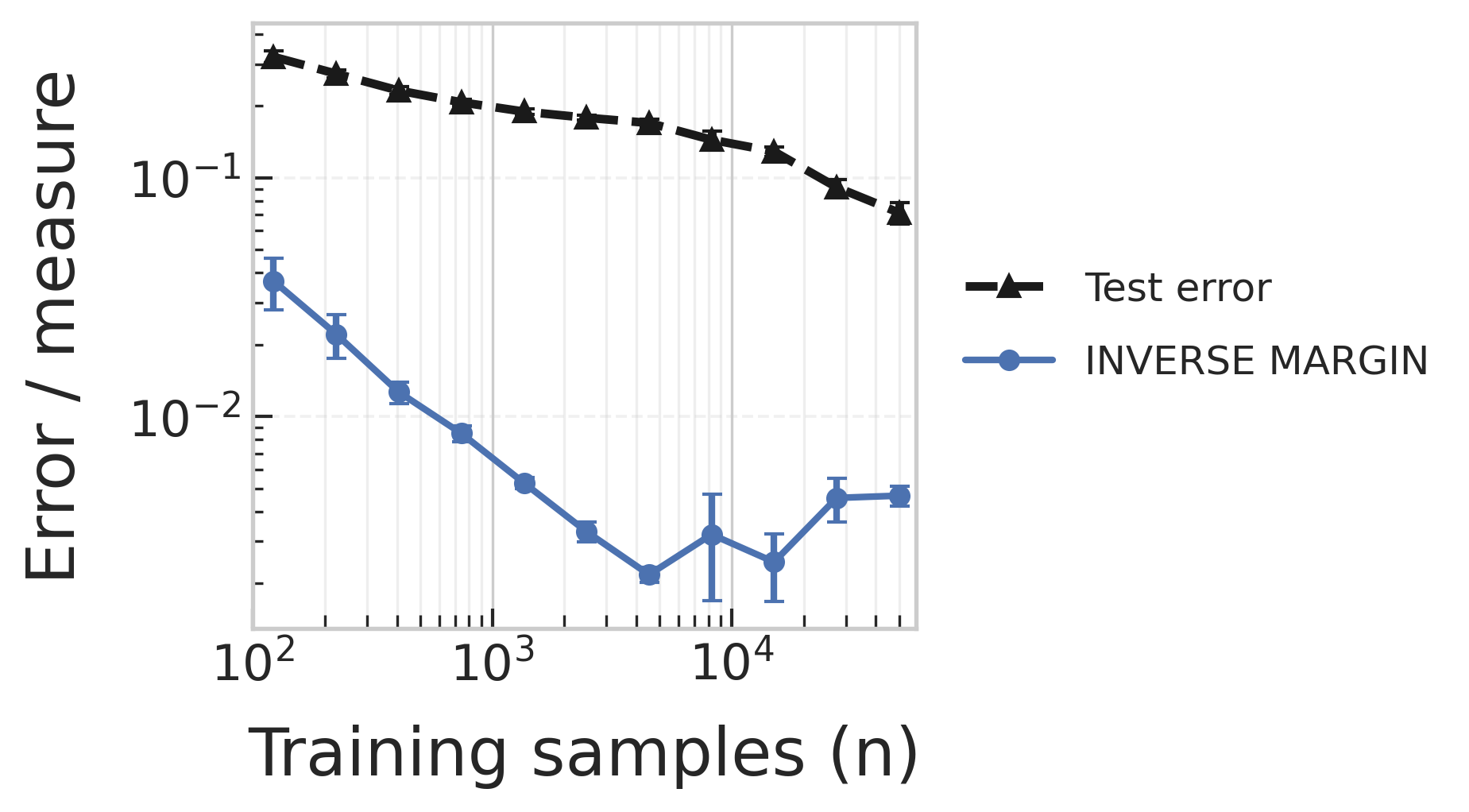}\hfill
  \includegraphics[width=0.48\linewidth]{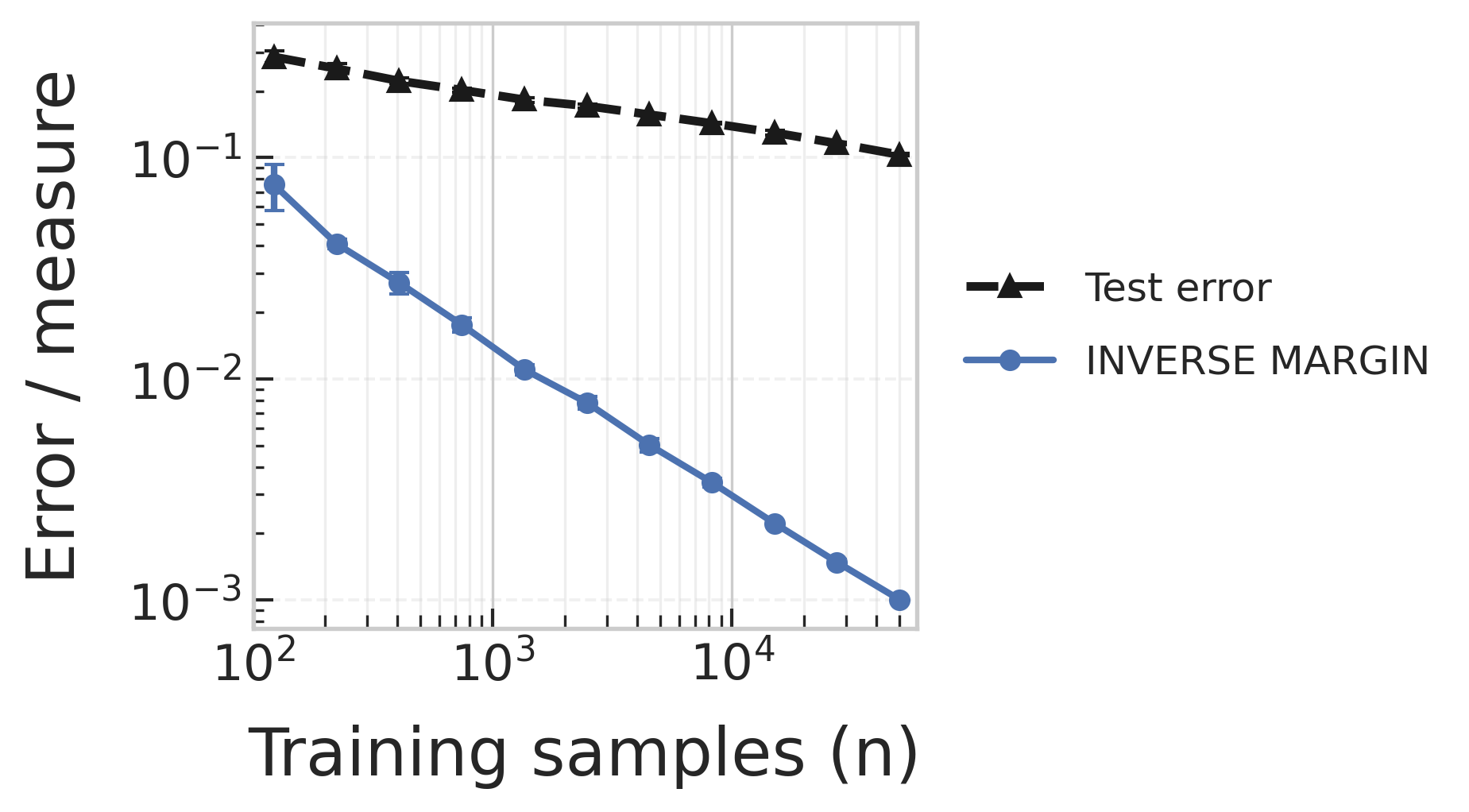}
  \caption{\textbf{Inverse margin, FCN on FashionMNIST ($\eta{=}10^{-2}$).} Left: \textsc{Adam} develops a mid-course bump; right: SGD+mom is monotone; both generalize similarly.}
  \label{fig:margin-fcn-fmnist-opt-shape}
\end{figure}

Figure~\ref{fig:margin-resnet-cifar-lr-shape} (left vs.\ right) shows that for \textsc{ResNet-50}/CIFAR-10 with Adam, $\eta{=}10^{-3}$ produces a smooth, power-law-like decline in inverse margin, whereas $\eta{=}10^{-2}$ stalls mid-training before resuming its drop; this kink has no analogue in the accuracy curve. At fixed $\eta{=}10^{-2}$ on \textsc{FCN}/FashionMNIST, \textsc{Adam} develops a clear bump after roughly $10^{3}$ samples while SGD’s curve decreases monotonically (Fig.~\ref{fig:margin-fcn-fmnist-opt-shape}); again, both reach similar generalization.

\subsection{Spectral metrics}
\begin{figure}[t]
  \centering
  \includegraphics[width=0.48\linewidth]{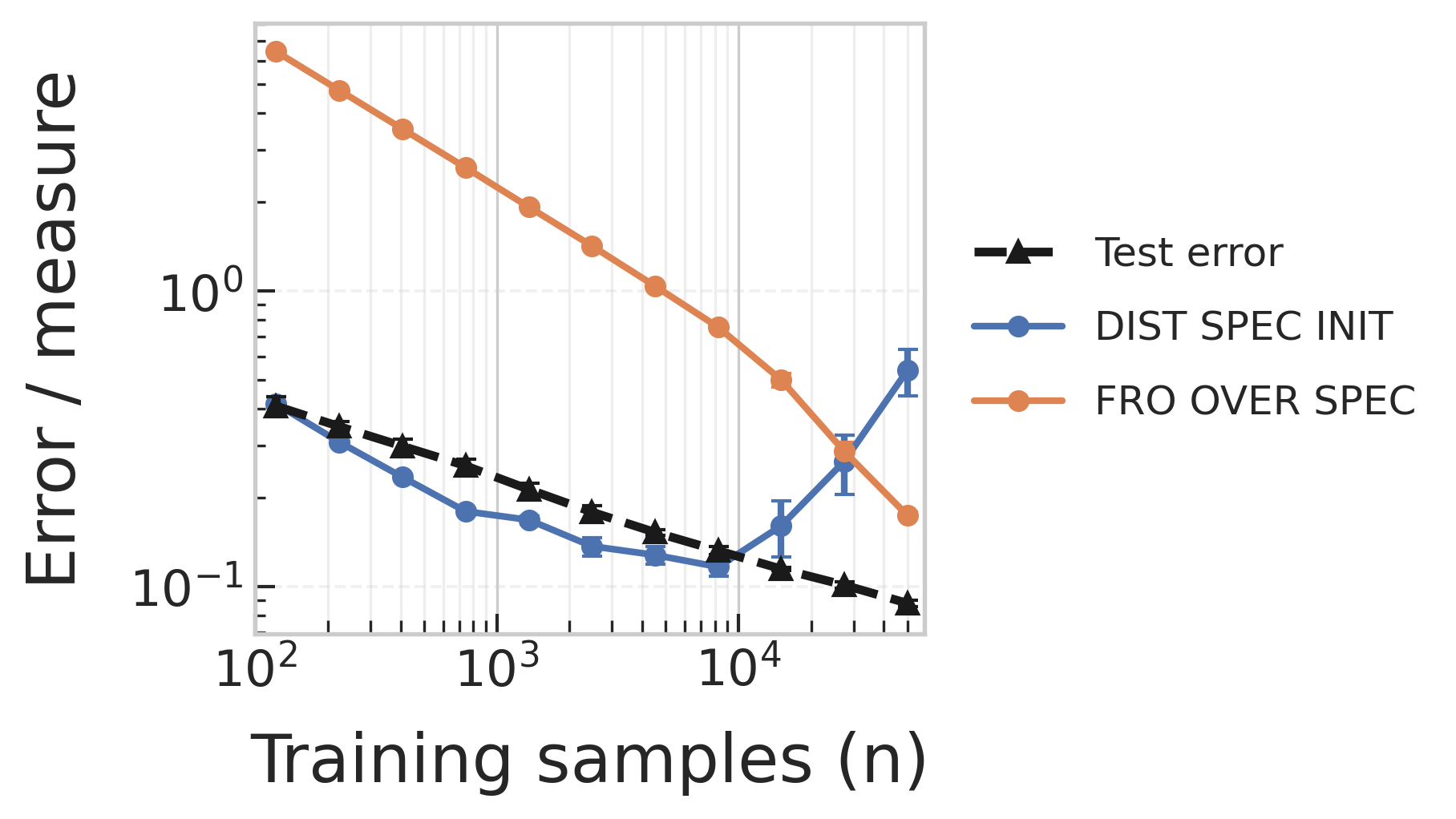}\hfill
  \includegraphics[width=0.48\linewidth]{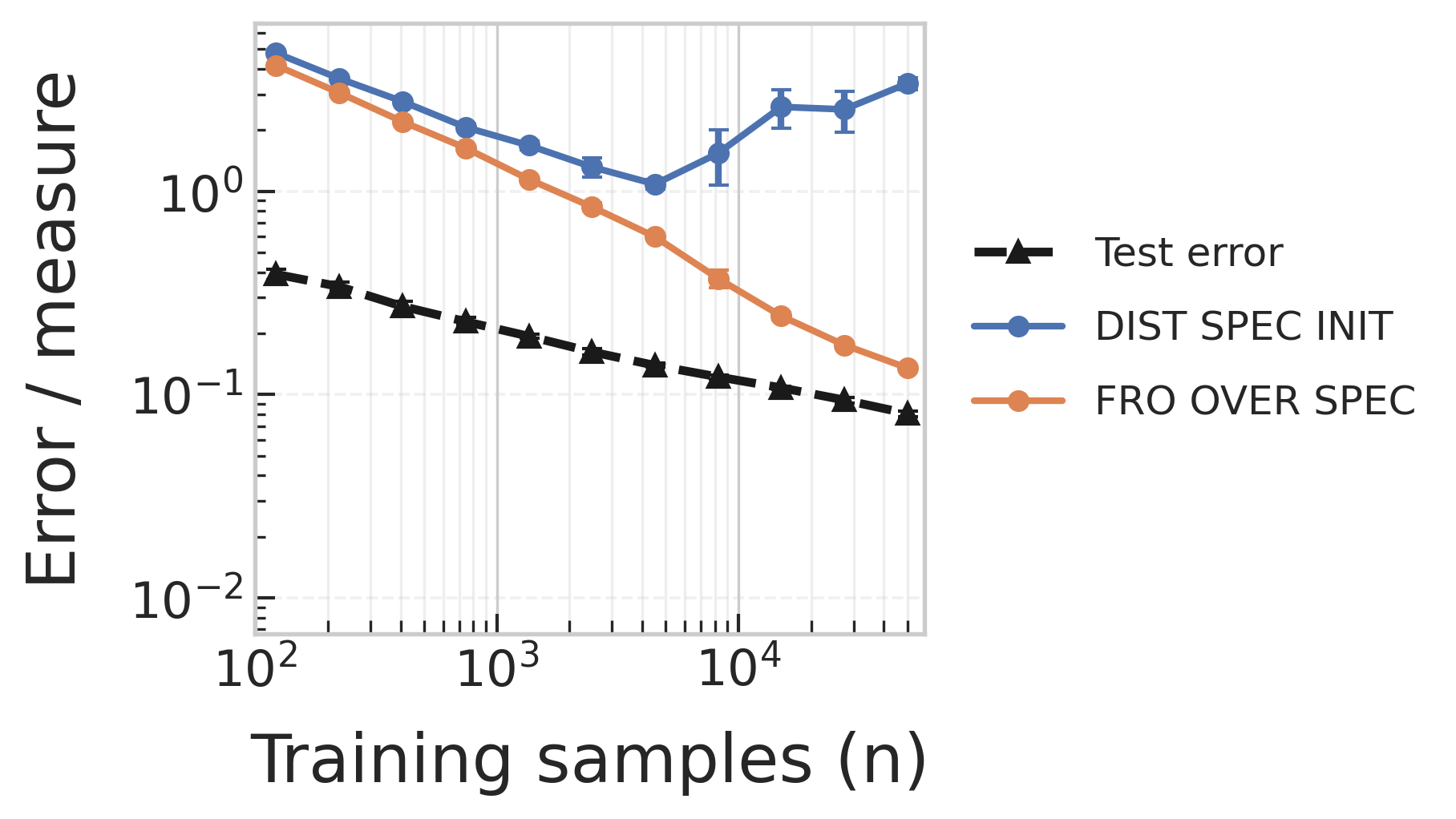}
  \caption{\textbf{Spectral metrics, DenseNet-121 on FashionMNIST (Adam).} Left: $\eta{=}10^{-3}$ drops then spikes, crossing FRO-OVER-SPEC; right: $\eta{=}10^{-2}$ stays monotone but reverses ordering.}
  \label{fig:spectral-densenet-fmnist-lr-shape}
\end{figure}

\begin{figure}[t]
  \centering
  \includegraphics[width=0.48\linewidth]{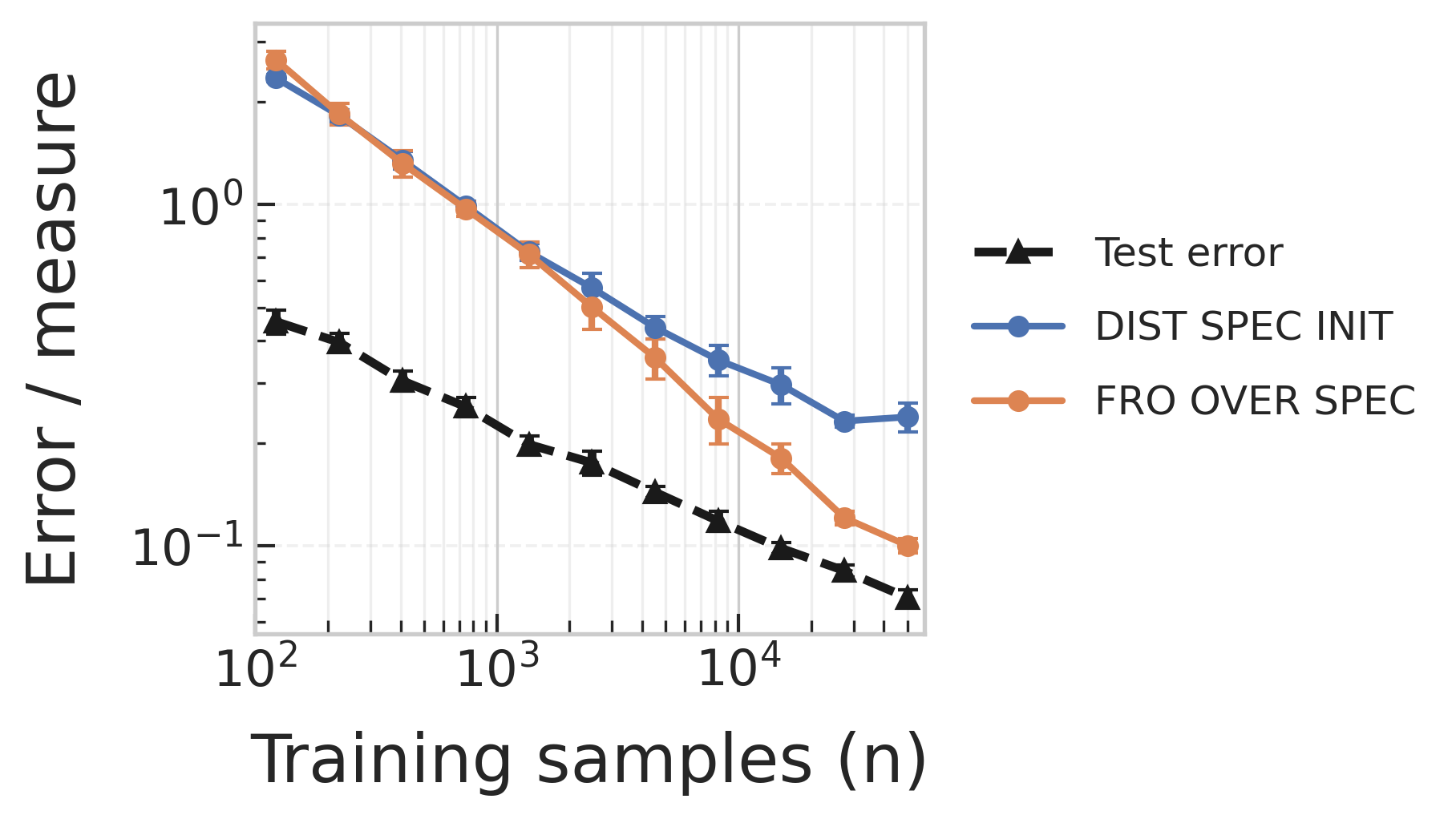}\hfill
  \includegraphics[width=0.48\linewidth]{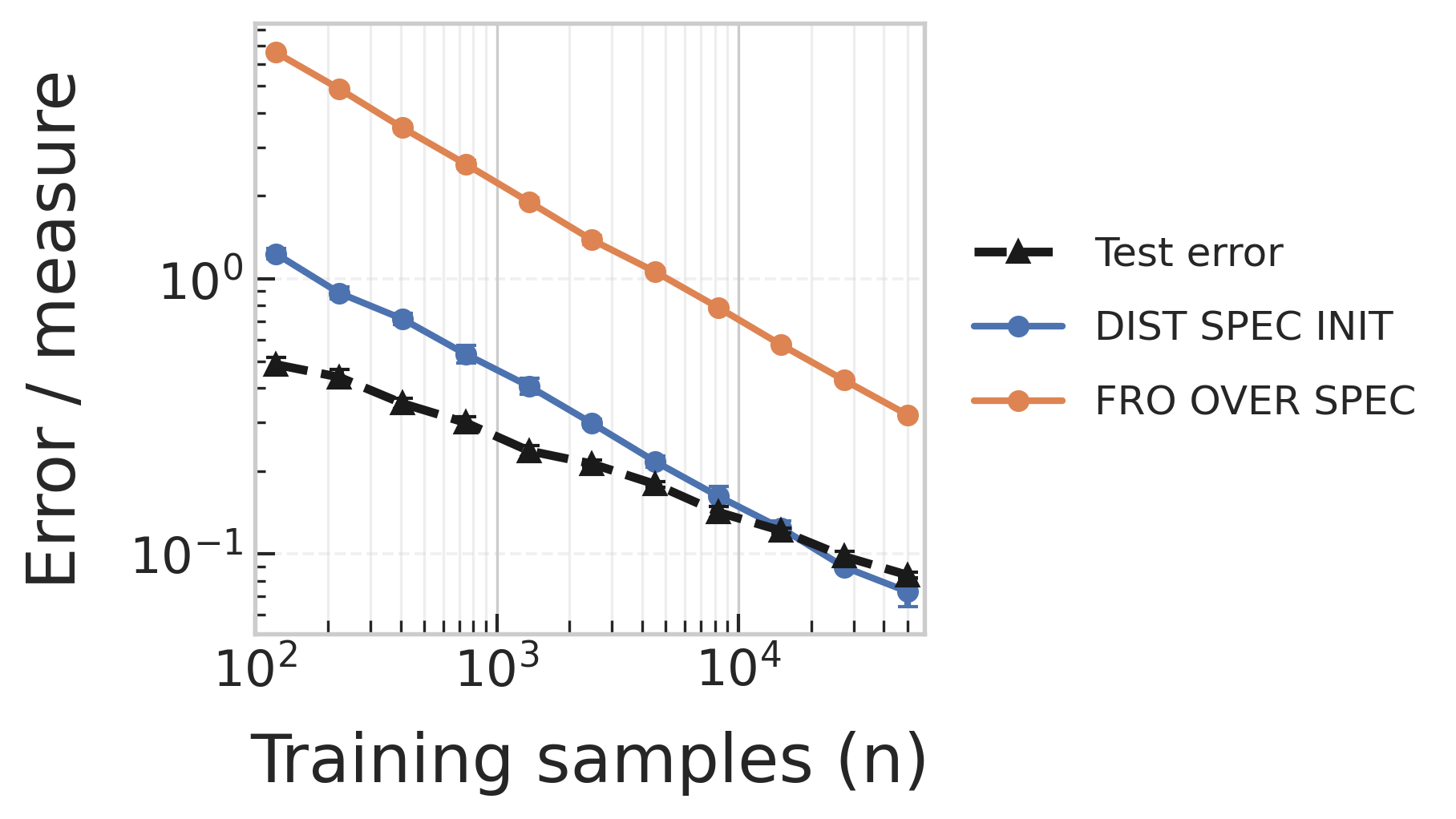}
  \caption{\textbf{Spectral metrics, ResNet-50 on FashionMNIST ($\eta{=}10^{-2}$).} Left: \textsc{Adam} exhibits a valley then rise; right: SGD+mom declines steadily; accuracy is similar.}
  \label{fig:spectral-resnet-fmnist-opt-shape}
\end{figure}

Spectral surrogates display late spikes and order reversals under the same minimal tweaks. With \textsc{DenseNet-121}/FashionMNIST and Adam, $\eta{=}10^{-3}$ sends the distance-from-initialization in spectral norm down and then sharply up late in training, crossing the \emph{FRO-OVER-SPEC} curve, whereas $\eta{=}10^{-2}$ keeps both traces monotone but flips their ordering (compare panels in Fig.~\ref{fig:spectral-densenet-fmnist-lr-shape}). On \textsc{ResNet-50}/FashionMNIST at $\eta{=}10^{-2}$, \textsc{Adam} shows a valley then rise, while SGD declines steadily (Fig.~\ref{fig:spectral-resnet-fmnist-opt-shape}); accuracy curves overlap in both pairs.

\subsection{PAC-Bayes bounds}
\begin{figure}[t]
  \centering
  \includegraphics[width=0.48\linewidth]{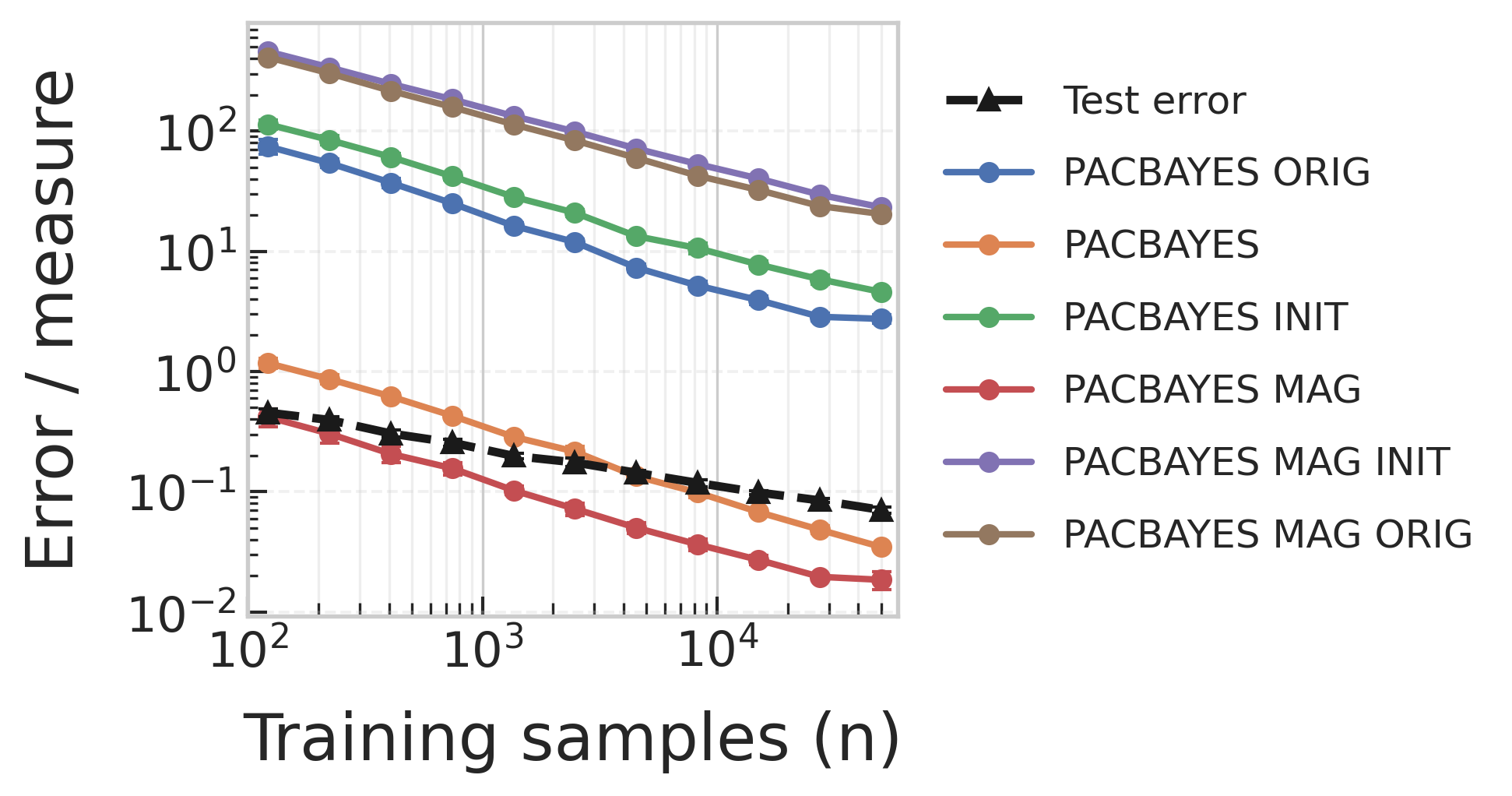}\hfill
  \includegraphics[width=0.48\linewidth]{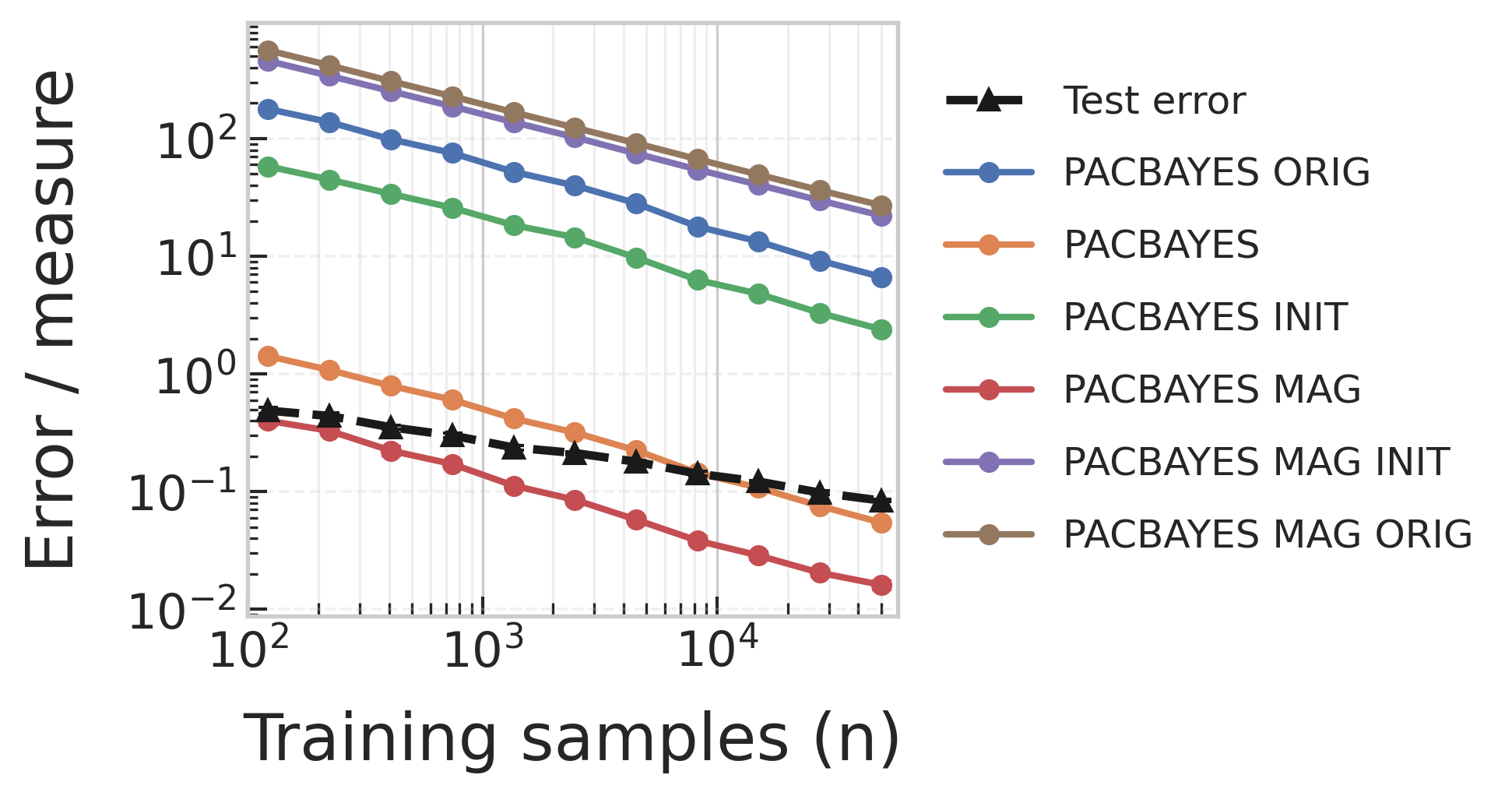}
  \caption{\textbf{PAC-Bayes, ResNet-50 on FashionMNIST ($\eta{=}10^{-2}$).} Left: \textsc{Adam} produces curved, kinked trajectories; right: SGD+mom is nearly straight; test error changes little.}
  \label{fig:pacbayes-resnet-fmnist-opt-shape}
\end{figure}

\begin{figure}[t]
  \centering
  \includegraphics[width=0.48\linewidth]{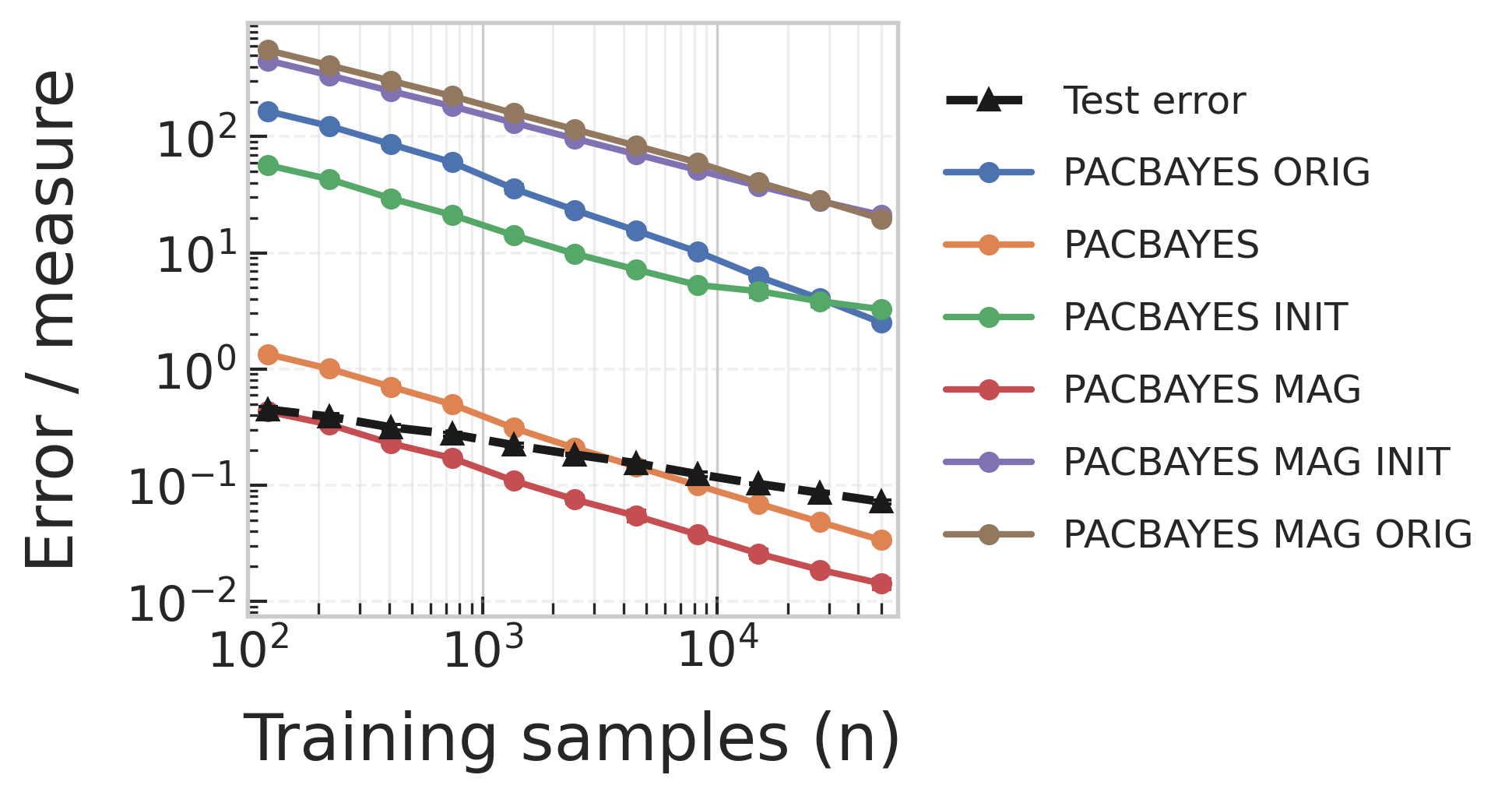}\hfill
  \includegraphics[width=0.48\linewidth]{pics_pdf/learning_curve_sep_2025_log/RESNET50-FashionMNIST/optimizer_ADAM__stop_true__lr_0.01/optimizer_ADAM__stop_true__lr_0.01__PAC-Bayes.png}
  \caption{\textbf{PAC-Bayes, ResNet-50 on FashionMNIST (Adam).} Left: $\eta{=}10^{-3}$; right: $\eta{=}10^{-2}$. Increasing $\eta$ induces late-training order swaps (e.g., {\sc MAG} vs.\ {\sc ORIG}) absent from the accuracy curves.}
  \label{fig:pacbayes-resnet-fmnist-lr-shape}
\end{figure}

Optimizer swaps and moderate learning-rate changes also produce curved versus straight “bound profiles” and late order crossings. For \textsc{ResNet-50}/FashionMNIST at $\eta{=}10^{-2}$, \textsc{Adam} yields kinked “banana” trajectories across variants, whereas SGD renders near-straight lines (Fig.~\ref{fig:pacbayes-resnet-fmnist-opt-shape}). Holding \textsc{Adam} fixed and raising $\eta$ from $10^{-3}$ to $10^{-2}$ reorders the variants late in training—for example, \textsc{PACBAYES MAG} and \textsc{PACBAYES ORIG} swap rank—even though accuracy shows no such crossing (Fig.~\ref{fig:pacbayes-resnet-fmnist-lr-shape}).

\subsection{VC-dimension proxy (robust baseline)}
\begin{figure}[t]
  \centering
  \includegraphics[width=0.48\linewidth]{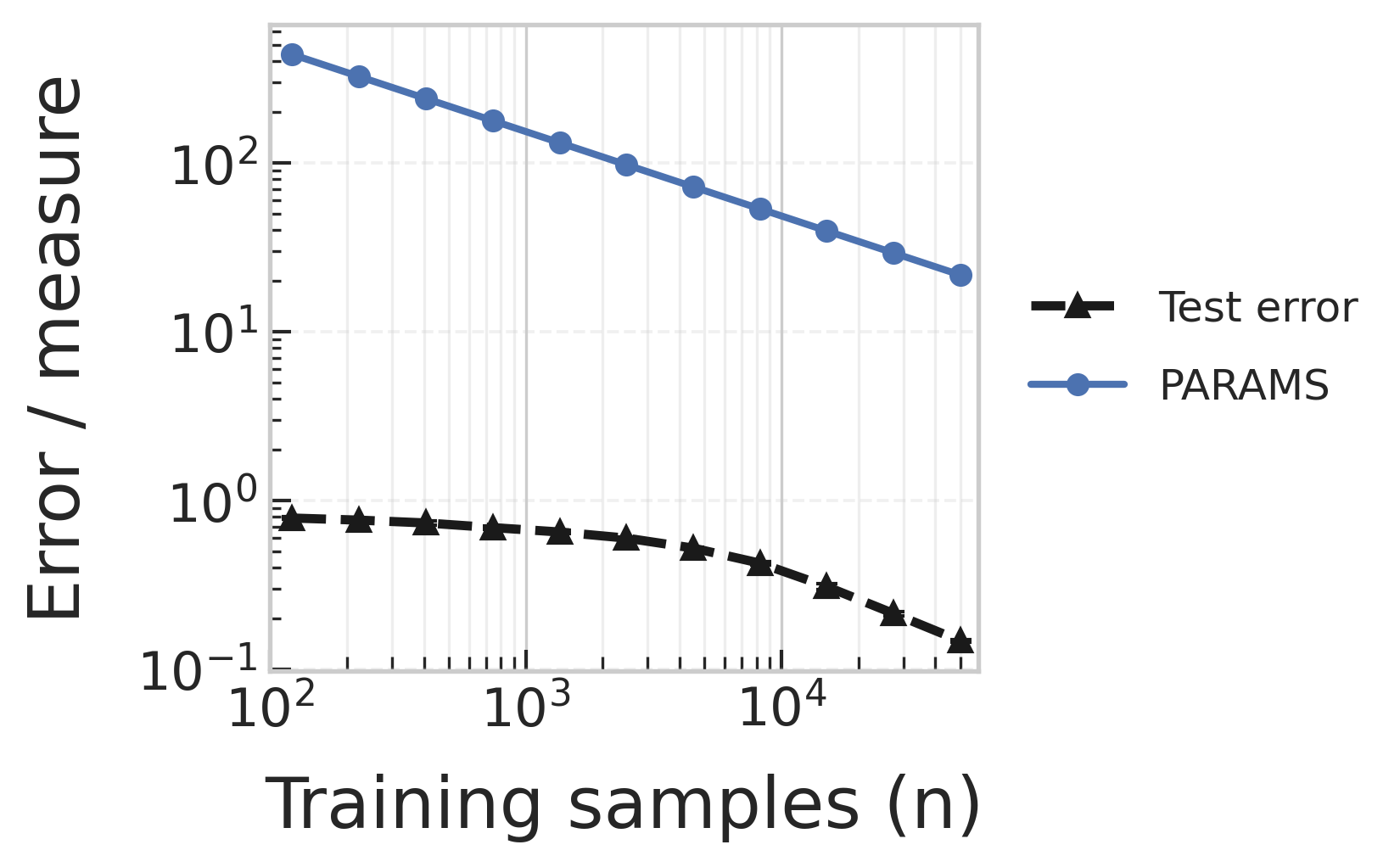}\hfill
  \includegraphics[width=0.48\linewidth]{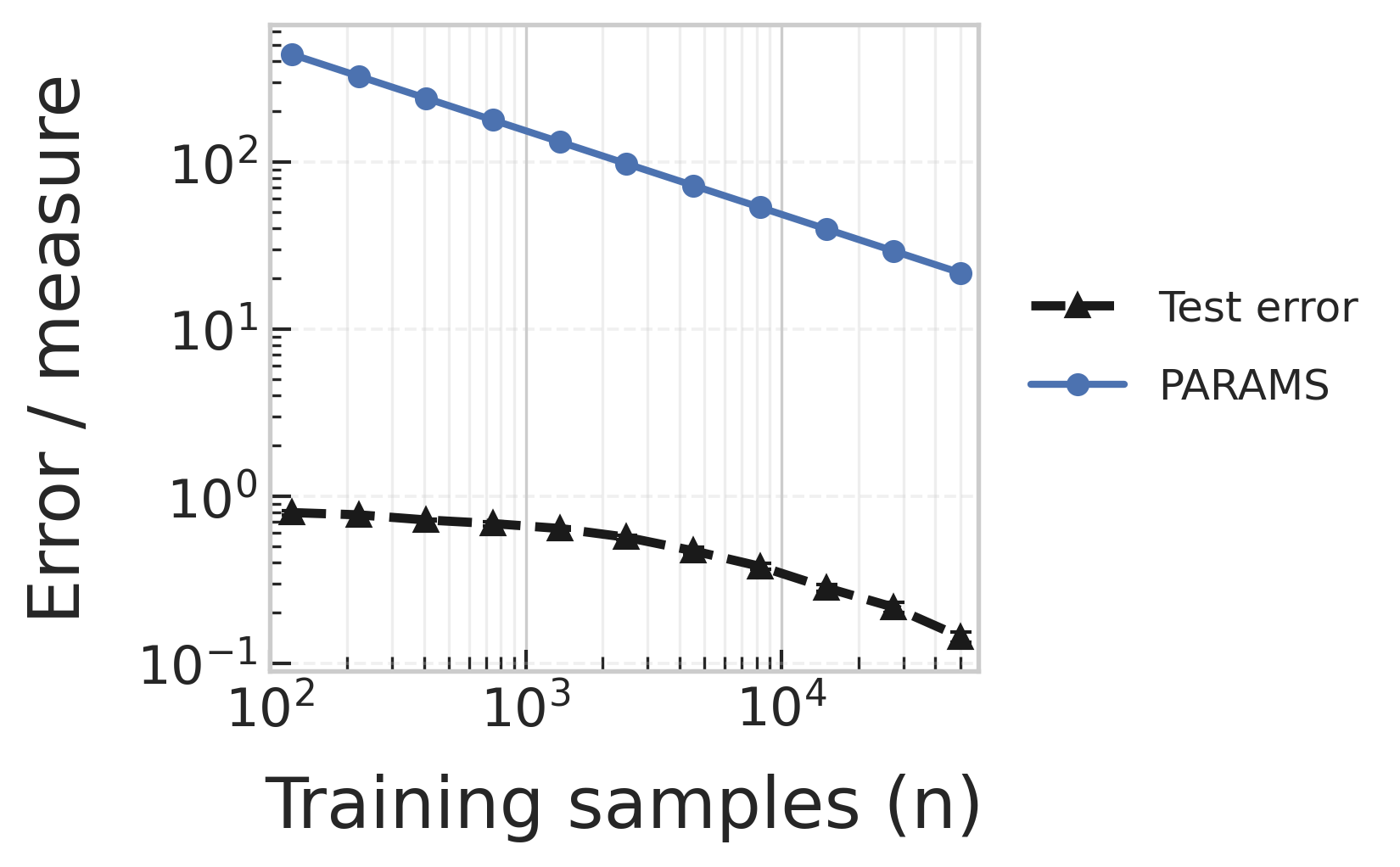}
  \caption{\textbf{VC-dimension proxy, ResNet-50 on CIFAR-10 (Adam).} Left: $\eta{=}10^{-3}$; right: $\eta{=}10^{-2}$. Nearly identical monotone decays; a useful control.}
  \label{fig:vc-resnet-cifar-lr-robust}
\end{figure}

As a control, Fig.~\ref{fig:vc-resnet-cifar-lr-robust} shows that the parameter-count proxy for \textsc{ResNet-50}/CIFAR-10 with \textsc{Adam} is strikingly shape-stable across $\eta{=}10^{-3}$ and $\eta{=}10^{-2}$: both traces are near-identical monotone decays, tracking one another while accuracy also aligns. This pair serves as a rare counterexample resistant to qualitative shifts.

\medskip
Across all families above, the figures make a consistent point: changing only the learning rate or swapping the optimizer can reshape the surrogate’s learning curve—introducing plateaus, rebounds, late spikes, or crossings—without a commensurate shift in test accuracy. This shape-level fragility extends far beyond Path norms and cautions against reading causal explanations of generalization from any single surrogate without dedicated stress tests \citep{dziugaite2020in,jiang2019fantastic}.

\section{Additional temporal behavior results: optimizer sensitivity across measure families}
\label{app:temporal-behavior}

This section extends the temporal analysis in Section~\ref{sec:temporal-fragility} by holding the task and hyperparameters fixed and changing only the optimizer. We train ResNet--50 on FashionMNIST with learning rate $0.01$ and no early stopping, and we track each measure across epochs. All panels use a logarithmic epoch axis; this makes the early regime and successive orders of magnitude more legible, while very late additive-only epochs occupy little horizontal extent unless they span a substantial multiplicative range. The red dashed vertical line in every panel marks the first epoch at which training accuracy reaches $100\%$; on a log axis this event is still easy to spot, but short post-interpolation intervals can appear visually narrow if they do not cover a large multiplicative window. To avoid duplication with the main text, we omit the path-norm panels here and focus on complementary families whose behavior further illustrates optimizer sensitivity.

The PAC--Bayes family provides a clean illustration of this theme. Under \textsc{Adam}, multiple bounds show slow, persistent growth after interpolation; on a log-time axis this appears as a steady positive slope across late decades of epochs, most clearly for \texttt{PACBAYES\_ORIG} and \texttt{PACBAYES\_INIT}. With momentum \textsc{SGD}, the same curves remain essentially flat within error bars once the dashed line is crossed, and on the log axis they sit nearly horizontal, emphasizing stability rather than drift.

\begin{figure}[t]
\centering
\subfigure[Adam]{%
  \includegraphics[width=.48\linewidth]{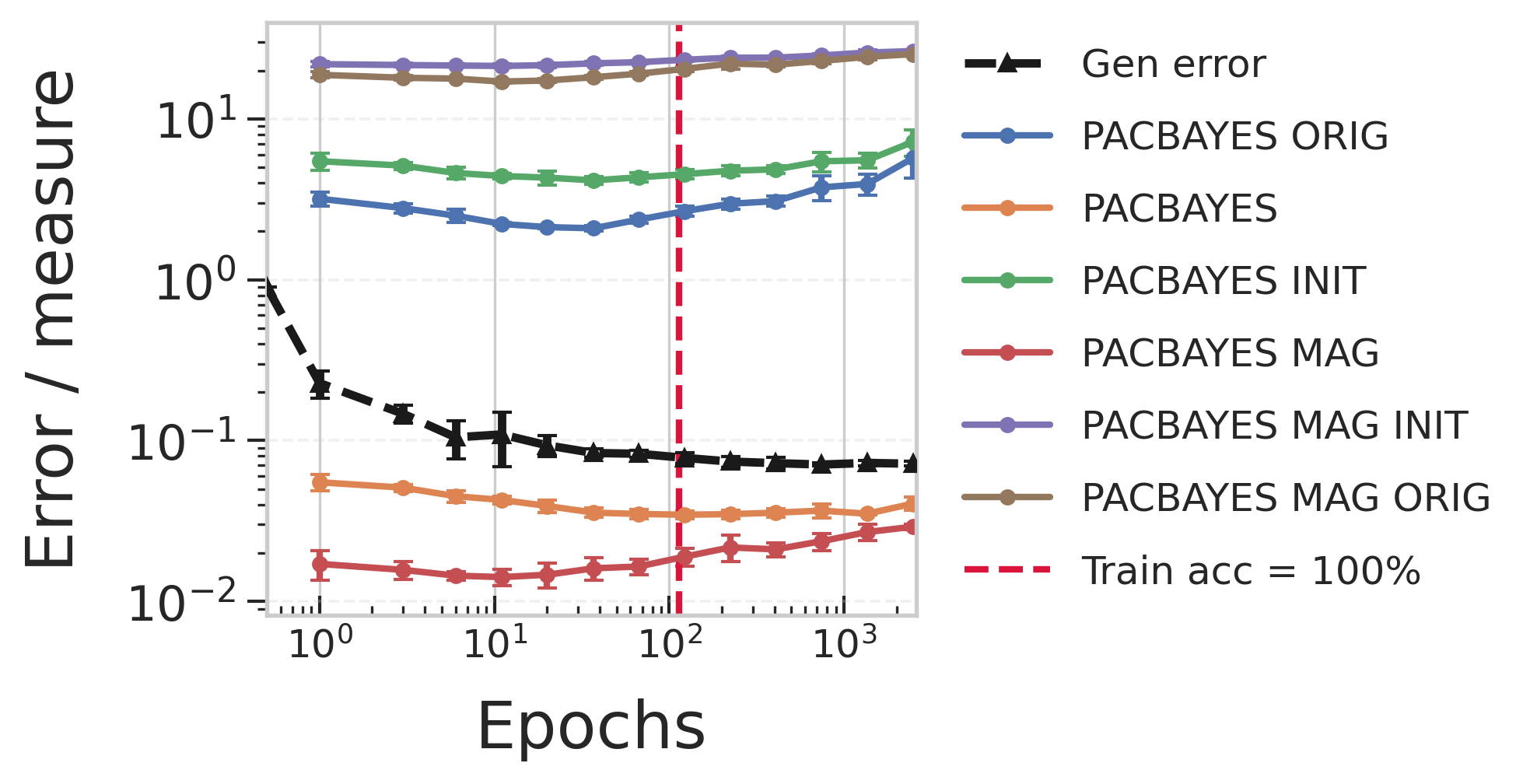}}
\hfill
\subfigure[\textsc{SGD} with momentum]{%
  \includegraphics[width=.48\linewidth]{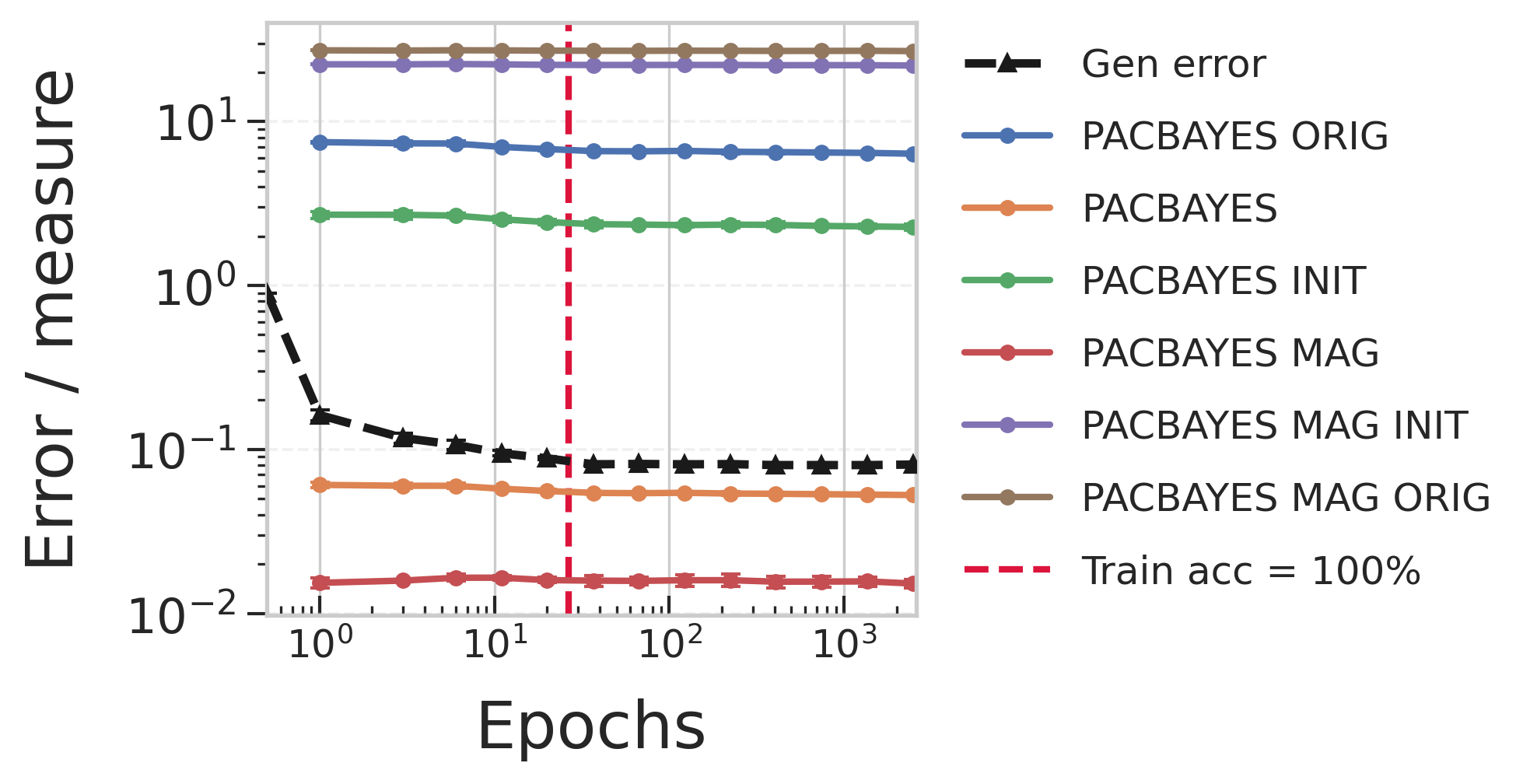}}
\caption{Temporal behavior for PAC--Bayes variants on ResNet--50/FashionMNIST at fixed learning rate $0.01$. The epoch axis is logarithmic and the red dashed vertical line marks the first $100\%$ training-accuracy epoch. \textsc{Adam} exhibits slow post‑accuracy growth (notably in \texttt{PACBAYES\_ORIG} and \texttt{PACBAYES\_INIT}), whereas \textsc{SGD} with momentum keeps the family flat once the dashed line is crossed.}
\label{fig:temporal-pacbayes}
\end{figure}

Measures tied to weight scale show the starkest divergence. In the Frobenius panel, \textsc{Adam} drives both the distance to initialization and the parameter norm upward almost monotonically after the model has interpolated; on a log-time axis this shows up as a persistent positive slope across late epochs. By contrast, momentum \textsc{SGD} leaves both traces effectively horizontal once the dashed line is passed, highlighting a stable plateau.

\begin{figure}[t]
\centering
\subfigure[Adam]{%
  \includegraphics[width=.48\linewidth]{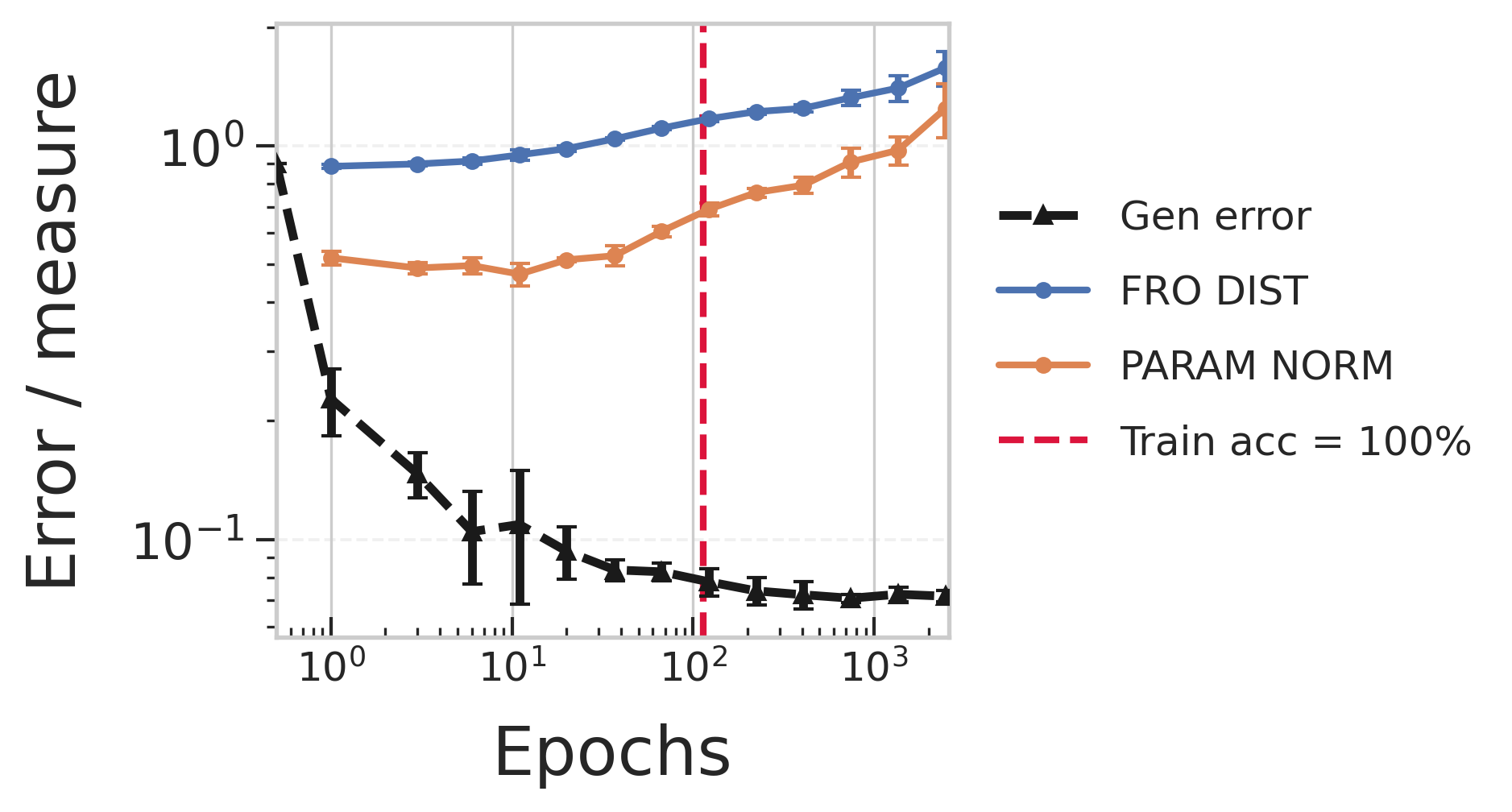}}
\hfill
\subfigure[\textsc{SGD} with momentum]{%
  \includegraphics[width=.48\linewidth]{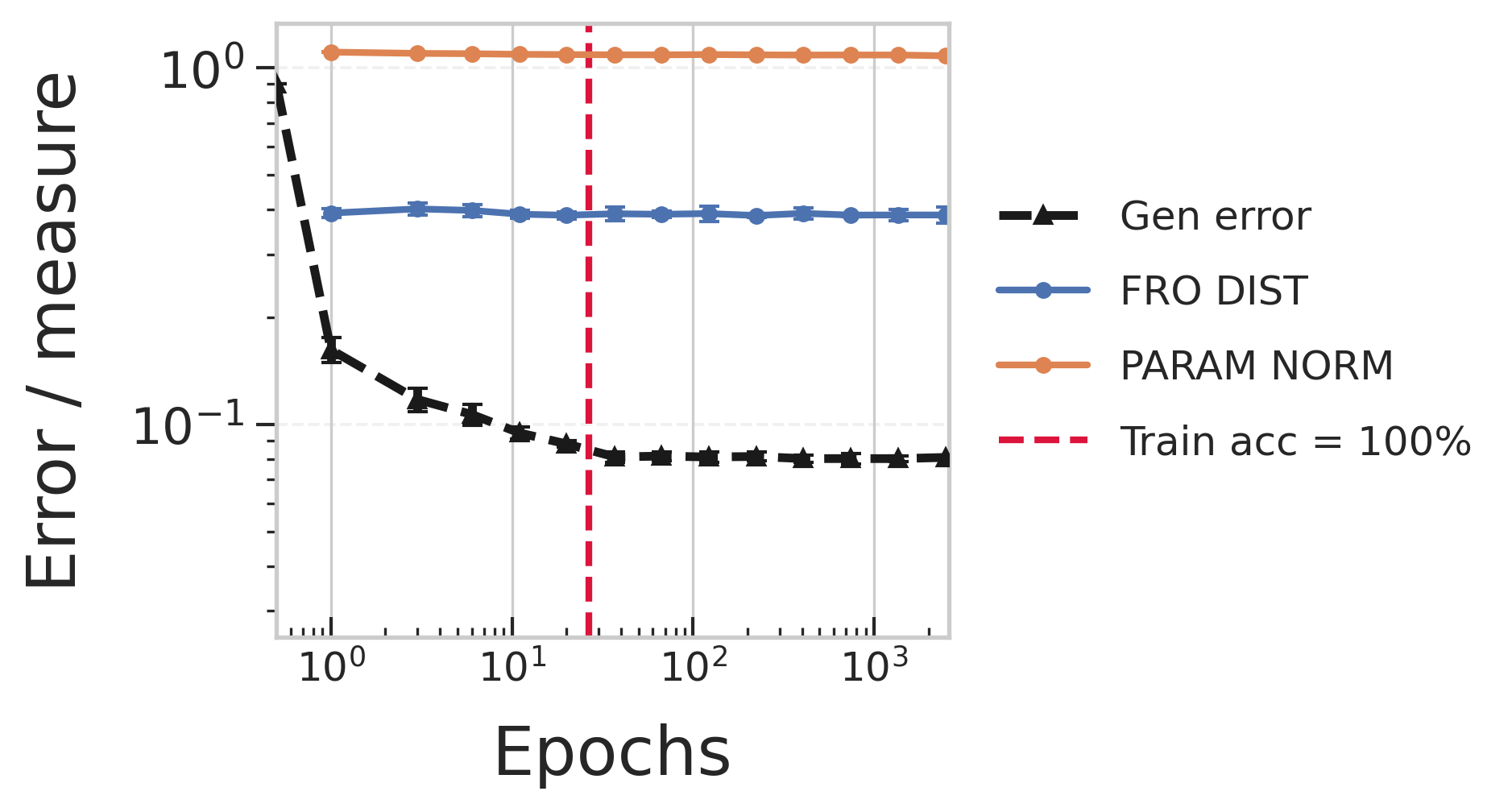}}
\caption{Frobenius distance and parameter norm through time. The epoch axis is logarithmic. \textsc{Adam} produces continued growth in both \texttt{FRO\_DIST} and \texttt{PARAM\_NORM} after $100\%$ training accuracy, while \textsc{SGD} with momentum holds them near a constant level.}
\label{fig:temporal-frobenius}
\end{figure}

Not every family bends under this perturbation. Both optimizers rapidly reduce the inverse‑margin surrogate early in training and then hold it near a floor. The log-time axis spreads out the initial drop, making the shared shape and timing transparent; \textsc{Adam} settles at a slightly higher level, but the trajectories otherwise coincide.

\begin{figure}[t]
\centering
\subfigure[Adam]{%
  \includegraphics[width=.48\linewidth]{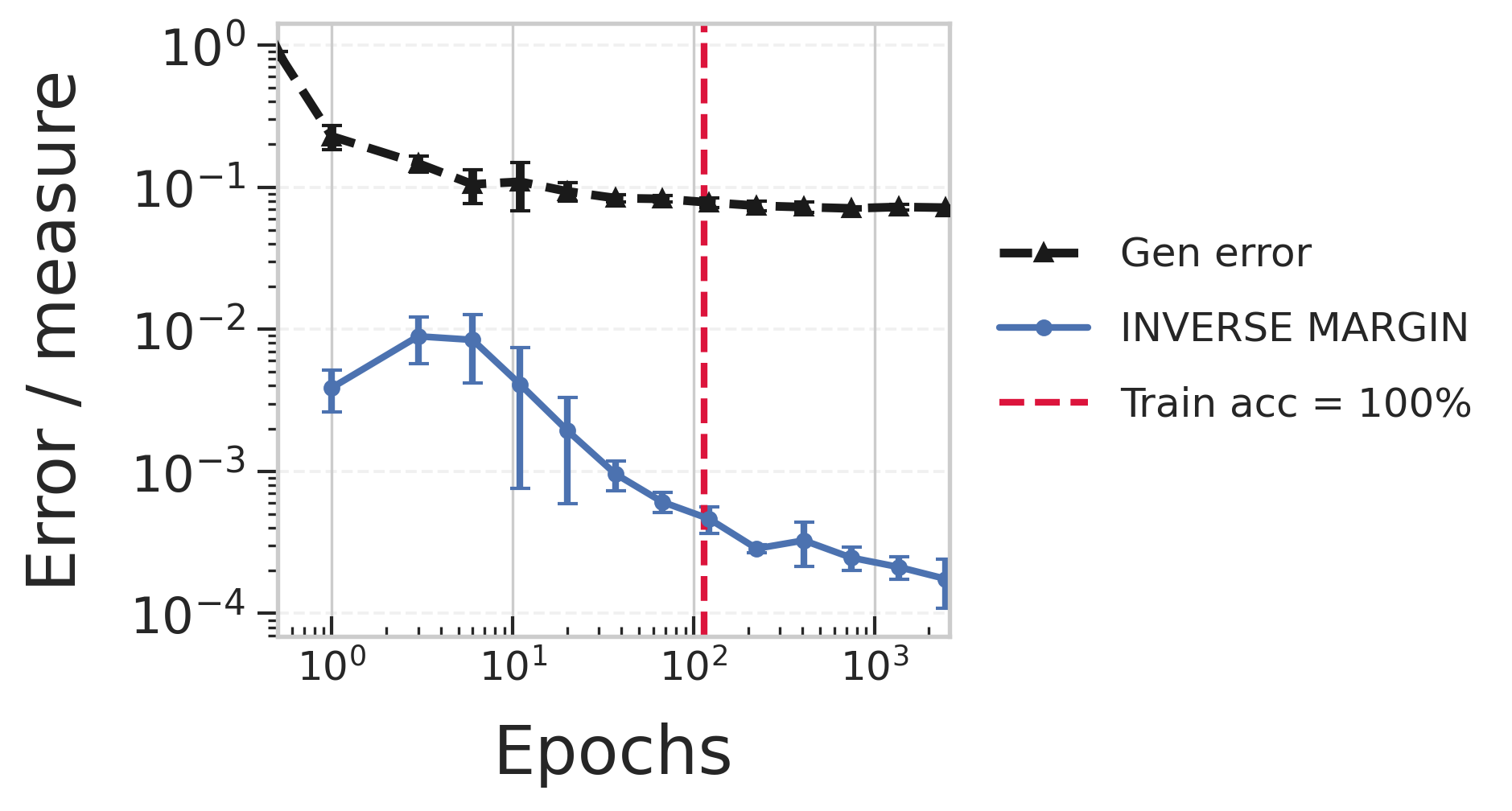}}
\hfill
\subfigure[\textsc{SGD} with momentum]{%
  \includegraphics[width=.48\linewidth]{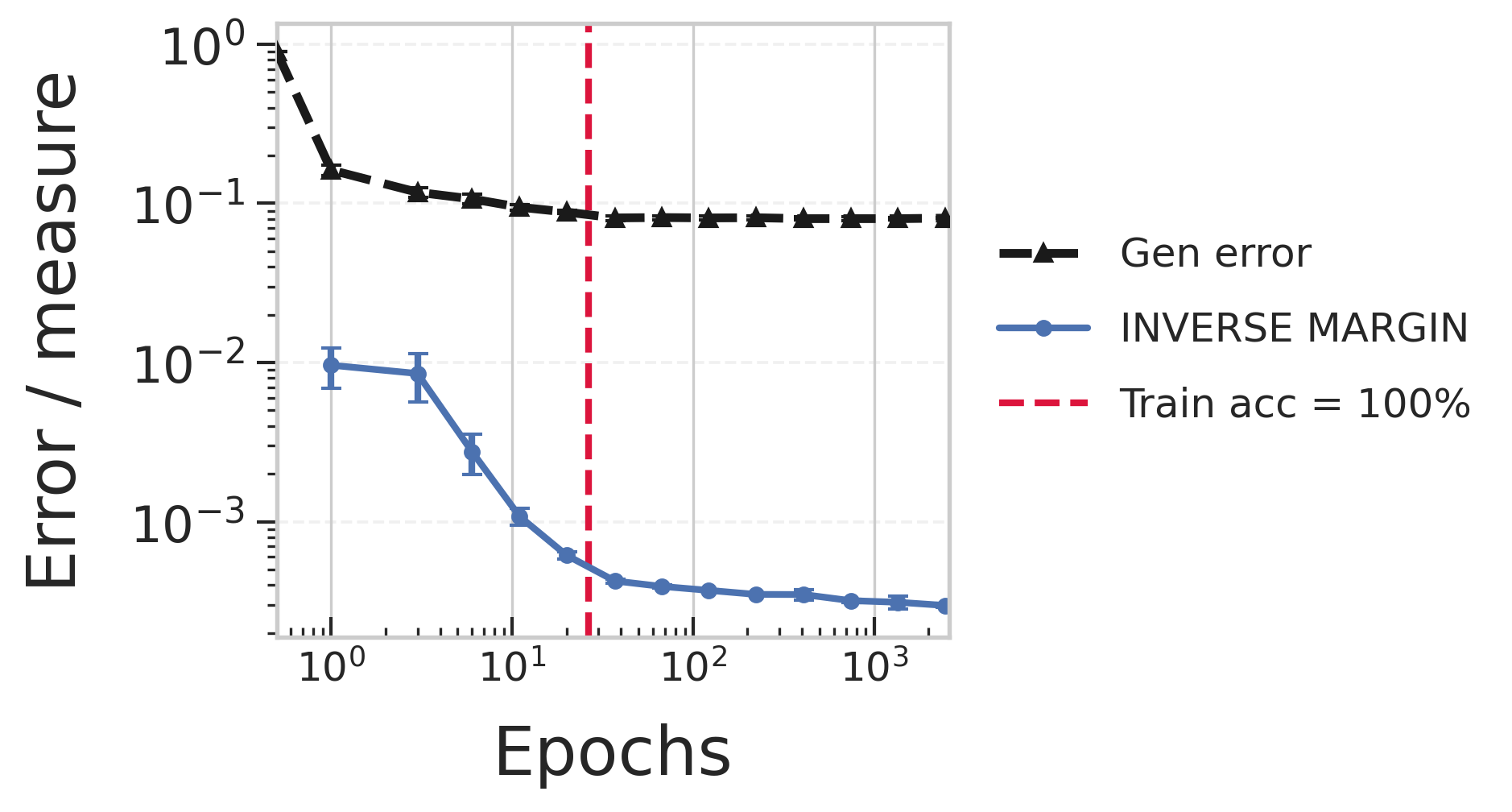}}
\caption{Inverse‑margin surrogate through time on a logarithmic epoch axis. Both optimizers shrink the measure quickly and then hover near a floor; \textsc{Adam} stabilizes at a slightly higher level but the overall shape is shared.}
\label{fig:temporal-margin}
\end{figure}

Spectral surrogates reveal a subtler but consistent imprint. Under \textsc{Adam}, the distance from initialization in spectral norm drifts upward over time; on log time the slope is small but positive beyond the dashed line. Under momentum \textsc{SGD}, the same quantity gently decreases from a plateau, appearing as a mild negative slope. The ratio \texttt{FRO\_OVER\_SPEC} also separates in level, hinting that the optimizer reshapes how mass is distributed across singular directions even when predictive performance is unchanged.

\begin{figure}[t]
\centering
\subfigure[Adam]{%
  \includegraphics[width=.48\linewidth]{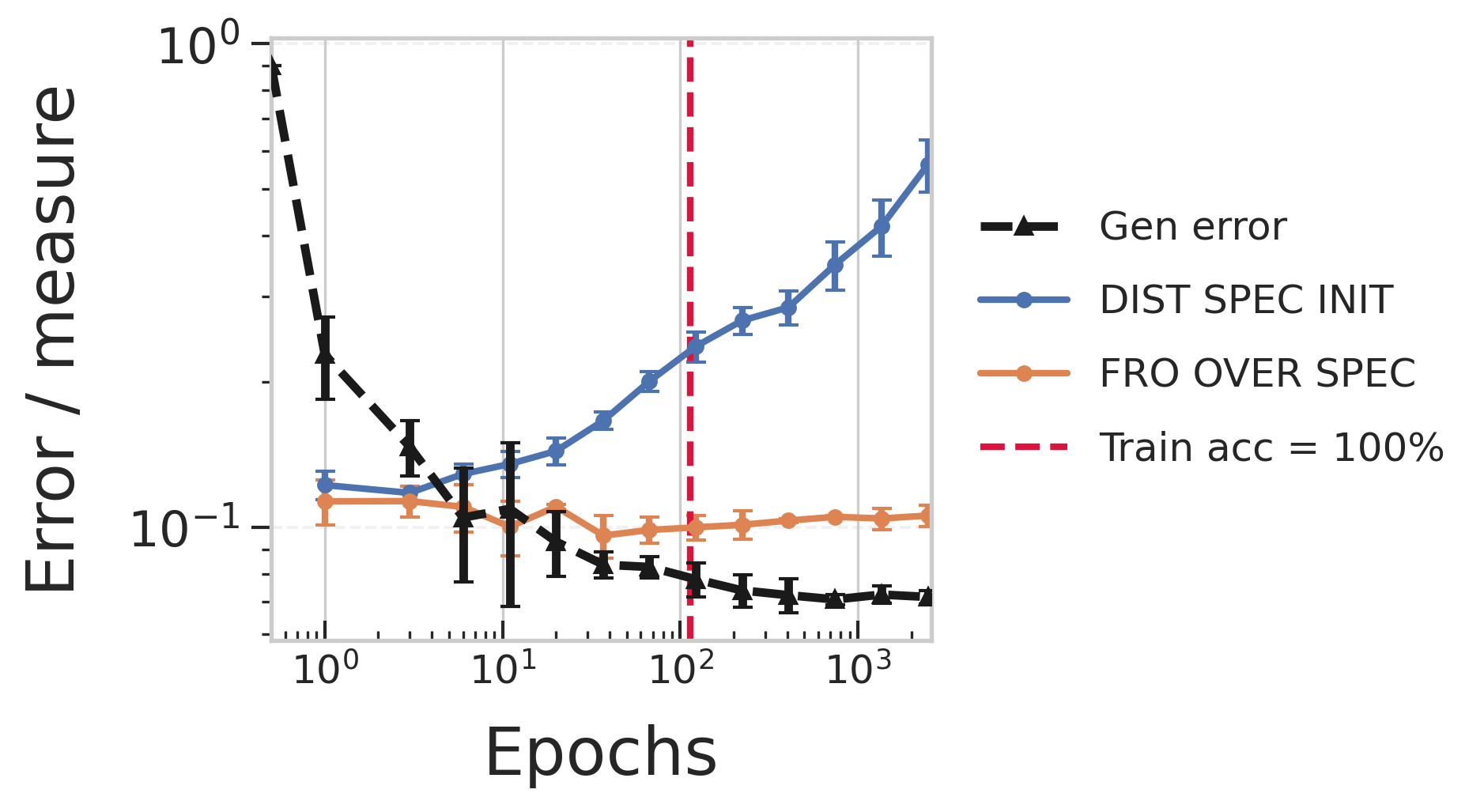}}
\hfill
\subfigure[\textsc{SGD} with momentum]{%
  \includegraphics[width=.48\linewidth]{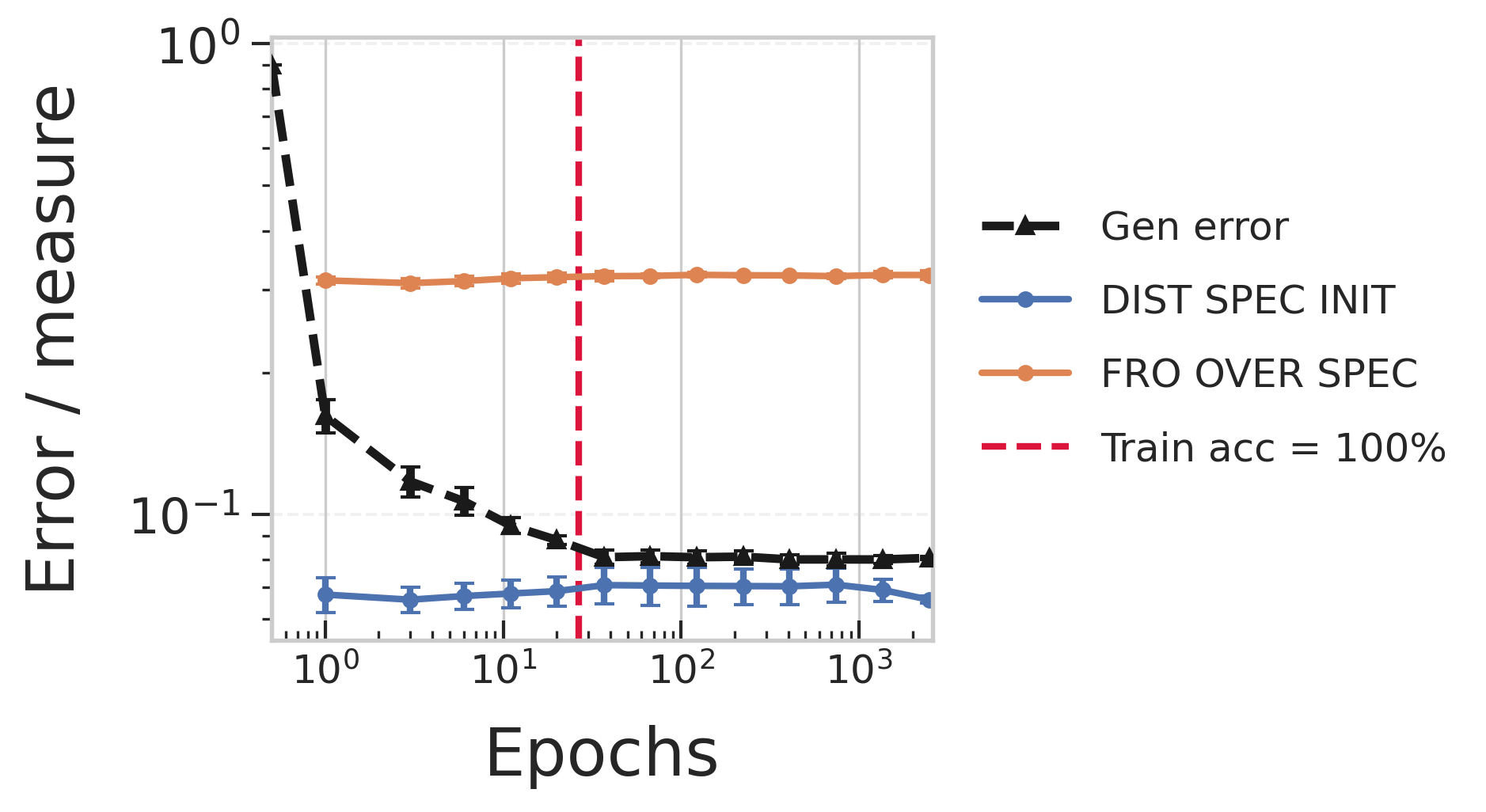}}
\caption{Spectral metrics through time. The epoch axis is logarithmic. \textsc{Adam} drives \texttt{DIST\_SPEC\_INIT} upward after interpolation, while momentum \textsc{SGD} yields a gentle decline; \texttt{FRO\_OVER\_SPEC} diverges slightly in level.}
\label{fig:temporal-spectral}
\end{figure}

As a neutral reference, the VC‑dimension proxy behaves identically across optimizers by construction, and the accompanying generalization error follows the same calm trajectory. The log-time axis makes this invariance explicit: the traces remain overlapped across decades of epochs, serving as anchors that remind us not all measures are sensitive to the optimizer change.

\begin{figure}[t]
\centering
\subfigure[Adam]{%
  \includegraphics[width=.48\linewidth]{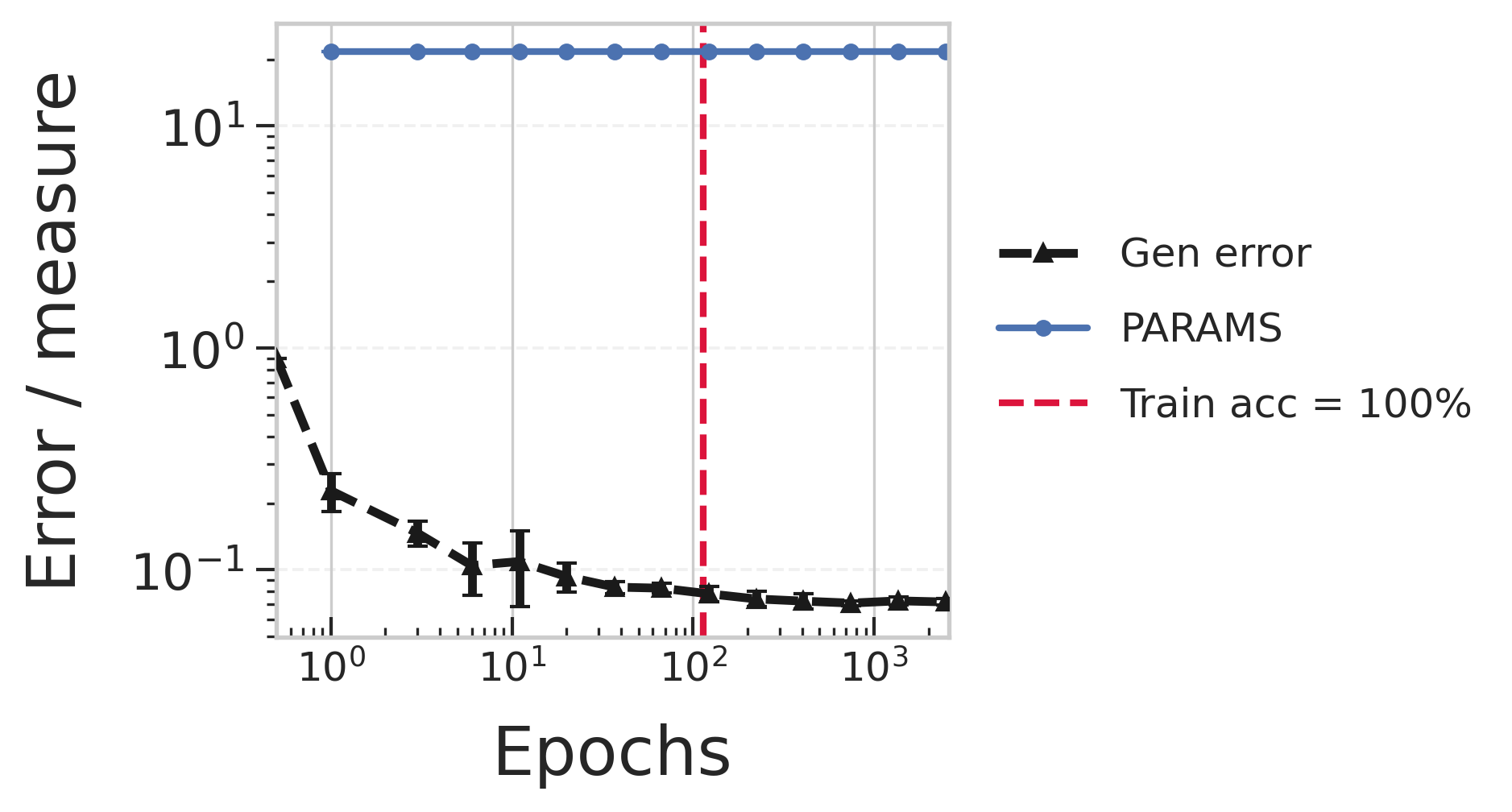}}
\hfill
\subfigure[\textsc{SGD} with momentum]{%
  \includegraphics[width=.48\linewidth]{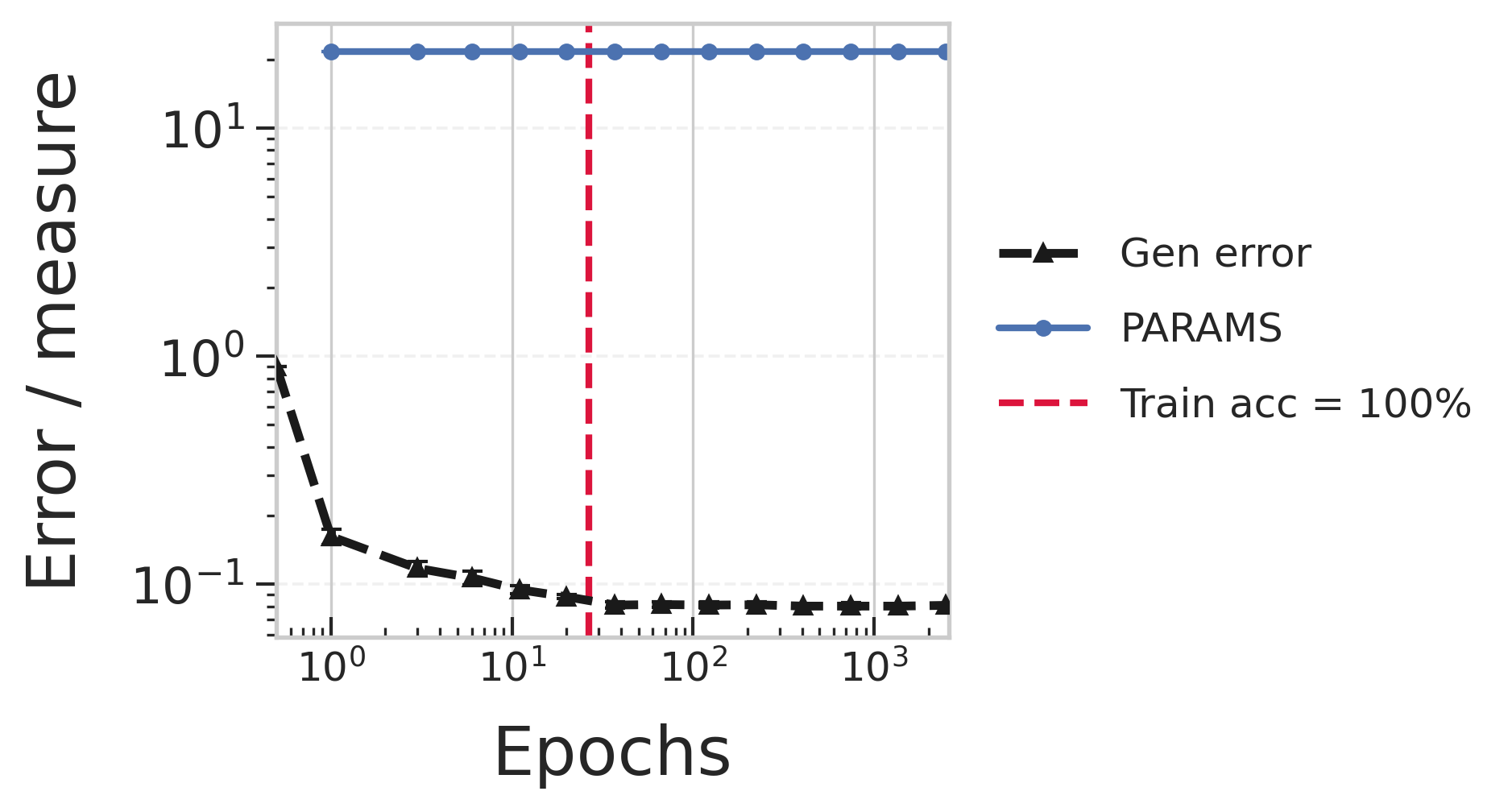}}
\caption{VC‑dimension proxy and generalization error through time on a logarithmic epoch axis. The parameter‑count surrogate is identical for both optimizers and the error trajectories are similarly calm, providing a neutral reference.}
\label{fig:temporal-vc}
\end{figure}

Taken together, these appendix figures broaden the temporal evidence. Several measure families that depend directly on weight scale or spectrum—Frobenius norms, spectral surrogates, and parts of the PAC--Bayes suite—react strongly to an optimizer swap despite matching accuracy, while others such as margin‑based quantities and the VC proxy remain largely stable. Reading these results in aggregate helps separate measure‑intrinsic behavior from optimizer‑driven drift and, with the log-time view, clarifies whether apparent motion reflects genuine multiplicative change or merely late-stage additive updates. For temporal behavior of path norms, see Section~\ref{sec:temporal-fragility}, where those panels are discussed in detail.

\section{Theory of scale-invariance: definitions and proofs}
\label{sec:proof_of_theorem_exp+wd}

In this section, we provide the formal definition of scale-invariant networks and the theorem regarding the equivalence of schedules (the "Exp++" factor) discussed in Section~\ref{sec:exp++lr}. We then provide the full proof.

\subsection{Definitions and Theorem Statement}

\begin{definition}[Scale invariant neural networks]
    Consider a parameterized neural network $f(\boldsymbol{\theta})$. 
    We say $f$ is \textit{scale invariant} if
    \begin{equation}
        \forall c  \in \mathbb{R}^{+}, f(\boldsymbol{\theta})=f(c \boldsymbol{\theta}) 
    \end{equation}
\end{definition}

\begin{theorem}[Equivalence of schedules in scale-invariant nets]
\label{theorem:exp+wd}
\emph{(GD+WD+LR fixed $\Leftrightarrow$ GD+WD$\searrow$ + LR$\nearrow$).}
Let $f(\theta_t)$ be scale-invariant and let training use \textsc{SGD} with momentum $\gamma$.
Introduce the shorthands
\begin{align}
\Delta_\lambda &\coloneqq (1-\gamma)^2 - 2(1+\gamma)\lambda\eta_0 + (\lambda\eta_0)^2,\\
\Xi(\alpha) &\coloneqq \frac{\alpha^2-\alpha(1-\lambda\eta_0+\gamma)+\gamma}{\eta_0}.
\end{align}
Define the interval endpoints
\begin{align}
\alpha_L &\coloneqq \frac{\gamma}{\,1-\lambda\eta_0+\gamma\,},\\
\alpha_- &\coloneqq \frac{1+\gamma-\lambda\eta_0-\sqrt{\Delta_\lambda}}{2},\\
\alpha_+ &\coloneqq \frac{1+\gamma-\lambda\eta_0+\sqrt{\Delta_\lambda}}{2},
\end{align}
and set
\begin{equation}
\mathcal{I}\coloneqq (\alpha_L,\ \alpha_-]\ \cup\ [\alpha_+,\,1).
\end{equation}

Consider the two updates (with $\theta_{-1}=\theta_0$, $\tilde\theta_0=\theta_0$, $\tilde\theta_{-1}=\alpha\,\theta_{-1}$):
\begin{align}
\tag{A}
\frac{\theta_t-\theta_{t-1}}{\eta_0}
&= \frac{\gamma(\theta_{t-1}-\theta_{t-2})}{\eta_0}
   - \nabla_\theta\!\Big(L(\theta_{t-1}) + \tfrac{\lambda}{2}\|\theta_{t-1}\|_2^2\Big),\\[2pt]
\tag{B}
\frac{\tilde\theta_t-\tilde\theta_{t-1}}{\tilde\eta_t}
&= \frac{\gamma(\tilde\theta_{t-1}-\tilde\theta_{t-2})}{\tilde\eta_{t-1}}
   - \nabla_\theta\!\Big(L(\tilde\theta_{t-1})
   + \tfrac{\tilde\lambda_t}{2}\|\tilde\theta_{t-1}\|_2^2\Big).
\end{align}

If $\alpha\in\mathcal{I}$, then the schedules
\begin{align}
\tilde\eta_t   &= \eta_0\,\alpha^{-2t-1},\\
\tilde\lambda_t&= \Xi(\alpha)\,\alpha^{2t-1}
                 + \frac{\gamma(1-\alpha)}{\eta_0\,\alpha}\,\mathbbm{1}\{t=0\}
\end{align}
ensure $\tilde\theta_t=\alpha^{-t}\theta_t$ for all $t\ge0$; hence (A) and (B) generate identical functions.
\end{theorem}

\begin{remark}[Parameter range and nonnegativity]
Assume
\begin{align}
\lambda\eta_0 &\le (1-\sqrt{\gamma})^2,\\
\alpha_L &< \alpha_- \quad\text{(equivalently, $\ \Delta_\lambda\ge 0$)}.
\end{align}
Then the interval $\mathcal{I}$ above is nonempty and the square‑root term is real; both conditions hold for common hyperparameters in practice.
\label{rmk:value_of_alpha}
\end{remark}

\subsection{Proof of Theorem~\ref{theorem:exp+wd}}

We first reintroduce the scale invariance lemma from \citet{arora2018theoretical,li2019an} which is the key source of intuition about scale-invariant networks.

\begin{lemma}[Scale-invariant networks]
    If $\forall c \in \mathbb{R}^{+}$, $L(\boldsymbol{\theta}) = L(c \boldsymbol{\theta}) $, then
    \begin{enumerate}
        \item $\left\langle\nabla_{\boldsymbol{\theta}} L, \boldsymbol{\theta}\right\rangle=0$
        \item $\left.\nabla_{\boldsymbol{\theta}} L\right|_{\boldsymbol{\theta}=\boldsymbol{\theta}_0}=\left.c \nabla_{\boldsymbol{\theta}} L\right|_{\boldsymbol{\theta}=c \boldsymbol{\theta}_0}$, for any $c>0$
    \end{enumerate}
\label{lemma:scale-invariance}
\end{lemma}

\begin{proof}
Treat $c$ as a differentiable variable. Clearly the derivative of $L$ w.r.t. $c$ is $0$.
    \begin{enumerate}
    \item   
    $ 0 = \frac{\partial L}{\partial c} =   \left\langle\nabla_{\boldsymbol{\theta}} L, \boldsymbol{\theta}\right\rangle$
    \item  Take gradient of the both sides of the equation $L(\boldsymbol{\theta}) = L(c \boldsymbol{\theta})$ w.r.t. $\boldsymbol{\theta}$
    and set
    $\boldsymbol{\theta}= \boldsymbol{\theta}_0$, we immediately arrive at the result.
    \end{enumerate}
\end{proof}

\noindent
We give the technical definition of the commonly used 
training algorithm \textsc{SGD} with momentum and weight decay (WD) (with respect to the L2 norm of the parameters) using a convenient 
form given in \citet{li2019an}, which is equivalent to the default implementation in Pytorch \citep{paszke2017automatic}.

\begin{definition}[\textsc{SGD} with momentum and WD]
    At iteration $t$, given the current parameters and learning rate $(\boldsymbol{\theta}_{t-1}$, $\eta_{t-1})$,
    the buffered parameters and learning rate $(\boldsymbol{\theta}_{t-2}$, $\eta_{t-2})$,
     momentum factor $\gamma$, current loss $L(\boldsymbol{\theta}_{t-1})$ and WD factor $\lambda_{t-1}$,
    update the parameters $\boldsymbol{\theta}_{t}$ as the following:
    \begin{equation}
        \frac{\boldsymbol{\theta}_t-\boldsymbol{\theta}_{t-1}}{\eta_{t-1}} = \gamma \frac{\boldsymbol{\theta}_{t-1}-\boldsymbol{\theta}_{t-2}}{\eta_{t-2}}-\nabla_{\boldsymbol{\theta}}\left(L\left(\boldsymbol{\theta}_{t-1}\right)+\frac{\lambda_{t-1}}{2}\left\|\boldsymbol{\theta}_{t-1}\right\|_2^2\right)
    \end{equation}
    \\
    For the boundary conditions, it is a common practice to set $\boldsymbol{\theta}_{-1} = \boldsymbol{\theta}_{0}$ and $\eta_{-1}$ 
    can be arbitrary. 

\end{definition}

From the above definition, it is easy to see that we can
represent the state of the algorithm 
using a four-tuple $\left(\boldsymbol{\theta}, \eta, \boldsymbol{\theta^{\prime}}, \eta^{\prime}\right)$, which stand for 
the parameters/learning rate at the current step and their buffers from the last step, respectively. 
A gradient descent step at time $t$
with momentum factor $\gamma$ and WD factor $\lambda$
can be seen as a mapping between two states:

\begin{itemize}
    \item A GD step with  momentum and WD: $ \> \operatorname{GD}_t^\rho\left(\boldsymbol{\theta}, \eta, \boldsymbol{\theta}^{\prime}, \eta^{\prime}\right)=\left(\rho \boldsymbol{\theta}+\eta\left(\gamma \frac{\boldsymbol{\theta}-\boldsymbol{\theta}^{\prime}}{\eta^{\prime}}-\nabla L(\boldsymbol{\theta})\right), \eta, \boldsymbol{\theta}, \eta\right)$
\end{itemize}
Here with WD factor being $\lambda$, $\rho$ is set to be $1-\lambda \eta$.
Furthermore, we define some extra basic mappings that can be composed together to represent the temporal behavior of the algorithm.

\begin{itemize}
    \item Scale current parameters $\boldsymbol{\theta}$: \tabto{4.9cm}
    $ \Pi_1^c\left(\boldsymbol{\theta}, \eta, \boldsymbol{\theta}^{\prime}, \eta^{\prime}\right)=\left(c \boldsymbol{\theta}, \eta, \boldsymbol{\theta}^{\prime}, \eta^{\prime}\right)$; 
    \item Scale current LR $\eta$: \tabto{4.9cm}
    $ \Pi_2^c\left(\boldsymbol{\theta}, \eta, \boldsymbol{\theta}^{\prime}, \eta^{\prime}\right)=\left(\boldsymbol{\theta}, c \eta, \boldsymbol{\theta}^{\prime}, \eta^{\prime}\right)$;
    \item Scale buffered parameter $\boldsymbol{\theta^{\prime}}$: \tabto{4.9cm}
     $\Pi_3^c\left(\boldsymbol{\theta}, \eta, \boldsymbol{\theta}^{\prime}, \eta^{\prime}\right)=\left(\boldsymbol{\theta}, \eta, c \boldsymbol{\theta}^{\prime}, \eta^{\prime}\right)$;
    \item  Scale buffered LR $\eta^{\prime}$: \tabto{4.9cm}
    $ \Pi_4^c\left(\boldsymbol{\theta}, \eta, \boldsymbol{\theta}^{\prime}, \eta^{\prime}\right)=\left(\boldsymbol{\theta}, \eta, \boldsymbol{\theta}^{\prime}, c \eta^{\prime}\right)$.
 
\end{itemize}

We know that in scale-invariant neural nets, two networks $f(\boldsymbol{\theta})$ and $f(\widetilde{\boldsymbol{\theta}})$
are equivalent if $\exists c > 0$ such that $ \widetilde{\boldsymbol{\theta}}=c\boldsymbol{\theta}$; Here we extend the equivalence
between weights to the equivalence between states of algorithms: 

\begin{definition}[Equivalent states]
    Two states $\left(\boldsymbol{\theta}, \eta,  \boldsymbol{\theta}^{\prime}, \eta^{\prime}\right)$ and $(\widetilde{\boldsymbol{\theta}}, \widetilde{\eta}, \widetilde{\boldsymbol{\theta}^{\prime}}, \widetilde{\eta^{\prime}})$
    are \textit{equivalent} iff $\exists c > 0 $ such that 
    $\left(\boldsymbol{\theta}, \eta, \boldsymbol{\theta}^{\prime}, \eta^{\prime}\right)=\left[\Pi_1^c \circ \Pi_2^{c^2} \circ \Pi_3^c \circ \Pi_4^{c^2}\right]\left(\widetilde{\boldsymbol{\theta}}, \widetilde{\eta}, \widetilde{\boldsymbol{\theta}^{\prime}}, \widetilde{\eta^{\prime}}\right)=\left(c \widetilde{\boldsymbol{\theta}}, c^2 \widetilde{\eta}, c \widetilde{\boldsymbol{\theta}^{\prime}}, c^2 \widetilde{\eta^{\prime}}\right)$,
    which is also noted as $\left(\boldsymbol{\theta}, \eta, \boldsymbol{\theta}^{\prime}, \eta^{\prime}\right) \stackrel{c}{\sim}(\widetilde{\boldsymbol{\theta}}, \widetilde{\eta}, \widetilde{\boldsymbol{\theta}^{\prime}}, \widetilde{\eta^{\prime}})$.
    We call $\left[\Pi_1^c \circ \Pi_2^{c^2} \circ \Pi_3^c \circ \Pi_4^{c^2}\right]$ as \textit{equivalent scaling} for all $c > 0$.
    
\end{definition}

Here we provide an intuitive explanation of why the equivalent scaling takes the form above.
If we
rearrange the first term of the R.H.S. of the GD update, and assume we 
are operating in a regime where $\eta^\prime = \eta$ \footnote{This just means in the eyes of the GD algorithm, the buffered LR and the current LR are the same. It does not exclude the possibility that we can still scale the current LR between GD updates. }, we have
\begin{equation}
    \boldsymbol{\theta}_{\text{update}} = (\rho + \gamma) \boldsymbol{\theta} - \eta \nabla L(\boldsymbol{\theta}) -
    \gamma \boldsymbol{\theta}^{\prime}
\label{equ:SGD_update_inituition}
\end{equation}

In order to keep the updated parameters in the same direction, the three terms in equation \ref{equ:SGD_update_inituition} need to have the same
scaling. From lemma \ref{lemma:scale-invariance} we know that when $\boldsymbol{\theta}$ is scaled by $c$, the gradient 
$ \nabla L(\boldsymbol{\theta})$ will actually be scaled by $\frac{1}{c}$. Hence for the second term $\eta \nabla L(\boldsymbol{\theta})$ 
to have the same amount of scaling as the first and third terms, $\eta$ has to be scaled by $c^2$.

The following lemma tells us that equivalent scaling commutes with GD update with momentum and WD, implying
that equivalence is preserved under GD updates. Hence we are free to stack GD updates and insert equivalent scaling anywhere in a sequence of basic maps 
without changing the network function.

\begin{lemma}[Equivalent scaling commutes with GD]
$\forall c, \rho>0$ 
and 
$t \geq 0$,
$$\text{GD}_t^\rho \circ\left[\Pi_1^c \circ \Pi_2^{c^2} \circ \Pi_3^c \circ \Pi_4^{c^2}\right]=\left[\Pi_1^c \circ \Pi_2^{c^2} \circ \Pi_3^c \circ \Pi_4^{c^2}\right] \circ \text{GD}_t^\rho.$$    
In other words, if $\left(\boldsymbol{\theta}, \eta, \boldsymbol{\theta}^{\prime}, \eta^{\prime}\right) \stackrel{c}{\sim}(\widetilde{\boldsymbol{\theta}}, \widetilde{\eta}, \widetilde{\boldsymbol{\theta}^{\prime}}, \widetilde{\eta^{\prime}})$ then 
$\text{GD}_t^\rho \left(\boldsymbol{\theta}, \eta, \boldsymbol{\theta}^{\prime}, \eta^{\prime}\right) \stackrel{c}{\sim} 
\text{GD}_t^\rho (\widetilde{\boldsymbol{\theta}}, \widetilde{\eta}, \widetilde{\boldsymbol{\theta}^{\prime}}, \widetilde{\eta^{\prime}})
$.
\label{lemma:commute}
\end{lemma}

\begin{proof}
    For any given state $(\boldsymbol{\theta}, \eta,  \boldsymbol{\theta}^{\prime}, \eta^{\prime})$, the L.H.S. of the equation is: 
    $$
    \begin{aligned}
         \text{GD}_t^\rho \circ\left[\Pi_1^c \circ \Pi_2^{c^2} \circ \Pi_3^c \circ \Pi_4^{c^2}\right](\boldsymbol{\theta}, \eta,  \boldsymbol{\theta}^{\prime}, \eta^{\prime})
        &=  \text{GD}_t^\rho (c \boldsymbol{\theta}, c^2 \eta, c \boldsymbol{\theta}^{\prime}, c^2 \eta^{\prime}) \\
        &=  \left(
        \rho c \boldsymbol{\theta} + c^2 \eta \left( \gamma \frac{c\boldsymbol{\theta} - c\boldsymbol{\theta^{\prime}}}{c^2\eta^{\prime}}
        - \nabla L(c\boldsymbol{\theta})
        \right), 
        c^2\eta,
        c\boldsymbol{\theta},
        c^2\eta
        \right) \\
        &= \left( c \left( \rho \boldsymbol{\theta} + \eta \left( \gamma \frac{\boldsymbol{\theta} - \boldsymbol{\theta^{\prime}}}{\eta^\prime} -\nabla L(\boldsymbol{\theta}) \right) 
        \right),
        c^2 \eta,
        c\boldsymbol{\theta},
        c^2\eta
        \right) \\
        &= \left[\Pi_1^c \circ \Pi_2^{c^2} \circ \Pi_3^c \circ \Pi_4^{c^2}\right] \circ \text{GD}_t^\rho (\boldsymbol{\theta}, \eta,  \boldsymbol{\theta}^{\prime}, \eta^{\prime})
    \end{aligned}
    $$
\end{proof}

Now comes the important step: in order to show the equivalence between two series of parameters with 
(fixed WD +  fixed LR)/(exponentially decreasing WD + exponentially increasing LR), respectively, we need to rewrite $\text{GD}_t^\rho$
as a composition of itself with varying WD factor and upscaling LR, conjugated with other scaling terms that cancel with each other eventually.
Here again we choose to work in the regime where the current and buffered LR are the same in the input of $\text{GD}_t^\rho$.

\begin{lemma}[Conjugated GD updates]
    For any input
    with equal current and buffered LR
    $\left(\boldsymbol{\theta}, \eta,  \boldsymbol{\theta}^{\prime}, \eta\right)$ and 
    $\forall \alpha \in  \left( Z_0, Z_1 \right] \cup \left[ Z_2, 1\right) $ \footnote{Technically $\alpha$ can be larger than $1$, but in that case we will be shrinking the LR between GD steps which is not what we mainly care about here.}
    , 
    we have
    $$
\text{GD}_t^\rho\left(\boldsymbol{\theta}, \eta, \boldsymbol{\theta}^{\prime}, \eta\right)=\left[\Pi_4^\alpha \circ \Pi_2^\alpha \circ \Pi_1^\alpha \circ \text{GD}^{\beta}_t \circ \Pi_2^{\alpha^{-1}} \circ \Pi_3^\alpha \circ \Pi_4^\alpha\right]\left(\boldsymbol{\theta}, \eta, \boldsymbol{\theta}^{\prime}, \eta\right)
$$
which can be written in the form of equivalent states:
\begin{equation}
\text{GD}_t^\rho\left(\boldsymbol{\theta}, \eta, \boldsymbol{\theta}^{\prime}, \eta\right) \stackrel{\alpha}{\sim}\left[\Pi_3^{\alpha^{-1}} \circ \Pi_4^{\alpha^{-1}} \circ \Pi_2^{\alpha^{-1}} \circ \text{GD}_t^\beta \circ \Pi_2^{\alpha^{-1}} \circ \Pi_3^\alpha \circ \Pi_4^\alpha\right]\left(\boldsymbol{\theta}, \eta, \boldsymbol{\theta}^{\prime}, \eta\right)
\label{equ:conjugated_GD_updates}
\end{equation}
where
\begin{itemize}
    \item $\beta = \frac{(\rho + \gamma)}{\alpha} - \frac{\gamma}{\alpha^2}$
    \item $Z_0 = \frac{\gamma}{1-\lambda \eta_0 + \gamma}$
    \item $Z_1 = \frac{1+\gamma-\lambda \eta_0-\sqrt{(1-\gamma)^2-2(1+\gamma) \lambda \eta_0+\lambda^2 \eta_0^2}}{2}$
    \item $Z_2 = \frac{1+\gamma-\lambda \eta_0+\sqrt{(1-\gamma)^2-2(1+\gamma) \lambda \eta_0+\lambda^2 \eta_0^2}}{2}$ 
    \footnote{It's easy to verify that $Z_2$ is always smaller than 1.}
\end{itemize}
\end{lemma}

\begin{proof}
    We directly verify the equivalence. The R.H.S. is:
    $$
    \begin{aligned}
        &\left[\Pi_4^\alpha \circ \Pi_2^\alpha \circ \Pi_1^\alpha \circ \text{GD}^{\beta}_t \circ \Pi_2^{\alpha^{-1}} \circ \Pi_3^\alpha \circ \Pi_4^\alpha\right]\left(\boldsymbol{\theta}, \eta, \boldsymbol{\theta}^{\prime}, \eta\right) \\
    = &\left[\Pi_4^\alpha \circ \Pi_2^\alpha \circ \Pi_1^\alpha\right] \circ \text{GD}^{\beta}_t \left( \boldsymbol{\theta}, \alpha^{-1} \eta, \alpha\boldsymbol{\theta}^{\prime}, \alpha\eta   \right) \\
     = &\left[\Pi_4^\alpha \circ \Pi_2^\alpha \circ \Pi_1^\alpha \right] \left( 
     \beta \boldsymbol{\theta} + \alpha^{-1} \eta \left( \gamma \frac{\boldsymbol{\theta} - \alpha\boldsymbol{\theta}^{\prime}}{\alpha\eta}  
     - \nabla L(\boldsymbol{\theta})
     \right), \alpha^{-1}\eta, \boldsymbol{\theta}, \alpha^{-1}\eta
     \right) \\
     = &\left(    
     \alpha \beta \boldsymbol{\theta} + \eta \left( \gamma   \frac{\boldsymbol{\theta} - \alpha\boldsymbol{\theta}^{\prime}}{\alpha\eta} - \nabla L(\boldsymbol{\theta}) \right), \eta, \boldsymbol{\theta}, \eta
     \right) \\
     = &\left(
     \left(\rho+\gamma\right) \boldsymbol{\theta} - \gamma \boldsymbol{\theta}^{\prime} - \eta\nabla L(\boldsymbol{\theta}),
     \eta, \boldsymbol{\theta}, \eta
     \right) \\
     = &\left(
     \rho\boldsymbol{\theta} + 
        \eta \left(\gamma \frac{\boldsymbol{\theta}-\boldsymbol{\theta}^{\prime}}{\eta} - \nabla L(\boldsymbol{\theta}) 
        \right),
        \eta, \boldsymbol{\theta}, \eta
     \right) \\
     = &\text{GD}_t^\rho\left(\boldsymbol{\theta}, \eta, \boldsymbol{\theta}^{\prime}, \eta\right)
    \end{aligned}
    $$
    The range of $\alpha$ can be easily shown by combining the following two constraints and assuming remark \ref{rmk:value_of_alpha} is true:
    \begin{itemize}
        \item $\alpha \in \left(0, 1\right];$
        \item $\frac{(\rho + \gamma)}{\alpha} - \frac{\gamma}{\alpha^2} \in \left(0, 1\right]$
    \end{itemize}
\end{proof}

Now we are ready to prove theorem~\ref{theorem:exp+wd}.
\begin{proof}[Proof of Theorem~\ref{theorem:exp+wd}]
    From the assumption we have the following equivalence between the boundary conditions of the two series of states:
    $$
    \begin{aligned}
        \left( \boldsymbol{\theta}_0, \eta_0, \boldsymbol{\theta}_{-1}, \eta_{-1} \right) &=
        \left( \boldsymbol{\theta}_0, \eta_0, \boldsymbol{\theta}_{0}, \eta_{0} \right)
        \\
        \left( \widetilde{\boldsymbol{\theta}}_0, \widetilde{\eta}_0, \widetilde{\boldsymbol{\theta}}_{-1}, \widetilde{\eta}_{-1} \right) &= 
    \left[ \Pi_2^{\alpha^{-1}} \circ \Pi_3^\alpha \circ \Pi_4^\alpha \right] 
    \left( \boldsymbol{\theta}_0, \eta_0, \boldsymbol{\theta}_{-1}, \eta_{-1} \right)
    \end{aligned}
    $$
    Lemma \ref{lemma:commute} tells us that equivalent states are still equivalent after both being transformed by a GD step. 
    Hence we can stack up on both sides of equation \ref{equ:conjugated_GD_updates} for a finite number of times. i.e. for $\forall t \ge 0$, we have 
    $$
    \begin{aligned}
        &\text{GD}^{\rho}_{t-1} \circ \text{GD}^{\rho}_{t-2} \circ \cdot \cdot \cdot \circ \text{GD}^{\rho}_{0} 
        \left(  
        \boldsymbol{\theta}_0, \eta_0, \boldsymbol{\theta}_{-1}, \eta_{-1}
        \right)  \\
         \stackrel{\alpha^t}{\sim} 
        & \left[ 
        \Pi_3^{\alpha^{-1}} \circ \Pi_4^{\alpha^{-1}} \circ \Pi_2^{\alpha^{-1}} \circ \text{GD}^{\beta}_{t-1} \circ \Pi_2^{\alpha^{-1}} \circ
        \Pi_3^{\alpha} \circ \Pi_4^{\alpha} 
        \right]  \\
         &\circ \cdot \cdot \cdot \circ
        \left[ 
        \Pi_3^{\alpha^{-1}} \circ \Pi_4^{\alpha^{-1}} \circ \Pi_2^{\alpha^{-1}}\right] \circ \text{GD}^{\beta}_{0}
        \left( \widetilde{\boldsymbol{\theta}}_0, \widetilde{\eta}_0, \widetilde{\boldsymbol{\theta}}_{-1}, \widetilde{\eta}_{-1} \right) \\
         \stackrel{\alpha^t}{\sim} 
        &\left[ 
        \Pi_3^{\alpha^{-1}} \circ \Pi_4^{\alpha^{-1}} \circ \Pi_2^{\alpha^{-1}} \circ \text{GD}^{\beta}_{t-1} \circ \Pi_2^{\alpha^{-2}} \circ
          \text{GD}^{\beta}_{t-2} \circ \cdot \cdot \cdot \circ  \Pi_2^{\alpha^{-2}} \circ\text{GD}^{\beta}_{0}
          \right] 
          \left( \widetilde{\boldsymbol{\theta}}_0, \widetilde{\eta}_0, \widetilde{\boldsymbol{\theta}}_{-1}, \widetilde{\eta}_{-1} \right)
    \end{aligned}
    $$
    which implies that 
    \begin{itemize}
        \item  $\boldsymbol{\theta}_t = \alpha^t \widetilde{\boldsymbol{\theta}}_{t}$
        \item $\widetilde{\eta}_{t} = \alpha ^{-2t} \widetilde{\eta}_{0} =  \alpha ^{-2t-1} \eta_0$
        \item $ \widetilde{\lambda}_{t} = \frac{1-\beta}{\widetilde{\eta}_{t}} $, except for $t=0$, in which case 
        $\widetilde{\lambda}_{0} = \frac{1-\beta + \frac{\gamma}{\alpha^2} - \frac{\gamma}{\alpha}}{\widetilde{\eta}_{0}}$
    \end{itemize}
    We note the special boundary condition for $\widetilde{\lambda}_{0}$ due to the fact that $\widetilde{\boldsymbol{\theta}}_{-1} \ne \widetilde{\boldsymbol{\theta}}_{0}$.
\end{proof}


\section{Additional label-corruption results: PAC–Bayes and Path Norms}
\label{app:label-corruption}

\noindent This appendix gathers the label–corruption results referenced in the main text for both PAC–Bayes–style measures and Path Norms.

\paragraph{PAC–Bayes measures.}
Under \textsc{Adam}, lowering the learning rate from $10^{-2}$ to $10^{-3}$ transforms a clean, steadily rising ramp (e.g., \texttt{PACBAYES\_ORIG} from $\sim 3.6$ past $4.7$) into a higher–lying but \emph{shallower} U–shape centred around $\sim 8$. 
At fixed learning rate $10^{-2}$, swapping \textsc{Adam} for \textsc{SGD} produces a striking level shift: \textsc{SGD} yields a \emph{high plateau} ($\sim 12$–$13$) with little curvature, while \textsc{Adam} traces a \emph{low}, clearly increasing arc ($\sim 3.6$–$4.7$). 
Even within the PAC–Bayes family, members respond differently by optimizer: \texttt{PACBAYES\_INIT} grows aggressively under \textsc{Adam} but only mildly under \textsc{SGD}. 
These patterns illustrate an (often) \emph{insensitive} or contradictory relationship between the measure and increasing label corruption—another facet of fragility.

\begin{figure}[t]
\centering
\subfigure[Adam, LR $=0.01$]{%
  \includegraphics[width=.48\columnwidth]{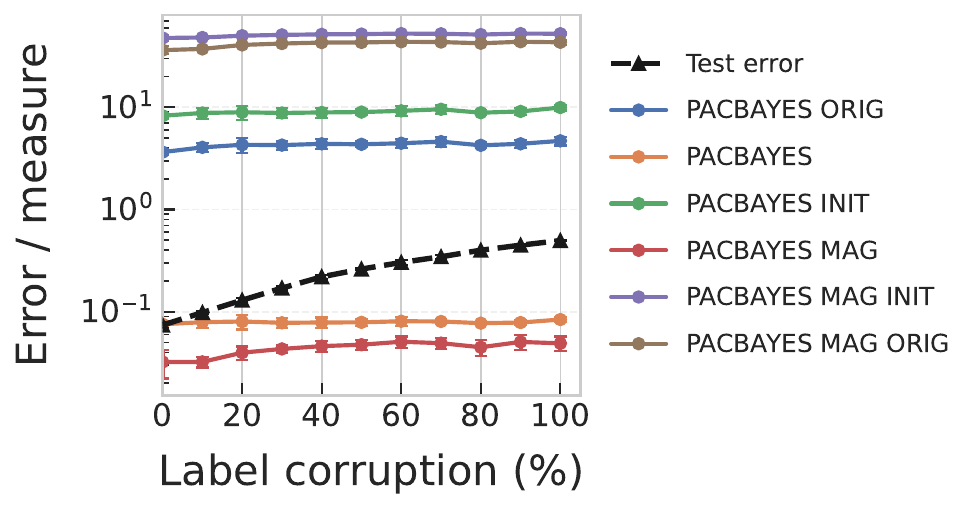}}
\subfigure[Adam, LR $=0.001$]{%
  \includegraphics[width=.48\columnwidth]{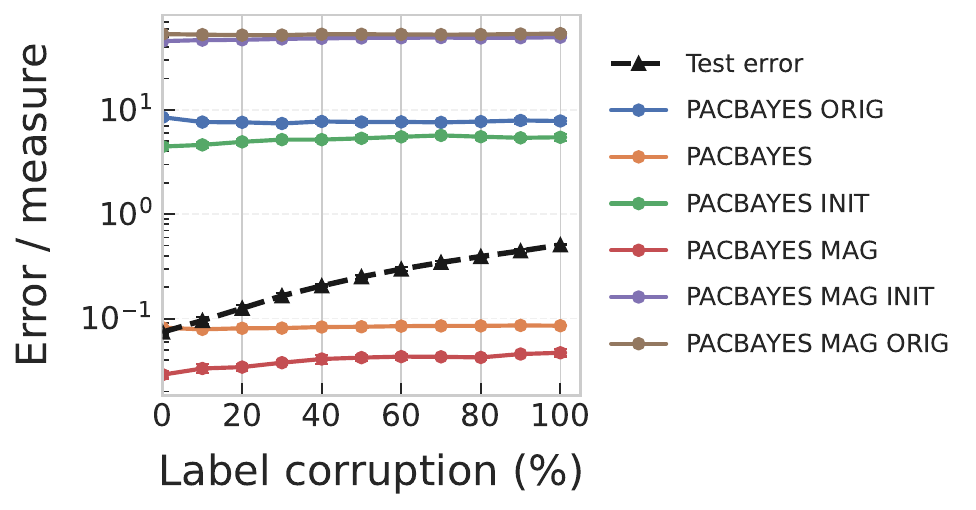}}
\caption{\textbf{PAC–Bayes measures vs.\ label corruption under Adam.} 
With $\mathrm{LR}{=}10^{-2}$, the family rises steadily with corruption; with $\mathrm{LR}{=}10^{-3}$ it sits higher overall and shows a shallow U–shape. 
\emph{All panels use 10{,}000 training samples.}}
\label{fig:pacfrag-adamlr}
\end{figure}

\begin{figure}[t]
\centering
\subfigure[Adam, LR $=0.01$]{%
  \includegraphics[width=.48\columnwidth]{pics_pdf/label_corruption_Oct2025/figures/RESNET50-FashionMNIST_binary/optimizer_ADAM__stop_false__lr_0.01/optimizer_ADAM__stop_false__lr_0.01__PAC-Bayes.pdf}}
\subfigure[SGD (momentum), LR $=0.01$]{%
  \includegraphics[width=.48\columnwidth]{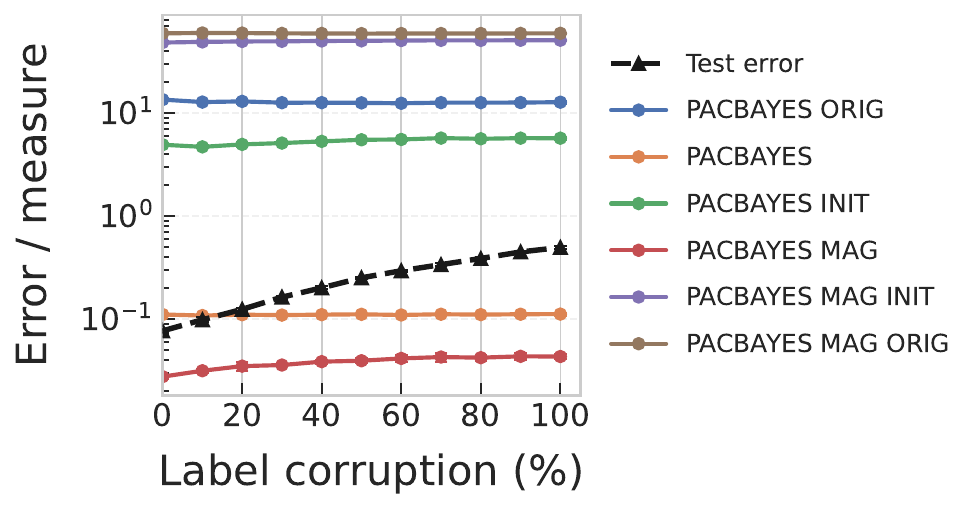}}
\caption{\textbf{PAC–Bayes measures vs.\ label corruption at fixed LR ($10^{-2}$), swapping only the optimizer.} 
Adam yields a low, steadily rising family (e.g., \texttt{PACBAYES\_ORIG} $\approx 3.6$–$4.7$), whereas SGD holds a high plateau ($\sim 12$–$13$) with minimal curvature; \texttt{PACBAYES\_INIT} grows far more under \textsc{Adam} than under SGD. 
\emph{All panels use 10{,}000 training samples.}}
\label{fig:pacfrag-opt}
\end{figure}

\begin{figure*}[t]
\centering
\subfigure[LR $=0.1$]{%
  \includegraphics[width=.48\linewidth]{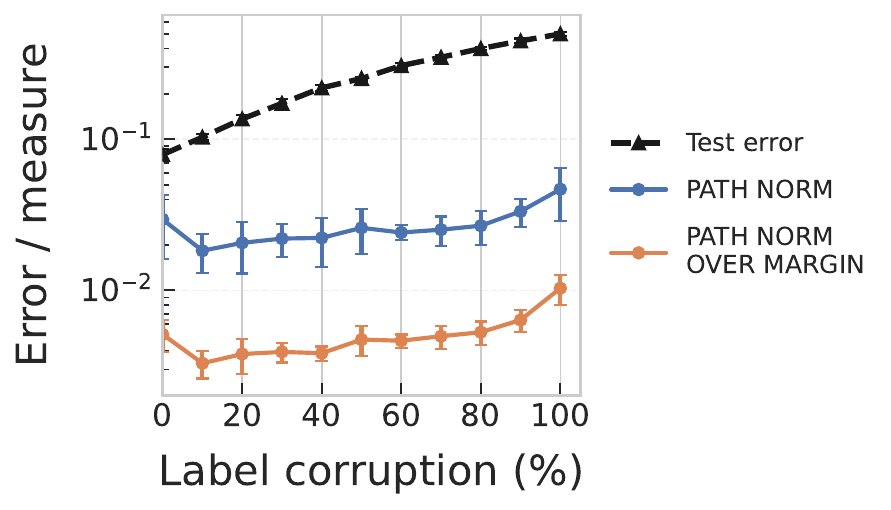}}
\subfigure[LR $=0.001$]{%
  \includegraphics[width=.48\linewidth]{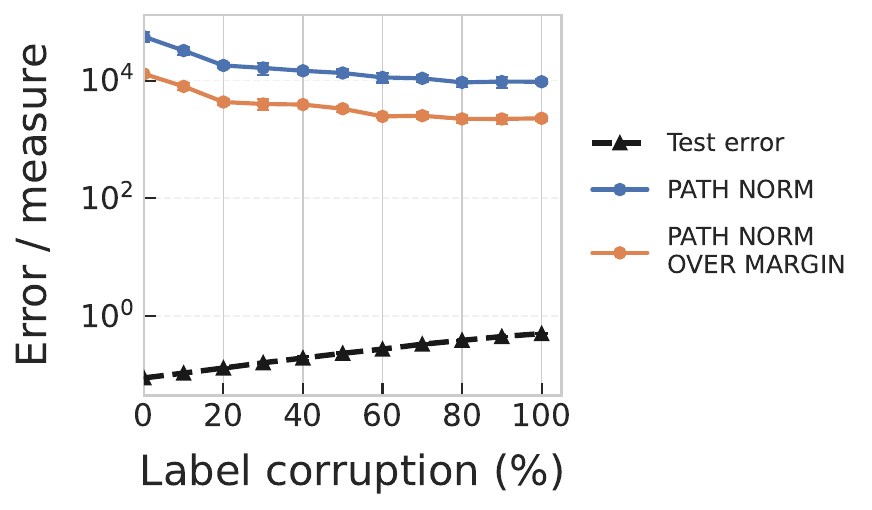}}
\caption{\textbf{Path norms vs.\ label corruption} with \emph{SGD (momentum)}; panels differ only in learning rate. 
At $\mathrm{LR}{=}10^{-3}$ the curve sits in $10^{4}$–$10^{5}$ and decays with corruption; at $\mathrm{LR}{=}10^{-1}$ it lives near zero and rises. 
Both the direction and the dynamic range flip—another instance of qualitative mismatch across a minimal training change. 
\emph{All panels use 10{,}000 training samples.}}
\label{fig:pathfrag-sgd}
\end{figure*}

\paragraph{Path norms under SGD (momentum).}
Mirroring the Adam case in the main text, \textsc{SGD} with momentum exhibits an equally striking flip (Fig.~\ref{fig:pathfrag-sgd}): at $\mathrm{LR}{=}10^{-3}$, the path-norm trajectory lives in the $10^{4}$–$10^{5}$ band and \emph{decays} steadily with corruption; at $\mathrm{LR}{=}10^{-1}$, the scale \emph{crashes} to nearly zero and the curve \emph{climbs} monotonically. 
Trend and scale both invert. 
If a reader tried to infer “harder data $\Rightarrow$ larger path norm” from one panel, the other would immediately contradict it.

\section{Quantitative fragility: detailed definitions and results}
\label{app:quant_fragility}

This appendix provides the formal definitions, algorithmic details, and full results for the quantitative analysis summarized in Section~\ref{sec:quantitative_fragility}.

\subsection{Formal Definitions}

To quantify fragility, we define metrics that capture how much a generalization measure fluctuates among runs with nearly identical test error.

\paragraph{Setup.}
Fix a dataset/architecture \emph{group} $g$ with trained runs $\mathcal{S}_g$.
For run $r\in\mathcal{S}_g$, let $\widehat{\varepsilon}_r\in[0,1]$ denote test error and
$C_r\in(0,\infty)$ the value of a post‑mortem measure $C$ (e.g., a norm proxy).
For an absolute error tolerance $\delta>0$, define the close‑error set
\begin{equation}
\label{eq:pi-delta}
\Pi_\delta(\mathcal{S}_g)
\;=\;
\bigl\{(r,s)\in\mathcal{S}_g^2:\ r<s,\ 
\lvert \widehat{\varepsilon}_r-\widehat{\varepsilon}_s\rvert\le \delta\bigr\}.
\end{equation}

\begin{definition}[Conditional Measure Spread (CMS)]
\label{def:cms}
The \emph{CMS} of $C$ in group $g$ at tolerance $\delta$ is
\begin{equation}
\label{eq:cms-def}
\mathrm{CMS}_\delta(C;g)
\;=\;
\operatorname{median}_{(r,s)\in \Pi_\delta(\mathcal{S}_g)}
\bigl\lvert \log C_r - \log C_s \bigr\rvert.
\end{equation}
\end{definition}

\noindent \textit{Interpretation.} $e^{\mathrm{CMS}_\delta}$ is the typical multiplicative factor by which $C$ varies across runs that are indistinguishable in test error up to $\delta$ (e.g., $\mathrm{CMS}_\delta\!\approx\!\ln 2$ means a factor‑two swing).

\paragraph{Randomness adjustment.}
To separate hyperparameter induced jumpiness from random‑seed jitter, let $H_r$ collect the
training hyperparameters for run $r$ (e.g., optimizer, learning rate, schedule flags,
early‑stopping rule, training‑set size), and let $\textsf{seed}_r$ be its random seed.
Split close‑error pairs into
\begin{align}
\label{eq:pi-seed}
\Pi_\delta^{\text{seed}}(g)
&=\bigl\{(r,s)\in \Pi_\delta(\mathcal{S}_g):\
H_r=H_s,\ \textsf{seed}_r\neq \textsf{seed}_s\bigr\},\\[-0.2em]
\label{eq:pi-inter}
\Pi_\delta^{\text{inter}}(g)
&=\bigl\{(r,s)\in \Pi_\delta(\mathcal{S}_g):\
H_r\neq H_s\bigr\}.
\end{align}
Define the corresponding spreads
\begin{align}
\label{eq:cms-seed}
\mathrm{CMS}^{\text{seed}}_\delta(C;g)
&=\operatorname{median}_{(r,s)\in \Pi_\delta^{\text{seed}}(g)}
\bigl\lvert \log C_r - \log C_s \bigr\rvert,\\[-0.2em]
\label{eq:cms-inter}
\mathrm{CMS}^{\text{inter}}_\delta(C;g)
&=\operatorname{median}_{(r,s)\in \Pi_\delta^{\text{inter}}(g)}
\bigl\lvert \log C_r - \log C_s \bigr\rvert.
\end{align}

\begin{definition}[eCMS (Excess CMS)]
\label{def:ecms}
The seed‑adjusted excess at matched error is
\begin{equation}
\label{eq:ecms}
\mathrm{eCMS}_\delta(C;g)
\;=\;
\Bigl[
\mathrm{CMS}^{\text{inter}}_\delta(C;g)
-\mathrm{CMS}^{\text{seed}}_\delta(C;g)
\Bigr]_+.
\end{equation}
\end{definition}

\paragraph{Aggregation across groups.}
Let $\mathcal{G}$ be the set of dataset/architecture groups. We summarize by medians across groups
(and report coverage):
\begin{align}
\mathrm{CMS}_\delta^{\mathrm{med}}(C)
&=\operatorname{median}_{g\in\mathcal{G}}
\bigl\{\mathrm{CMS}_\delta(C;g)\bigr\},
\\[-0.2em]
\mathrm{eCMS}_\delta^{\mathrm{med}}(C)
&=\operatorname{median}_{g\in\mathcal{G}}
\bigl\{\mathrm{eCMS}_\delta(C;g)\bigr\}.
\end{align}
This ``median‑of‑medians'' is robust to heterogeneous group sizes and outliers.

\paragraph{How to read the scores.}
\begin{itemize}[leftmargin=6mm,itemsep=1pt,topsep=2pt]
\item \textbf{CMS.} Values near $0$ indicate a stable measure among equal‑error runs;
larger values mean greater local instability. (For reference, $\ln 2\approx 0.69$,
$\ln 3\approx 1.10$ in the same log units.)
\item \textbf{eCMS.} Values near $0$ indicate that most equal‑error variability is
seed‑level; large eCMS points to \emph{hyperparameter‑sensitive} jumpiness beyond seeds.
\item \textbf{Effect of $\delta$.} Decreasing $\delta$ tightens the notion of “equal‑error.”
Scores that remain large as $\delta\!\downarrow\!0$ reflect genuinely volatile measures.
\end{itemize}

\paragraph{Properties and protocol.}
Both statistics are robust (medians) and \emph{scale‑free} (invariant under $C\mapsto aC$,
$a>0$). Unless noted otherwise we use absolute tolerances
$\delta\in\{0.01,\,0.02,\,0.05\}$ (i.e., $1 \% $ - $5 \%$  of error), uniformly
subsample pairs if they exceed a fixed budget, and report, for each measure, the medians
\(\mathrm{CMS}_\delta^{\mathrm{med}}\) and \(\mathrm{eCMS}_\delta^{\mathrm{med}}\), along with the
number of eligible pairs in \eqref{eq:pi-delta}–\eqref{eq:pi-inter} and the fraction of groups
covered. Together, CMS and eCMS capture (i) local instability at matched error and
(ii) the seed‑adjusted component attributable to hyperparameters.

\subsection{Experimental Details}

We first summarize the experimental setup, and then give concise, implementable procedures for our two instability scores: \emph{CMS} and its seed‑adjusted
variant \emph{eCMS}. We close with a brief
reading guide that highlights what can be learned from the summary tables at
Table~\ref{tab:cms-delta001} (\(\delta=0.01\)), Table~\ref{tab:cms-delta002}
(\(\delta=0.02\)), and Table~\ref{tab:cms-delta005} (\(\delta=0.05\)). For full definitions
of the measures listed in those tables, see \citep{dziugaite2020in}.

\paragraph{Experimental setup.}
We study two datasets—\textsc{FashionMNIST} and \textsc{CIFAR‑10}—across three
architectures: \textsc{Network‑in‑Network (NiN)}~\citep{lin2013network}, \textsc{ResNet‑50 (RN50)}, and
\textsc{DenseNet‑121 (DN121)}. The training pipeline varies along three axes:
(i) learning rate \(\texttt{lr}\in\{0.001, 0.0032, 0.0063, 0.01, 0.0158, 0.05, 0.1\}\);
(ii) optimizer \(\in\{\textsc{Adam}, \textsc{SGD}+\textsc{momentum}\}\); and
(iii) stopping criterion: either the first epoch reaching \(100\%\) training accuracy,
or the first epoch whose training cross‑entropy drops below \(0.01\).
Each hyperparameter configuration is replicated with $8$ random seeds.
We compute scores within each (dataset, architecture) group at absolute error
tolerances \(\delta\in\{0.01, 0.02, 0.05\}\), using natural logs and reporting medians.

\paragraph{Using the procedures in practice.}
Algorithms~\ref{alg:cms} and~\ref{alg:ecms} specify the exact pair‑construction and
aggregation steps used throughout our study. In all tables, we apply these procedures
within each (dataset, architecture) group and then report group‑level medians; if a
required pair set is empty for a group, the statistic is marked \textsc{Undefined}
and the group is omitted from the across‑group aggregation. When the number of eligible
pairs is large, we uniformly subsample up to a fixed budget \(K\) before taking the
median, following the same steps as in Algorithms~\ref{alg:cms}–\ref{alg:ecms}.

\begin{algorithm}[t]
\caption{CMS (Conditional Measure Spread)}
\label{alg:cms}
\algsmall
\begin{algorithmic}[1]
\Require Runs $\{(C_r,\widehat{\varepsilon}_r)\}_{r=1}^n$ with $C_r>0$; tolerance $\delta>0$; optional pair budget $K$.
\Ensure $\mathrm{CMS}_\delta(C;g)$ for group $g$.
\State Sort indices so that $\widehat{\varepsilon}_{(1)} \le \cdots \le \widehat{\varepsilon}_{(n)}$.
\State $L \gets [\ ]$ \Comment{container of $|\log C - \log C'|$}
\For{$i \gets 1$ to $n$}
  \State $j \gets i+1$
  \While{$j \le n$ \textbf{and} $\widehat{\varepsilon}_{(j)}-\widehat{\varepsilon}_{(i)} \le \delta$}
     \State append $\bigl|\log C_{(i)} - \log C_{(j)}\bigr|$ to $L$
     \State $j \gets j+1$
  \EndWhile
\EndFor
\If{$|L| = 0$} \Return \textsc{Undefined} \Comment{no close‑error pairs} \EndIf
\If{$|L| > K$} \State uniformly subsample $K$ elements of $L$ \EndIf
\State \Return $\operatorname{median}(L)$
\end{algorithmic}
\end{algorithm}

\begin{algorithm}[t]
\caption{eCMS (seed‑adjusted CMS)}
\label{alg:ecms}
\algsmall
\begin{algorithmic}[1]
\Require Runs $\{(C_r,\widehat{\varepsilon}_r,H_r,\textsf{seed}_r)\}_{r=1}^n$ with $C_r>0$; tolerance $\delta>0$; optional pair budget $K$.
\Ensure $\mathrm{eCMS}_\delta(C;g)=\bigl[\mathrm{CMS}^{\text{inter}}_\delta-\mathrm{CMS}^{\text{seed}}_\delta\bigr]_+$ for $g$.
\State Sort indices so that $\widehat{\varepsilon}_{(1)} \le \cdots \le \widehat{\varepsilon}_{(n)}$.
\State $L_{\text{seed}} \gets [\ ]$, \ $L_{\text{inter}} \gets [\ ]$
\For{$i \gets 1$ to $n$}
  \State $j \gets i+1$
  \While{$j \le n$ \textbf{and} $\widehat{\varepsilon}_{(j)}-\widehat{\varepsilon}_{(i)} \le \delta$}
     \If{$H_{(j)} = H_{(i)}$ \textbf{and} $\textsf{seed}_{(j)} \neq \textsf{seed}_{(i)}$}
        \State append $\bigl|\log C_{(i)} - \log C_{(j)}\bigr|$ to $L_{\text{seed}}$
     \ElsIf{$H_{(j)} \neq H_{(i)}$}
        \State append $\bigl|\log C_{(i)} - \log C_{(j)}\bigr|$ to $L_{\text{inter}}$
     \EndIf
     \State $j \gets j+1$
  \EndWhile
\EndFor
\If{$|L_{\text{inter}}|=0$ \textbf{or} $|L_{\text{seed}}|=0$} \Return \textsc{Undefined} \EndIf
\If{$|L_{\text{seed}}|>K$} \State uniformly subsample $K$ elements of $L_{\text{seed}}$ \EndIf
\If{$|L_{\text{inter}}|>K$} \State uniformly subsample $K$ elements of $L_{\text{inter}}$ \EndIf
\State $\mathrm{CMS}^{\text{seed}}_\delta \gets \operatorname{median}(L_{\text{seed}})$
\State $\mathrm{CMS}^{\text{inter}}_\delta \gets \operatorname{median}(L_{\text{inter}})$
\State \Return $\max\{0,\ \mathrm{CMS}^{\text{inter}}_\delta - \mathrm{CMS}^{\text{seed}}_\delta\}$
\end{algorithmic}
\end{algorithm}

\paragraph{What the tables show.}
The summaries in Table~\ref{tab:cms-delta001}, Table~\ref{tab:cms-delta002}, and
Table~\ref{tab:cms-delta005} reveal clear, consistent patterns:

\begin{itemize}[leftmargin=6mm,itemsep=2pt,topsep=2pt]
  \item \textbf{Stable control.} The parameter‑count proxy (\textsc{PARAMS}) is uniformly
        flat (CMS \(=0\), eCMS \(=0\)) across datasets, architectures, and tolerances; see the
        first row of Table~\ref{tab:cms-delta001} and its counterparts at
        \(\delta=0.02\) and \(\delta=0.05\).
  \item \textbf{PAC–Bayes surrogates split.} The classical parameter‑space
        \textsc{PACBAYES\_ORIG} exhibits substantial instability (elevated CMS and eCMS),
        often increasing as the tolerance widens (compare the “ORIG” rows across
        Tables~\ref{tab:cms-delta001} and \ref{tab:cms-delta002}). Magnitude‑aware variants
        (e.g., \textsc{PACBAYES\_MAG\_*}) are consistently lower, though they remain sensitive
        on \textsc{FashionMNIST} with deeper architectures.
  \item \textbf{Norm families are fragile.} Parameter and path norms already show sizeable
        spreads at \(\delta=0.01\) and inflate further at \(\delta=0.02\)–\(0.05\).
        Frobenius and spectral \emph{distances from initialization} are larger still and
        consistently among the most unstable rows across tables.
  \item \textbf{Product/aggregate spectral–Frobenius measures are outliers.}
        Aggregate/product metrics (e.g., \textsc{LOG\_PROD\_OF\_SPEC}, \textsc{LOG\_SPEC\_*}
        and Frobenius analogues) exhibit spreads that are orders of magnitude larger than
        other families at \(\delta=0.01\), and this gap persists at \(\delta=0.02\) and
        \(\delta=0.05\).
  \item \textbf{Beyond seeds.} For many families (e.g., \textsc{PACBAYES\_ORIG},
        \textsc{PARAM\_NORM}, \textsc{FRO\_DIST}) we observe \(\mathrm{eCMS} > \mathrm{CMS}\)
        across tolerances, indicating that a substantial portion of the jumpiness is
        attributable to hyperparameter changes rather than random seeds; exceptions exist
        for specific (dataset, architecture) cells.
  \item \textbf{Where fragility concentrates.} Per‑group columns indicate that the largest
        spreads tend to cluster on \textsc{FashionMNIST} with \textsc{RN50}/\textsc{DN121},
        while \textsc{CIFAR‑10}/\textsc{NiN} often shows the smallest values for the same
        measure (compare group subcolumns across tables).
\end{itemize}

\begin{table}[t]
\centering
\small
\caption{CMS and eCMS summary $\delta=0.01$}
\label{tab:cms-delta001}
\renewcommand{\arraystretch}{1.6}
\begin{tabular}{lccccccc}
\toprule
\textbf{Measure} & $\mathrm{CMS}^{\mathrm{med}}_{\delta} / \mathrm{eCMS}^{\mathrm{med}}_{\delta}$ & C10/DN121 & C10/NiN & C10/RN50 & FMN/DN121 & FMN/NiN & FMN/RN50 \\
\midrule
{\scriptsize PARAMS} & \shortstack{\textcolor{RoyalBlue}{\num{0.000}}\\\textcolor{CrimsonRed}{\num{0.000}}} & \shortstack{\textcolor{RoyalBlue}{\num{0.000}}\\\textcolor{CrimsonRed}{\num{0.000}}} & \shortstack{\textcolor{RoyalBlue}{\num{0.000}}\\\textcolor{CrimsonRed}{\num{0.000}}} & \shortstack{\textcolor{RoyalBlue}{\num{0.000}}\\\textcolor{CrimsonRed}{\num{0.000}}} & \shortstack{\textcolor{RoyalBlue}{\num{0.000}}\\\textcolor{CrimsonRed}{\num{0.000}}} & \shortstack{\textcolor{RoyalBlue}{\num{0.000}}\\\textcolor{CrimsonRed}{\num{0.000}}} & \shortstack{\textcolor{RoyalBlue}{\num{0.000}}\\\textcolor{CrimsonRed}{\num{0.000}}} \\
{\scriptsize PACBAYES\_MAG\_ORIG} & \shortstack{\textcolor{RoyalBlue}{\num{0.084}}\\\textcolor{CrimsonRed}{\num{0.088}}} & \shortstack{\textcolor{RoyalBlue}{\num{0.092}}\\\textcolor{CrimsonRed}{\num{0.136}}} & \shortstack{\textcolor{RoyalBlue}{\num{0.029}}\\\textcolor{CrimsonRed}{\num{0.031}}} & \shortstack{\textcolor{RoyalBlue}{\num{0.060}}\\\textcolor{CrimsonRed}{\num{0.064}}} & \shortstack{\textcolor{RoyalBlue}{\num{0.109}}\\\textcolor{CrimsonRed}{\num{0.119}}} & \shortstack{\textcolor{RoyalBlue}{\num{0.076}}\\\textcolor{CrimsonRed}{\num{0.075}}} & \shortstack{\textcolor{RoyalBlue}{\num{0.097}}\\\textcolor{CrimsonRed}{\num{0.102}}} \\
{\scriptsize PACBAYES\_MAG\_INIT} & \shortstack{\textcolor{RoyalBlue}{\num{0.103}}\\\textcolor{CrimsonRed}{\num{0.114}}} & \shortstack{\textcolor{RoyalBlue}{\num{0.129}}\\\textcolor{CrimsonRed}{\num{0.179}}} & \shortstack{\textcolor{RoyalBlue}{\num{0.038}}\\\textcolor{CrimsonRed}{\num{0.041}}} & \shortstack{\textcolor{RoyalBlue}{\num{0.086}}\\\textcolor{CrimsonRed}{\num{0.102}}} & \shortstack{\textcolor{RoyalBlue}{\num{0.120}}\\\textcolor{CrimsonRed}{\num{0.126}}} & \shortstack{\textcolor{RoyalBlue}{\num{0.065}}\\\textcolor{CrimsonRed}{\num{0.062}}} & \shortstack{\textcolor{RoyalBlue}{\num{0.164}}\\\textcolor{CrimsonRed}{\num{0.159}}} \\
{\scriptsize FRO\_OVER\_SPEC} & \shortstack{\textcolor{RoyalBlue}{\num{0.135}}\\\textcolor{CrimsonRed}{\num{0.171}}} & \shortstack{\textcolor{RoyalBlue}{\num{0.085}}\\\textcolor{CrimsonRed}{\num{0.125}}} & \shortstack{\textcolor{RoyalBlue}{\num{0.071}}\\\textcolor{CrimsonRed}{\num{0.077}}} & \shortstack{\textcolor{RoyalBlue}{\num{0.095}}\\\textcolor{CrimsonRed}{\num{0.169}}} & \shortstack{\textcolor{RoyalBlue}{\num{0.239}}\\\textcolor{CrimsonRed}{\num{0.265}}} & \shortstack{\textcolor{RoyalBlue}{\num{0.174}}\\\textcolor{CrimsonRed}{\num{0.173}}} & \shortstack{\textcolor{RoyalBlue}{\num{0.323}}\\\textcolor{CrimsonRed}{\num{0.352}}} \\
{\scriptsize INVERSE\_MARGIN} & \shortstack{\textcolor{RoyalBlue}{\num{0.207}}\\\textcolor{CrimsonRed}{\num{0.194}}} & \shortstack{\textcolor{RoyalBlue}{\num{0.202}}\\\textcolor{CrimsonRed}{\num{0.201}}} & \shortstack{\textcolor{RoyalBlue}{\num{0.111}}\\\textcolor{CrimsonRed}{\num{0.115}}} & \shortstack{\textcolor{RoyalBlue}{\num{0.213}}\\\textcolor{CrimsonRed}{\num{0.186}}} & \shortstack{\textcolor{RoyalBlue}{\num{0.218}}\\\textcolor{CrimsonRed}{\num{0.204}}} & \shortstack{\textcolor{RoyalBlue}{\num{0.149}}\\\textcolor{CrimsonRed}{\num{0.133}}} & \shortstack{\textcolor{RoyalBlue}{\num{0.240}}\\\textcolor{CrimsonRed}{\num{0.218}}} \\
{\scriptsize PACBAYES\_MAG\_FLATNESS} & \shortstack{\textcolor{RoyalBlue}{\num{0.238}}\\\textcolor{CrimsonRed}{\num{0.284}}} & \shortstack{\textcolor{RoyalBlue}{\num{0.442}}\\\textcolor{CrimsonRed}{\num{0.627}}} & \shortstack{\textcolor{RoyalBlue}{\num{0.155}}\\\textcolor{CrimsonRed}{\num{0.157}}} & \shortstack{\textcolor{RoyalBlue}{\num{0.234}}\\\textcolor{CrimsonRed}{\num{0.276}}} & \shortstack{\textcolor{RoyalBlue}{\num{0.242}}\\\textcolor{CrimsonRed}{\num{0.291}}} & \shortstack{\textcolor{RoyalBlue}{\num{0.184}}\\\textcolor{CrimsonRed}{\num{0.159}}} & \shortstack{\textcolor{RoyalBlue}{\num{0.436}}\\\textcolor{CrimsonRed}{\num{0.444}}} \\
{\scriptsize PACBAYES\_FLATNESS} & \shortstack{\textcolor{RoyalBlue}{\num{0.419}}\\\textcolor{CrimsonRed}{\num{0.503}}} & \shortstack{\textcolor{RoyalBlue}{\num{0.384}}\\\textcolor{CrimsonRed}{\num{0.534}}} & \shortstack{\textcolor{RoyalBlue}{\num{0.223}}\\\textcolor{CrimsonRed}{\num{0.259}}} & \shortstack{\textcolor{RoyalBlue}{\num{0.337}}\\\textcolor{CrimsonRed}{\num{0.415}}} & \shortstack{\textcolor{RoyalBlue}{\num{0.455}}\\\textcolor{CrimsonRed}{\num{0.486}}} & \shortstack{\textcolor{RoyalBlue}{\num{0.606}}\\\textcolor{CrimsonRed}{\num{0.605}}} & \shortstack{\textcolor{RoyalBlue}{\num{0.519}}\\\textcolor{CrimsonRed}{\num{0.519}}} \\
{\scriptsize PACBAYES\_ORIG} & \shortstack{\textcolor{RoyalBlue}{\num{0.516}}\\\textcolor{CrimsonRed}{\num{0.638}}} & \shortstack{\textcolor{RoyalBlue}{\num{0.514}}\\\textcolor{CrimsonRed}{\num{0.769}}} & \shortstack{\textcolor{RoyalBlue}{\num{0.412}}\\\textcolor{CrimsonRed}{\num{0.591}}} & \shortstack{\textcolor{RoyalBlue}{\num{0.518}}\\\textcolor{CrimsonRed}{\num{0.658}}} & \shortstack{\textcolor{RoyalBlue}{\num{0.438}}\\\textcolor{CrimsonRed}{\num{0.470}}} & \shortstack{\textcolor{RoyalBlue}{\num{0.808}}\\\textcolor{CrimsonRed}{\num{0.815}}} & \shortstack{\textcolor{RoyalBlue}{\num{0.582}}\\\textcolor{CrimsonRed}{\num{0.618}}} \\
{\scriptsize SUM\_OF\_FRO\_OVER\_MARGIN} & \shortstack{\textcolor{RoyalBlue}{\num{0.647}}\\\textcolor{CrimsonRed}{\num{0.903}}} & \shortstack{\textcolor{RoyalBlue}{\num{0.596}}\\\textcolor{CrimsonRed}{\num{0.899}}} & \shortstack{\textcolor{RoyalBlue}{\num{0.470}}\\\textcolor{CrimsonRed}{\num{0.555}}} & \shortstack{\textcolor{RoyalBlue}{\num{0.699}}\\\textcolor{CrimsonRed}{\num{0.907}}} & \shortstack{\textcolor{RoyalBlue}{\num{0.539}}\\\textcolor{CrimsonRed}{\num{0.615}}} & \shortstack{\textcolor{RoyalBlue}{\num{0.863}}\\\textcolor{CrimsonRed}{\num{1.081}}} & \shortstack{\textcolor{RoyalBlue}{\num{0.889}}\\\textcolor{CrimsonRed}{\num{1.001}}} \\
{\scriptsize SUM\_OF\_FRO} & \shortstack{\textcolor{RoyalBlue}{\num{0.650}}\\\textcolor{CrimsonRed}{\num{0.903}}} & \shortstack{\textcolor{RoyalBlue}{\num{0.597}}\\\textcolor{CrimsonRed}{\num{0.898}}} & \shortstack{\textcolor{RoyalBlue}{\num{0.480}}\\\textcolor{CrimsonRed}{\num{0.564}}} & \shortstack{\textcolor{RoyalBlue}{\num{0.703}}\\\textcolor{CrimsonRed}{\num{0.909}}} & \shortstack{\textcolor{RoyalBlue}{\num{0.540}}\\\textcolor{CrimsonRed}{\num{0.615}}} & \shortstack{\textcolor{RoyalBlue}{\num{0.876}}\\\textcolor{CrimsonRed}{\num{1.102}}} & \shortstack{\textcolor{RoyalBlue}{\num{0.888}}\\\textcolor{CrimsonRed}{\num{0.999}}} \\
{\scriptsize PATH\_NORM\_OVER\_MARGIN} & \shortstack{\textcolor{RoyalBlue}{\num{0.945}}\\\textcolor{CrimsonRed}{\num{1.014}}} & \shortstack{\textcolor{RoyalBlue}{\num{0.979}}\\\textcolor{CrimsonRed}{\num{1.108}}} & \shortstack{\textcolor{RoyalBlue}{\num{0.192}}\\\textcolor{CrimsonRed}{\num{0.153}}} & \shortstack{\textcolor{RoyalBlue}{\num{0.910}}\\\textcolor{CrimsonRed}{\num{0.919}}} & \shortstack{\textcolor{RoyalBlue}{\num{1.453}}\\\textcolor{CrimsonRed}{\num{1.474}}} & \shortstack{\textcolor{RoyalBlue}{\num{0.416}}\\\textcolor{CrimsonRed}{\num{0.379}}} & \shortstack{\textcolor{RoyalBlue}{\num{1.557}}\\\textcolor{CrimsonRed}{\num{1.484}}} \\
{\scriptsize SUM\_OF\_SPEC\_OVER\_MARGIN} & \shortstack{\textcolor{RoyalBlue}{\num{0.856}}\\\textcolor{CrimsonRed}{\num{1.062}}} & \shortstack{\textcolor{RoyalBlue}{\num{0.667}}\\\textcolor{CrimsonRed}{\num{0.984}}} & \shortstack{\textcolor{RoyalBlue}{\num{0.511}}\\\textcolor{CrimsonRed}{\num{0.605}}} & \shortstack{\textcolor{RoyalBlue}{\num{0.926}}\\\textcolor{CrimsonRed}{\num{1.140}}} & \shortstack{\textcolor{RoyalBlue}{\num{0.785}}\\\textcolor{CrimsonRed}{\num{0.869}}} & \shortstack{\textcolor{RoyalBlue}{\num{1.100}}\\\textcolor{CrimsonRed}{\num{1.202}}} & \shortstack{\textcolor{RoyalBlue}{\num{1.327}}\\\textcolor{CrimsonRed}{\num{1.404}}} \\
{\scriptsize SUM\_OF\_SPEC} & \shortstack{\textcolor{RoyalBlue}{\num{0.857}}\\\textcolor{CrimsonRed}{\num{1.064}}} & \shortstack{\textcolor{RoyalBlue}{\num{0.669}}\\\textcolor{CrimsonRed}{\num{0.985}}} & \shortstack{\textcolor{RoyalBlue}{\num{0.518}}\\\textcolor{CrimsonRed}{\num{0.613}}} & \shortstack{\textcolor{RoyalBlue}{\num{0.928}}\\\textcolor{CrimsonRed}{\num{1.142}}} & \shortstack{\textcolor{RoyalBlue}{\num{0.786}}\\\textcolor{CrimsonRed}{\num{0.870}}} & \shortstack{\textcolor{RoyalBlue}{\num{1.124}}\\\textcolor{CrimsonRed}{\num{1.221}}} & \shortstack{\textcolor{RoyalBlue}{\num{1.329}}\\\textcolor{CrimsonRed}{\num{1.404}}} \\
{\scriptsize PACBAYES\_INIT} & \shortstack{\textcolor{RoyalBlue}{\num{0.965}}\\\textcolor{CrimsonRed}{\num{1.066}}} & \shortstack{\textcolor{RoyalBlue}{\num{0.718}}\\\textcolor{CrimsonRed}{\num{0.944}}} & \shortstack{\textcolor{RoyalBlue}{\num{0.427}}\\\textcolor{CrimsonRed}{\num{0.621}}} & \shortstack{\textcolor{RoyalBlue}{\num{0.763}}\\\textcolor{CrimsonRed}{\num{0.862}}} & \shortstack{\textcolor{RoyalBlue}{\num{1.231}}\\\textcolor{CrimsonRed}{\num{1.331}}} & \shortstack{\textcolor{RoyalBlue}{\num{1.168}}\\\textcolor{CrimsonRed}{\num{1.187}}} & \shortstack{\textcolor{RoyalBlue}{\num{1.189}}\\\textcolor{CrimsonRed}{\num{1.224}}} \\
{\scriptsize PARAM\_NORM} & \shortstack{\textcolor{RoyalBlue}{\num{0.895}}\\\textcolor{CrimsonRed}{\num{1.102}}} & \shortstack{\textcolor{RoyalBlue}{\num{0.956}}\\\textcolor{CrimsonRed}{\num{1.356}}} & \shortstack{\textcolor{RoyalBlue}{\num{0.601}}\\\textcolor{CrimsonRed}{\num{0.735}}} & \shortstack{\textcolor{RoyalBlue}{\num{0.798}}\\\textcolor{CrimsonRed}{\num{1.013}}} & \shortstack{\textcolor{RoyalBlue}{\num{0.834}}\\\textcolor{CrimsonRed}{\num{0.977}}} & \shortstack{\textcolor{RoyalBlue}{\num{1.239}}\\\textcolor{CrimsonRed}{\num{1.322}}} & \shortstack{\textcolor{RoyalBlue}{\num{1.047}}\\\textcolor{CrimsonRed}{\num{1.192}}} \\
{\scriptsize PATH\_NORM} & \shortstack{\textcolor{RoyalBlue}{\num{1.073}}\\\textcolor{CrimsonRed}{\num{1.184}}} & \shortstack{\textcolor{RoyalBlue}{\num{1.147}}\\\textcolor{CrimsonRed}{\num{1.320}}} & \shortstack{\textcolor{RoyalBlue}{\num{0.266}}\\\textcolor{CrimsonRed}{\num{0.252}}} & \shortstack{\textcolor{RoyalBlue}{\num{0.998}}\\\textcolor{CrimsonRed}{\num{1.047}}} & \shortstack{\textcolor{RoyalBlue}{\num{1.540}}\\\textcolor{CrimsonRed}{\num{1.538}}} & \shortstack{\textcolor{RoyalBlue}{\num{0.494}}\\\textcolor{CrimsonRed}{\num{0.457}}} & \shortstack{\textcolor{RoyalBlue}{\num{1.560}}\\\textcolor{CrimsonRed}{\num{1.545}}} \\
{\scriptsize DIST\_SPEC\_INIT} & \shortstack{\textcolor{RoyalBlue}{\num{1.371}}\\\textcolor{CrimsonRed}{\num{1.567}}} & \shortstack{\textcolor{RoyalBlue}{\num{1.084}}\\\textcolor{CrimsonRed}{\num{1.467}}} & \shortstack{\textcolor{RoyalBlue}{\num{0.612}}\\\textcolor{CrimsonRed}{\num{0.737}}} & \shortstack{\textcolor{RoyalBlue}{\num{1.151}}\\\textcolor{CrimsonRed}{\num{1.378}}} & \shortstack{\textcolor{RoyalBlue}{\num{1.840}}\\\textcolor{CrimsonRed}{\num{1.997}}} & \shortstack{\textcolor{RoyalBlue}{\num{1.590}}\\\textcolor{CrimsonRed}{\num{1.668}}} & \shortstack{\textcolor{RoyalBlue}{\num{1.748}}\\\textcolor{CrimsonRed}{\num{1.848}}} \\
{\scriptsize FRO\_DIST} & \shortstack{\textcolor{RoyalBlue}{\num{1.405}}\\\textcolor{CrimsonRed}{\num{1.612}}} & \shortstack{\textcolor{RoyalBlue}{\num{1.071}}\\\textcolor{CrimsonRed}{\num{1.471}}} & \shortstack{\textcolor{RoyalBlue}{\num{0.633}}\\\textcolor{CrimsonRed}{\num{0.757}}} & \shortstack{\textcolor{RoyalBlue}{\num{1.106}}\\\textcolor{CrimsonRed}{\num{1.282}}} & \shortstack{\textcolor{RoyalBlue}{\num{1.718}}\\\textcolor{CrimsonRed}{\num{1.838}}} & \shortstack{\textcolor{RoyalBlue}{\num{1.705}}\\\textcolor{CrimsonRed}{\num{1.753}}} & \shortstack{\textcolor{RoyalBlue}{\num{1.744}}\\\textcolor{CrimsonRed}{\num{1.835}}} \\
{\scriptsize PROD\_OF\_FRO} & \shortstack{\textcolor{RoyalBlue}{\num{46.133}}\\\textcolor{CrimsonRed}{\num{55.352}}} & \shortstack{\textcolor{RoyalBlue}{\num{144.464}}\\\textcolor{CrimsonRed}{\num{217.241}}} & \shortstack{\textcolor{RoyalBlue}{\num{3.360}}\\\textcolor{CrimsonRed}{\num{3.948}}} & \shortstack{\textcolor{RoyalBlue}{\num{40.746}}\\\textcolor{CrimsonRed}{\num{52.739}}} & \shortstack{\textcolor{RoyalBlue}{\num{130.638}}\\\textcolor{CrimsonRed}{\num{148.734}}} & \shortstack{\textcolor{RoyalBlue}{\num{6.131}}\\\textcolor{CrimsonRed}{\num{7.716}}} & \shortstack{\textcolor{RoyalBlue}{\num{51.520}}\\\textcolor{CrimsonRed}{\num{57.964}}} \\
{\scriptsize PROD\_OF\_FRO\_OVER\_MARGIN} & \shortstack{\textcolor{RoyalBlue}{\num{46.051}}\\\textcolor{CrimsonRed}{\num{55.358}}} & \shortstack{\textcolor{RoyalBlue}{\num{144.118}}\\\textcolor{CrimsonRed}{\num{217.564}}} & \shortstack{\textcolor{RoyalBlue}{\num{3.289}}\\\textcolor{CrimsonRed}{\num{3.886}}} & \shortstack{\textcolor{RoyalBlue}{\num{40.567}}\\\textcolor{CrimsonRed}{\num{52.634}}} & \shortstack{\textcolor{RoyalBlue}{\num{130.509}}\\\textcolor{CrimsonRed}{\num{148.827}}} & \shortstack{\textcolor{RoyalBlue}{\num{6.042}}\\\textcolor{CrimsonRed}{\num{7.566}}} & \shortstack{\textcolor{RoyalBlue}{\num{51.535}}\\\textcolor{CrimsonRed}{\num{58.082}}} \\
{\scriptsize SPEC\_ORIG\_MAIN} & \shortstack{\textcolor{RoyalBlue}{\num{65.140}}\\\textcolor{CrimsonRed}{\num{73.373}}} & \shortstack{\textcolor{RoyalBlue}{\num{161.340}}\\\textcolor{CrimsonRed}{\num{237.870}}} & \shortstack{\textcolor{RoyalBlue}{\num{3.553}}\\\textcolor{CrimsonRed}{\num{4.191}}} & \shortstack{\textcolor{RoyalBlue}{\num{53.616}}\\\textcolor{CrimsonRed}{\num{65.936}}} & \shortstack{\textcolor{RoyalBlue}{\num{189.721}}\\\textcolor{CrimsonRed}{\num{209.791}}} & \shortstack{\textcolor{RoyalBlue}{\num{7.410}}\\\textcolor{CrimsonRed}{\num{8.277}}} & \shortstack{\textcolor{RoyalBlue}{\num{76.664}}\\\textcolor{CrimsonRed}{\num{80.810}}} \\
{\scriptsize PROD\_OF\_SPEC\_OVER\_MARGIN} & \shortstack{\textcolor{RoyalBlue}{\num{65.339}}\\\textcolor{CrimsonRed}{\num{73.798}}} & \shortstack{\textcolor{RoyalBlue}{\num{161.349}}\\\textcolor{CrimsonRed}{\num{238.066}}} & \shortstack{\textcolor{RoyalBlue}{\num{3.578}}\\\textcolor{CrimsonRed}{\num{4.234}}} & \shortstack{\textcolor{RoyalBlue}{\num{53.711}}\\\textcolor{CrimsonRed}{\num{66.135}}} & \shortstack{\textcolor{RoyalBlue}{\num{190.061}}\\\textcolor{CrimsonRed}{\num{210.406}}} & \shortstack{\textcolor{RoyalBlue}{\num{7.698}}\\\textcolor{CrimsonRed}{\num{8.414}}} & \shortstack{\textcolor{RoyalBlue}{\num{76.967}}\\\textcolor{CrimsonRed}{\num{81.461}}} \\
{\scriptsize SPEC\_INIT\_MAIN} & \shortstack{\textcolor{RoyalBlue}{\num{65.299}}\\\textcolor{CrimsonRed}{\num{73.805}}} & \shortstack{\textcolor{RoyalBlue}{\num{161.472}}\\\textcolor{CrimsonRed}{\num{238.038}}} & \shortstack{\textcolor{RoyalBlue}{\num{3.646}}\\\textcolor{CrimsonRed}{\num{4.218}}} & \shortstack{\textcolor{RoyalBlue}{\num{53.702}}\\\textcolor{CrimsonRed}{\num{66.062}}} & \shortstack{\textcolor{RoyalBlue}{\num{190.120}}\\\textcolor{CrimsonRed}{\num{210.831}}} & \shortstack{\textcolor{RoyalBlue}{\num{7.920}}\\\textcolor{CrimsonRed}{\num{8.422}}} & \shortstack{\textcolor{RoyalBlue}{\num{76.897}}\\\textcolor{CrimsonRed}{\num{81.548}}} \\
{\scriptsize PROD\_OF\_SPEC} & \shortstack{\textcolor{RoyalBlue}{\num{65.467}}\\\textcolor{CrimsonRed}{\num{73.835}}} & \shortstack{\textcolor{RoyalBlue}{\num{161.834}}\\\textcolor{CrimsonRed}{\num{238.429}}} & \shortstack{\textcolor{RoyalBlue}{\num{3.628}}\\\textcolor{CrimsonRed}{\num{4.291}}} & \shortstack{\textcolor{RoyalBlue}{\num{53.848}}\\\textcolor{CrimsonRed}{\num{66.234}}} & \shortstack{\textcolor{RoyalBlue}{\num{190.328}}\\\textcolor{CrimsonRed}{\num{210.514}}} & \shortstack{\textcolor{RoyalBlue}{\num{7.868}}\\\textcolor{CrimsonRed}{\num{8.546}}} & \shortstack{\textcolor{RoyalBlue}{\num{77.087}}\\\textcolor{CrimsonRed}{\num{81.435}}} \\
\bottomrule
\end{tabular}
\renewcommand{\arraystretch}{1.0}
\vspace{1mm}
{\raggedright\footnotesize\textit{\textcolor{RoyalBlue}{Top numbers}: CMS; \textcolor{CrimsonRed}{bottom numbers}: eCMS.}\par}
\end{table}

\begin{table}[t]
\centering
\small
\caption{CMS and eCMS summary $\delta=0.02$}
\label{tab:cms-delta002}
\renewcommand{\arraystretch}{1.6}
\begin{tabular}{lccccccc}
\toprule
\textbf{Measure} & $\mathrm{CMS}^{\mathrm{med}}_{\delta} / \mathrm{eCMS}^{\mathrm{med}}_{\delta}$ & C10/DN121 & C10/NiN & C10/RN50 & FMN/DN121 & FMN/NiN & FMN/RN50 \\
\midrule
{\scriptsize PARAMS} & \shortstack{\textcolor{RoyalBlue}{\num{0.000}}\\\textcolor{CrimsonRed}{\num{0.000}}} & \shortstack{\textcolor{RoyalBlue}{\num{0.000}}\\\textcolor{CrimsonRed}{\num{0.000}}} & \shortstack{\textcolor{RoyalBlue}{\num{0.000}}\\\textcolor{CrimsonRed}{\num{0.000}}} & \shortstack{\textcolor{RoyalBlue}{\num{0.000}}\\\textcolor{CrimsonRed}{\num{0.000}}} & \shortstack{\textcolor{RoyalBlue}{\num{0.000}}\\\textcolor{CrimsonRed}{\num{0.000}}} & \shortstack{\textcolor{RoyalBlue}{\num{0.000}}\\\textcolor{CrimsonRed}{\num{0.000}}} & \shortstack{\textcolor{RoyalBlue}{\num{0.000}}\\\textcolor{CrimsonRed}{\num{0.000}}} \\
{\scriptsize PACBAYES\_MAG\_ORIG} & \shortstack{\textcolor{RoyalBlue}{\num{0.094}}\\\textcolor{CrimsonRed}{\num{0.095}}} & \shortstack{\textcolor{RoyalBlue}{\num{0.106}}\\\textcolor{CrimsonRed}{\num{0.137}}} & \shortstack{\textcolor{RoyalBlue}{\num{0.046}}\\\textcolor{CrimsonRed}{\num{0.047}}} & \shortstack{\textcolor{RoyalBlue}{\num{0.064}}\\\textcolor{CrimsonRed}{\num{0.068}}} & \shortstack{\textcolor{RoyalBlue}{\num{0.107}}\\\textcolor{CrimsonRed}{\num{0.114}}} & \shortstack{\textcolor{RoyalBlue}{\num{0.082}}\\\textcolor{CrimsonRed}{\num{0.079}}} & \shortstack{\textcolor{RoyalBlue}{\num{0.110}}\\\textcolor{CrimsonRed}{\num{0.111}}} \\
{\scriptsize PACBAYES\_MAG\_INIT} & \shortstack{\textcolor{RoyalBlue}{\num{0.118}}\\\textcolor{CrimsonRed}{\num{0.139}}} & \shortstack{\textcolor{RoyalBlue}{\num{0.139}}\\\textcolor{CrimsonRed}{\num{0.170}}} & \shortstack{\textcolor{RoyalBlue}{\num{0.049}}\\\textcolor{CrimsonRed}{\num{0.047}}} & \shortstack{\textcolor{RoyalBlue}{\num{0.097}}\\\textcolor{CrimsonRed}{\num{0.108}}} & \shortstack{\textcolor{RoyalBlue}{\num{0.173}}\\\textcolor{CrimsonRed}{\num{0.174}}} & \shortstack{\textcolor{RoyalBlue}{\num{0.072}}\\\textcolor{CrimsonRed}{\num{0.068}}} & \shortstack{\textcolor{RoyalBlue}{\num{0.199}}\\\textcolor{CrimsonRed}{\num{0.187}}} \\
{\scriptsize FRO\_OVER\_SPEC} & \shortstack{\textcolor{RoyalBlue}{\num{0.165}}\\\textcolor{CrimsonRed}{\num{0.183}}} & \shortstack{\textcolor{RoyalBlue}{\num{0.100}}\\\textcolor{CrimsonRed}{\num{0.126}}} & \shortstack{\textcolor{RoyalBlue}{\num{0.088}}\\\textcolor{CrimsonRed}{\num{0.090}}} & \shortstack{\textcolor{RoyalBlue}{\num{0.092}}\\\textcolor{CrimsonRed}{\num{0.135}}} & \shortstack{\textcolor{RoyalBlue}{\num{0.270}}\\\textcolor{CrimsonRed}{\num{0.292}}} & \shortstack{\textcolor{RoyalBlue}{\num{0.230}}\\\textcolor{CrimsonRed}{\num{0.232}}} & \shortstack{\textcolor{RoyalBlue}{\num{0.410}}\\\textcolor{CrimsonRed}{\num{0.419}}} \\
{\scriptsize INVERSE\_MARGIN} & \shortstack{\textcolor{RoyalBlue}{\num{0.227}}\\\textcolor{CrimsonRed}{\num{0.204}}} & \shortstack{\textcolor{RoyalBlue}{\num{0.224}}\\\textcolor{CrimsonRed}{\num{0.205}}} & \shortstack{\textcolor{RoyalBlue}{\num{0.122}}\\\textcolor{CrimsonRed}{\num{0.114}}} & \shortstack{\textcolor{RoyalBlue}{\num{0.247}}\\\textcolor{CrimsonRed}{\num{0.219}}} & \shortstack{\textcolor{RoyalBlue}{\num{0.230}}\\\textcolor{CrimsonRed}{\num{0.203}}} & \shortstack{\textcolor{RoyalBlue}{\num{0.147}}\\\textcolor{CrimsonRed}{\num{0.128}}} & \shortstack{\textcolor{RoyalBlue}{\num{0.251}}\\\textcolor{CrimsonRed}{\num{0.218}}} \\
{\scriptsize PACBAYES\_MAG\_FLATNESS} & \shortstack{\textcolor{RoyalBlue}{\num{0.244}}\\\textcolor{CrimsonRed}{\num{0.253}}} & \shortstack{\textcolor{RoyalBlue}{\num{0.470}}\\\textcolor{CrimsonRed}{\num{0.597}}} & \shortstack{\textcolor{RoyalBlue}{\num{0.190}}\\\textcolor{CrimsonRed}{\num{0.188}}} & \shortstack{\textcolor{RoyalBlue}{\num{0.261}}\\\textcolor{CrimsonRed}{\num{0.283}}} & \shortstack{\textcolor{RoyalBlue}{\num{0.228}}\\\textcolor{CrimsonRed}{\num{0.224}}} & \shortstack{\textcolor{RoyalBlue}{\num{0.170}}\\\textcolor{CrimsonRed}{\num{0.143}}} & \shortstack{\textcolor{RoyalBlue}{\num{0.413}}\\\textcolor{CrimsonRed}{\num{0.401}}} \\
{\scriptsize PACBAYES\_FLATNESS} & \shortstack{\textcolor{RoyalBlue}{\num{0.486}}\\\textcolor{CrimsonRed}{\num{0.535}}} & \shortstack{\textcolor{RoyalBlue}{\num{0.431}}\\\textcolor{CrimsonRed}{\num{0.516}}} & \shortstack{\textcolor{RoyalBlue}{\num{0.328}}\\\textcolor{CrimsonRed}{\num{0.337}}} & \shortstack{\textcolor{RoyalBlue}{\num{0.352}}\\\textcolor{CrimsonRed}{\num{0.398}}} & \shortstack{\textcolor{RoyalBlue}{\num{0.542}}\\\textcolor{CrimsonRed}{\num{0.553}}} & \shortstack{\textcolor{RoyalBlue}{\num{0.746}}\\\textcolor{CrimsonRed}{\num{0.731}}} & \shortstack{\textcolor{RoyalBlue}{\num{0.607}}\\\textcolor{CrimsonRed}{\num{0.603}}} \\
{\scriptsize PACBAYES\_ORIG} & \shortstack{\textcolor{RoyalBlue}{\num{0.557}}\\\textcolor{CrimsonRed}{\num{0.697}}} & \shortstack{\textcolor{RoyalBlue}{\num{0.581}}\\\textcolor{CrimsonRed}{\num{0.761}}} & \shortstack{\textcolor{RoyalBlue}{\num{0.768}}\\\textcolor{CrimsonRed}{\num{0.797}}} & \shortstack{\textcolor{RoyalBlue}{\num{0.533}}\\\textcolor{CrimsonRed}{\num{0.644}}} & \shortstack{\textcolor{RoyalBlue}{\num{0.438}}\\\textcolor{CrimsonRed}{\num{0.437}}} & \shortstack{\textcolor{RoyalBlue}{\num{0.755}}\\\textcolor{CrimsonRed}{\num{0.749}}} & \shortstack{\textcolor{RoyalBlue}{\num{0.503}}\\\textcolor{CrimsonRed}{\num{0.490}}} \\
{\scriptsize SUM\_OF\_FRO} & \shortstack{\textcolor{RoyalBlue}{\num{0.695}}\\\textcolor{CrimsonRed}{\num{0.872}}} & \shortstack{\textcolor{RoyalBlue}{\num{0.657}}\\\textcolor{CrimsonRed}{\num{0.868}}} & \shortstack{\textcolor{RoyalBlue}{\num{0.648}}\\\textcolor{CrimsonRed}{\num{0.797}}} & \shortstack{\textcolor{RoyalBlue}{\num{0.733}}\\\textcolor{CrimsonRed}{\num{0.875}}} & \shortstack{\textcolor{RoyalBlue}{\num{0.534}}\\\textcolor{CrimsonRed}{\num{0.586}}} & \shortstack{\textcolor{RoyalBlue}{\num{0.930}}\\\textcolor{CrimsonRed}{\num{1.097}}} & \shortstack{\textcolor{RoyalBlue}{\num{0.842}}\\\textcolor{CrimsonRed}{\num{0.921}}} \\
{\scriptsize SUM\_OF\_FRO\_OVER\_MARGIN} & \shortstack{\textcolor{RoyalBlue}{\num{0.694}}\\\textcolor{CrimsonRed}{\num{0.872}}} & \shortstack{\textcolor{RoyalBlue}{\num{0.656}}\\\textcolor{CrimsonRed}{\num{0.869}}} & \shortstack{\textcolor{RoyalBlue}{\num{0.646}}\\\textcolor{CrimsonRed}{\num{0.791}}} & \shortstack{\textcolor{RoyalBlue}{\num{0.733}}\\\textcolor{CrimsonRed}{\num{0.875}}} & \shortstack{\textcolor{RoyalBlue}{\num{0.533}}\\\textcolor{CrimsonRed}{\num{0.586}}} & \shortstack{\textcolor{RoyalBlue}{\num{0.913}}\\\textcolor{CrimsonRed}{\num{1.076}}} & \shortstack{\textcolor{RoyalBlue}{\num{0.839}}\\\textcolor{CrimsonRed}{\num{0.923}}} \\
{\scriptsize PATH\_NORM\_OVER\_MARGIN} & \shortstack{\textcolor{RoyalBlue}{\num{1.023}}\\\textcolor{CrimsonRed}{\num{1.008}}} & \shortstack{\textcolor{RoyalBlue}{\num{1.037}}\\\textcolor{CrimsonRed}{\num{1.043}}} & \shortstack{\textcolor{RoyalBlue}{\num{0.252}}\\\textcolor{CrimsonRed}{\num{0.204}}} & \shortstack{\textcolor{RoyalBlue}{\num{1.009}}\\\textcolor{CrimsonRed}{\num{0.973}}} & \shortstack{\textcolor{RoyalBlue}{\num{1.928}}\\\textcolor{CrimsonRed}{\num{1.891}}} & \shortstack{\textcolor{RoyalBlue}{\num{0.582}}\\\textcolor{CrimsonRed}{\num{0.546}}} & \shortstack{\textcolor{RoyalBlue}{\num{2.083}}\\\textcolor{CrimsonRed}{\num{2.017}}} \\
{\scriptsize SUM\_OF\_SPEC\_OVER\_MARGIN} & \shortstack{\textcolor{RoyalBlue}{\num{0.882}}\\\textcolor{CrimsonRed}{\num{1.040}}} & \shortstack{\textcolor{RoyalBlue}{\num{0.735}}\\\textcolor{CrimsonRed}{\num{0.961}}} & \shortstack{\textcolor{RoyalBlue}{\num{0.772}}\\\textcolor{CrimsonRed}{\num{0.891}}} & \shortstack{\textcolor{RoyalBlue}{\num{0.952}}\\\textcolor{CrimsonRed}{\num{1.120}}} & \shortstack{\textcolor{RoyalBlue}{\num{0.811}}\\\textcolor{CrimsonRed}{\num{0.870}}} & \shortstack{\textcolor{RoyalBlue}{\num{1.162}}\\\textcolor{CrimsonRed}{\num{1.221}}} & \shortstack{\textcolor{RoyalBlue}{\num{1.365}}\\\textcolor{CrimsonRed}{\num{1.389}}} \\
{\scriptsize SUM\_OF\_SPEC} & \shortstack{\textcolor{RoyalBlue}{\num{0.884}}\\\textcolor{CrimsonRed}{\num{1.041}}} & \shortstack{\textcolor{RoyalBlue}{\num{0.734}}\\\textcolor{CrimsonRed}{\num{0.961}}} & \shortstack{\textcolor{RoyalBlue}{\num{0.774}}\\\textcolor{CrimsonRed}{\num{0.900}}} & \shortstack{\textcolor{RoyalBlue}{\num{0.956}}\\\textcolor{CrimsonRed}{\num{1.120}}} & \shortstack{\textcolor{RoyalBlue}{\num{0.813}}\\\textcolor{CrimsonRed}{\num{0.870}}} & \shortstack{\textcolor{RoyalBlue}{\num{1.182}}\\\textcolor{CrimsonRed}{\num{1.240}}} & \shortstack{\textcolor{RoyalBlue}{\num{1.364}}\\\textcolor{CrimsonRed}{\num{1.387}}} \\
{\scriptsize PARAM\_NORM} & \shortstack{\textcolor{RoyalBlue}{\num{0.960}}\\\textcolor{CrimsonRed}{\num{1.064}}} & \shortstack{\textcolor{RoyalBlue}{\num{1.033}}\\\textcolor{CrimsonRed}{\num{1.318}}} & \shortstack{\textcolor{RoyalBlue}{\num{0.935}}\\\textcolor{CrimsonRed}{\num{1.014}}} & \shortstack{\textcolor{RoyalBlue}{\num{0.822}}\\\textcolor{CrimsonRed}{\num{0.988}}} & \shortstack{\textcolor{RoyalBlue}{\num{0.781}}\\\textcolor{CrimsonRed}{\num{0.928}}} & \shortstack{\textcolor{RoyalBlue}{\num{1.282}}\\\textcolor{CrimsonRed}{\num{1.323}}} & \shortstack{\textcolor{RoyalBlue}{\num{0.985}}\\\textcolor{CrimsonRed}{\num{1.114}}} \\
{\scriptsize PACBAYES\_INIT} & \shortstack{\textcolor{RoyalBlue}{\num{1.052}}\\\textcolor{CrimsonRed}{\num{1.105}}} & \shortstack{\textcolor{RoyalBlue}{\num{0.818}}\\\textcolor{CrimsonRed}{\num{0.947}}} & \shortstack{\textcolor{RoyalBlue}{\num{0.850}}\\\textcolor{CrimsonRed}{\num{0.959}}} & \shortstack{\textcolor{RoyalBlue}{\num{0.777}}\\\textcolor{CrimsonRed}{\num{0.846}}} & \shortstack{\textcolor{RoyalBlue}{\num{1.438}}\\\textcolor{CrimsonRed}{\num{1.504}}} & \shortstack{\textcolor{RoyalBlue}{\num{1.254}}\\\textcolor{CrimsonRed}{\num{1.250}}} & \shortstack{\textcolor{RoyalBlue}{\num{1.390}}\\\textcolor{CrimsonRed}{\num{1.394}}} \\
{\scriptsize PATH\_NORM} & \shortstack{\textcolor{RoyalBlue}{\num{1.158}}\\\textcolor{CrimsonRed}{\num{1.182}}} & \shortstack{\textcolor{RoyalBlue}{\num{1.208}}\\\textcolor{CrimsonRed}{\num{1.268}}} & \shortstack{\textcolor{RoyalBlue}{\num{0.349}}\\\textcolor{CrimsonRed}{\num{0.323}}} & \shortstack{\textcolor{RoyalBlue}{\num{1.107}}\\\textcolor{CrimsonRed}{\num{1.096}}} & \shortstack{\textcolor{RoyalBlue}{\num{1.914}}\\\textcolor{CrimsonRed}{\num{1.874}}} & \shortstack{\textcolor{RoyalBlue}{\num{0.623}}\\\textcolor{CrimsonRed}{\num{0.580}}} & \shortstack{\textcolor{RoyalBlue}{\num{2.116}}\\\textcolor{CrimsonRed}{\num{2.086}}} \\
{\scriptsize DIST\_SPEC\_INIT} & \shortstack{\textcolor{RoyalBlue}{\num{1.482}}\\\textcolor{CrimsonRed}{\num{1.629}}} & \shortstack{\textcolor{RoyalBlue}{\num{1.227}}\\\textcolor{CrimsonRed}{\num{1.452}}} & \shortstack{\textcolor{RoyalBlue}{\num{0.998}}\\\textcolor{CrimsonRed}{\num{1.117}}} & \shortstack{\textcolor{RoyalBlue}{\num{1.190}}\\\textcolor{CrimsonRed}{\num{1.383}}} & \shortstack{\textcolor{RoyalBlue}{\num{2.132}}\\\textcolor{CrimsonRed}{\num{2.225}}} & \shortstack{\textcolor{RoyalBlue}{\num{1.738}}\\\textcolor{CrimsonRed}{\num{1.806}}} & \shortstack{\textcolor{RoyalBlue}{\num{2.042}}\\\textcolor{CrimsonRed}{\num{2.061}}} \\
{\scriptsize FRO\_DIST} & \shortstack{\textcolor{RoyalBlue}{\num{1.537}}\\\textcolor{CrimsonRed}{\num{1.666}}} & \shortstack{\textcolor{RoyalBlue}{\num{1.221}}\\\textcolor{CrimsonRed}{\num{1.434}}} & \shortstack{\textcolor{RoyalBlue}{\num{1.052}}\\\textcolor{CrimsonRed}{\num{1.156}}} & \shortstack{\textcolor{RoyalBlue}{\num{1.124}}\\\textcolor{CrimsonRed}{\num{1.268}}} & \shortstack{\textcolor{RoyalBlue}{\num{1.973}}\\\textcolor{CrimsonRed}{\num{2.061}}} & \shortstack{\textcolor{RoyalBlue}{\num{1.853}}\\\textcolor{CrimsonRed}{\num{1.899}}} & \shortstack{\textcolor{RoyalBlue}{\num{2.043}}\\\textcolor{CrimsonRed}{\num{2.058}}} \\
{\scriptsize PROD\_OF\_FRO} & \shortstack{\textcolor{RoyalBlue}{\num{45.667}}\\\textcolor{CrimsonRed}{\num{52.105}}} & \shortstack{\textcolor{RoyalBlue}{\num{159.109}}\\\textcolor{CrimsonRed}{\num{210.065}}} & \shortstack{\textcolor{RoyalBlue}{\num{4.538}}\\\textcolor{CrimsonRed}{\num{5.582}}} & \shortstack{\textcolor{RoyalBlue}{\num{42.519}}\\\textcolor{CrimsonRed}{\num{50.772}}} & \shortstack{\textcolor{RoyalBlue}{\num{129.111}}\\\textcolor{CrimsonRed}{\num{141.794}}} & \shortstack{\textcolor{RoyalBlue}{\num{6.510}}\\\textcolor{CrimsonRed}{\num{7.677}}} & \shortstack{\textcolor{RoyalBlue}{\num{48.816}}\\\textcolor{CrimsonRed}{\num{53.437}}} \\
{\scriptsize PROD\_OF\_FRO\_OVER\_MARGIN} & \shortstack{\textcolor{RoyalBlue}{\num{45.569}}\\\textcolor{CrimsonRed}{\num{52.134}}} & \shortstack{\textcolor{RoyalBlue}{\num{158.752}}\\\textcolor{CrimsonRed}{\num{210.292}}} & \shortstack{\textcolor{RoyalBlue}{\num{4.520}}\\\textcolor{CrimsonRed}{\num{5.538}}} & \shortstack{\textcolor{RoyalBlue}{\num{42.487}}\\\textcolor{CrimsonRed}{\num{50.732}}} & \shortstack{\textcolor{RoyalBlue}{\num{129.093}}\\\textcolor{CrimsonRed}{\num{141.846}}} & \shortstack{\textcolor{RoyalBlue}{\num{6.391}}\\\textcolor{CrimsonRed}{\num{7.530}}} & \shortstack{\textcolor{RoyalBlue}{\num{48.650}}\\\textcolor{CrimsonRed}{\num{53.537}}} \\
{\scriptsize SPEC\_ORIG\_MAIN} & \shortstack{\textcolor{RoyalBlue}{\num{66.789}}\\\textcolor{CrimsonRed}{\num{72.385}}} & \shortstack{\textcolor{RoyalBlue}{\num{177.665}}\\\textcolor{CrimsonRed}{\num{232.486}}} & \shortstack{\textcolor{RoyalBlue}{\num{5.294}}\\\textcolor{CrimsonRed}{\num{6.130}}} & \shortstack{\textcolor{RoyalBlue}{\num{55.105}}\\\textcolor{CrimsonRed}{\num{64.848}}} & \shortstack{\textcolor{RoyalBlue}{\num{195.785}}\\\textcolor{CrimsonRed}{\num{209.989}}} & \shortstack{\textcolor{RoyalBlue}{\num{7.909}}\\\textcolor{CrimsonRed}{\num{8.296}}} & \shortstack{\textcolor{RoyalBlue}{\num{78.473}}\\\textcolor{CrimsonRed}{\num{79.923}}} \\
{\scriptsize PROD\_OF\_SPEC} & \shortstack{\textcolor{RoyalBlue}{\num{67.284}}\\\textcolor{CrimsonRed}{\num{72.701}}} & \shortstack{\textcolor{RoyalBlue}{\num{177.627}}\\\textcolor{CrimsonRed}{\num{232.516}}} & \shortstack{\textcolor{RoyalBlue}{\num{5.420}}\\\textcolor{CrimsonRed}{\num{6.298}}} & \shortstack{\textcolor{RoyalBlue}{\num{55.458}}\\\textcolor{CrimsonRed}{\num{64.973}}} & \shortstack{\textcolor{RoyalBlue}{\num{196.634}}\\\textcolor{CrimsonRed}{\num{210.576}}} & \shortstack{\textcolor{RoyalBlue}{\num{8.272}}\\\textcolor{CrimsonRed}{\num{8.678}}} & \shortstack{\textcolor{RoyalBlue}{\num{79.109}}\\\textcolor{CrimsonRed}{\num{80.429}}} \\
{\scriptsize PROD\_OF\_SPEC\_OVER\_MARGIN} & \shortstack{\textcolor{RoyalBlue}{\num{67.196}}\\\textcolor{CrimsonRed}{\num{72.740}}} & \shortstack{\textcolor{RoyalBlue}{\num{177.751}}\\\textcolor{CrimsonRed}{\num{232.603}}} & \shortstack{\textcolor{RoyalBlue}{\num{5.401}}\\\textcolor{CrimsonRed}{\num{6.235}}} & \shortstack{\textcolor{RoyalBlue}{\num{55.227}}\\\textcolor{CrimsonRed}{\num{64.942}}} & \shortstack{\textcolor{RoyalBlue}{\num{196.343}}\\\textcolor{CrimsonRed}{\num{210.513}}} & \shortstack{\textcolor{RoyalBlue}{\num{8.133}}\\\textcolor{CrimsonRed}{\num{8.548}}} & \shortstack{\textcolor{RoyalBlue}{\num{79.165}}\\\textcolor{CrimsonRed}{\num{80.539}}} \\
{\scriptsize SPEC\_INIT\_MAIN} & \shortstack{\textcolor{RoyalBlue}{\num{67.235}}\\\textcolor{CrimsonRed}{\num{72.897}}} & \shortstack{\textcolor{RoyalBlue}{\num{177.757}}\\\textcolor{CrimsonRed}{\num{232.667}}} & \shortstack{\textcolor{RoyalBlue}{\num{5.432}}\\\textcolor{CrimsonRed}{\num{6.237}}} & \shortstack{\textcolor{RoyalBlue}{\num{55.198}}\\\textcolor{CrimsonRed}{\num{64.927}}} & \shortstack{\textcolor{RoyalBlue}{\num{197.145}}\\\textcolor{CrimsonRed}{\num{210.896}}} & \shortstack{\textcolor{RoyalBlue}{\num{8.288}}\\\textcolor{CrimsonRed}{\num{8.751}}} & \shortstack{\textcolor{RoyalBlue}{\num{79.271}}\\\textcolor{CrimsonRed}{\num{80.867}}} \\
\bottomrule
\end{tabular}
\renewcommand{\arraystretch}{1.0}
\vspace{1mm}
{\raggedright\footnotesize\textit{\textcolor{RoyalBlue}{Top numbers}: CMS; \textcolor{CrimsonRed}{bottom numbers}: eCMS.}\par}
\end{table}

\begin{table}[t]
\centering
\small
\caption{CMS and eCMS summary $\delta=0.05$}
\label{tab:cms-delta005}
\renewcommand{\arraystretch}{1.6}
\begin{tabular}{lccccccc}
\toprule
\textbf{Measure} & $\mathrm{CMS}^{\mathrm{med}}_{\delta} / \mathrm{eCMS}^{\mathrm{med}}_{\delta}$ & C10/DN121 & C10/NiN & C10/RN50 & FMN/DN121 & FMN/NiN & FMN/RN50 \\
\midrule
{\scriptsize PARAMS} & \shortstack{\textcolor{RoyalBlue}{\num{0.000}}\\\textcolor{CrimsonRed}{\num{0.000}}} & \shortstack{\textcolor{RoyalBlue}{\num{0.000}}\\\textcolor{CrimsonRed}{\num{0.000}}} & \shortstack{\textcolor{RoyalBlue}{\num{0.000}}\\\textcolor{CrimsonRed}{\num{0.000}}} & \shortstack{\textcolor{RoyalBlue}{\num{0.000}}\\\textcolor{CrimsonRed}{\num{0.000}}} & \shortstack{\textcolor{RoyalBlue}{\num{0.000}}\\\textcolor{CrimsonRed}{\num{0.000}}} & \shortstack{\textcolor{RoyalBlue}{\num{0.000}}\\\textcolor{CrimsonRed}{\num{0.000}}} & \shortstack{\textcolor{RoyalBlue}{\num{0.000}}\\\textcolor{CrimsonRed}{\num{0.000}}} \\
{\scriptsize PACBAYES\_MAG\_ORIG} & \shortstack{\textcolor{RoyalBlue}{\num{0.096}}\\\textcolor{CrimsonRed}{\num{0.096}}} & \shortstack{\textcolor{RoyalBlue}{\num{0.117}}\\\textcolor{CrimsonRed}{\num{0.131}}} & \shortstack{\textcolor{RoyalBlue}{\num{0.056}}\\\textcolor{CrimsonRed}{\num{0.054}}} & \shortstack{\textcolor{RoyalBlue}{\num{0.075}}\\\textcolor{CrimsonRed}{\num{0.076}}} & \shortstack{\textcolor{RoyalBlue}{\num{0.110}}\\\textcolor{CrimsonRed}{\num{0.113}}} & \shortstack{\textcolor{RoyalBlue}{\num{0.082}}\\\textcolor{CrimsonRed}{\num{0.079}}} & \shortstack{\textcolor{RoyalBlue}{\num{0.120}}\\\textcolor{CrimsonRed}{\num{0.119}}} \\
{\scriptsize PACBAYES\_MAG\_INIT} & \shortstack{\textcolor{RoyalBlue}{\num{0.136}}\\\textcolor{CrimsonRed}{\num{0.144}}} & \shortstack{\textcolor{RoyalBlue}{\num{0.133}}\\\textcolor{CrimsonRed}{\num{0.143}}} & \shortstack{\textcolor{RoyalBlue}{\num{0.067}}\\\textcolor{CrimsonRed}{\num{0.064}}} & \shortstack{\textcolor{RoyalBlue}{\num{0.139}}\\\textcolor{CrimsonRed}{\num{0.145}}} & \shortstack{\textcolor{RoyalBlue}{\num{0.286}}\\\textcolor{CrimsonRed}{\num{0.289}}} & \shortstack{\textcolor{RoyalBlue}{\num{0.073}}\\\textcolor{CrimsonRed}{\num{0.068}}} & \shortstack{\textcolor{RoyalBlue}{\num{0.302}}\\\textcolor{CrimsonRed}{\num{0.290}}} \\
{\scriptsize FRO\_OVER\_SPEC} & \shortstack{\textcolor{RoyalBlue}{\num{0.188}}\\\textcolor{CrimsonRed}{\num{0.198}}} & \shortstack{\textcolor{RoyalBlue}{\num{0.138}}\\\textcolor{CrimsonRed}{\num{0.162}}} & \shortstack{\textcolor{RoyalBlue}{\num{0.143}}\\\textcolor{CrimsonRed}{\num{0.143}}} & \shortstack{\textcolor{RoyalBlue}{\num{0.095}}\\\textcolor{CrimsonRed}{\num{0.111}}} & \shortstack{\textcolor{RoyalBlue}{\num{0.369}}\\\textcolor{CrimsonRed}{\num{0.389}}} & \shortstack{\textcolor{RoyalBlue}{\num{0.233}}\\\textcolor{CrimsonRed}{\num{0.234}}} & \shortstack{\textcolor{RoyalBlue}{\num{0.561}}\\\textcolor{CrimsonRed}{\num{0.569}}} \\
{\scriptsize INVERSE\_MARGIN} & \shortstack{\textcolor{RoyalBlue}{\num{0.246}}\\\textcolor{CrimsonRed}{\num{0.210}}} & \shortstack{\textcolor{RoyalBlue}{\num{0.228}}\\\textcolor{CrimsonRed}{\num{0.191}}} & \shortstack{\textcolor{RoyalBlue}{\num{0.167}}\\\textcolor{CrimsonRed}{\num{0.155}}} & \shortstack{\textcolor{RoyalBlue}{\num{0.265}}\\\textcolor{CrimsonRed}{\num{0.229}}} & \shortstack{\textcolor{RoyalBlue}{\num{0.323}}\\\textcolor{CrimsonRed}{\num{0.297}}} & \shortstack{\textcolor{RoyalBlue}{\num{0.145}}\\\textcolor{CrimsonRed}{\num{0.127}}} & \shortstack{\textcolor{RoyalBlue}{\num{0.364}}\\\textcolor{CrimsonRed}{\num{0.326}}} \\
{\scriptsize PACBAYES\_MAG\_FLATNESS} & \shortstack{\textcolor{RoyalBlue}{\num{0.299}}\\\textcolor{CrimsonRed}{\num{0.311}}} & \shortstack{\textcolor{RoyalBlue}{\num{0.395}}\\\textcolor{CrimsonRed}{\num{0.442}}} & \shortstack{\textcolor{RoyalBlue}{\num{0.221}}\\\textcolor{CrimsonRed}{\num{0.204}}} & \shortstack{\textcolor{RoyalBlue}{\num{0.310}}\\\textcolor{CrimsonRed}{\num{0.347}}} & \shortstack{\textcolor{RoyalBlue}{\num{0.288}}\\\textcolor{CrimsonRed}{\num{0.275}}} & \shortstack{\textcolor{RoyalBlue}{\num{0.170}}\\\textcolor{CrimsonRed}{\num{0.141}}} & \shortstack{\textcolor{RoyalBlue}{\num{0.411}}\\\textcolor{CrimsonRed}{\num{0.381}}} \\
{\scriptsize PACBAYES\_FLATNESS} & \shortstack{\textcolor{RoyalBlue}{\num{0.627}}\\\textcolor{CrimsonRed}{\num{0.628}}} & \shortstack{\textcolor{RoyalBlue}{\num{0.448}}\\\textcolor{CrimsonRed}{\num{0.478}}} & \shortstack{\textcolor{RoyalBlue}{\num{0.539}}\\\textcolor{CrimsonRed}{\num{0.540}}} & \shortstack{\textcolor{RoyalBlue}{\num{0.392}}\\\textcolor{CrimsonRed}{\num{0.426}}} & \shortstack{\textcolor{RoyalBlue}{\num{0.715}}\\\textcolor{CrimsonRed}{\num{0.717}}} & \shortstack{\textcolor{RoyalBlue}{\num{0.754}}\\\textcolor{CrimsonRed}{\num{0.738}}} & \shortstack{\textcolor{RoyalBlue}{\num{0.796}}\\\textcolor{CrimsonRed}{\num{0.782}}} \\
{\scriptsize PACBAYES\_ORIG} & \shortstack{\textcolor{RoyalBlue}{\num{0.558}}\\\textcolor{CrimsonRed}{\num{0.665}}} & \shortstack{\textcolor{RoyalBlue}{\num{0.550}}\\\textcolor{CrimsonRed}{\num{0.655}}} & \shortstack{\textcolor{RoyalBlue}{\num{0.972}}\\\textcolor{CrimsonRed}{\num{0.989}}} & \shortstack{\textcolor{RoyalBlue}{\num{0.566}}\\\textcolor{CrimsonRed}{\num{0.676}}} & \shortstack{\textcolor{RoyalBlue}{\num{0.473}}\\\textcolor{CrimsonRed}{\num{0.461}}} & \shortstack{\textcolor{RoyalBlue}{\num{0.751}}\\\textcolor{CrimsonRed}{\num{0.744}}} & \shortstack{\textcolor{RoyalBlue}{\num{0.505}}\\\textcolor{CrimsonRed}{\num{0.470}}} \\
{\scriptsize SUM\_OF\_FRO\_OVER\_MARGIN} & \shortstack{\textcolor{RoyalBlue}{\num{0.781}}\\\textcolor{CrimsonRed}{\num{0.862}}} & \shortstack{\textcolor{RoyalBlue}{\num{0.622}}\\\textcolor{CrimsonRed}{\num{0.757}}} & \shortstack{\textcolor{RoyalBlue}{\num{1.033}}\\\textcolor{CrimsonRed}{\num{1.132}}} & \shortstack{\textcolor{RoyalBlue}{\num{0.763}}\\\textcolor{CrimsonRed}{\num{0.917}}} & \shortstack{\textcolor{RoyalBlue}{\num{0.498}}\\\textcolor{CrimsonRed}{\num{0.545}}} & \shortstack{\textcolor{RoyalBlue}{\num{0.911}}\\\textcolor{CrimsonRed}{\num{1.070}}} & \shortstack{\textcolor{RoyalBlue}{\num{0.799}}\\\textcolor{CrimsonRed}{\num{0.807}}} \\
{\scriptsize SUM\_OF\_FRO} & \shortstack{\textcolor{RoyalBlue}{\num{0.785}}\\\textcolor{CrimsonRed}{\num{0.867}}} & \shortstack{\textcolor{RoyalBlue}{\num{0.623}}\\\textcolor{CrimsonRed}{\num{0.757}}} & \shortstack{\textcolor{RoyalBlue}{\num{1.045}}\\\textcolor{CrimsonRed}{\num{1.141}}} & \shortstack{\textcolor{RoyalBlue}{\num{0.765}}\\\textcolor{CrimsonRed}{\num{0.918}}} & \shortstack{\textcolor{RoyalBlue}{\num{0.501}}\\\textcolor{CrimsonRed}{\num{0.547}}} & \shortstack{\textcolor{RoyalBlue}{\num{0.929}}\\\textcolor{CrimsonRed}{\num{1.090}}} & \shortstack{\textcolor{RoyalBlue}{\num{0.805}}\\\textcolor{CrimsonRed}{\num{0.816}}} \\
{\scriptsize PARAM\_NORM} & \shortstack{\textcolor{RoyalBlue}{\num{0.955}}\\\textcolor{CrimsonRed}{\num{1.097}}} & \shortstack{\textcolor{RoyalBlue}{\num{0.985}}\\\textcolor{CrimsonRed}{\num{1.164}}} & \shortstack{\textcolor{RoyalBlue}{\num{1.359}}\\\textcolor{CrimsonRed}{\num{1.445}}} & \shortstack{\textcolor{RoyalBlue}{\num{0.854}}\\\textcolor{CrimsonRed}{\num{1.030}}} & \shortstack{\textcolor{RoyalBlue}{\num{0.702}}\\\textcolor{CrimsonRed}{\num{0.774}}} & \shortstack{\textcolor{RoyalBlue}{\num{1.281}}\\\textcolor{CrimsonRed}{\num{1.321}}} & \shortstack{\textcolor{RoyalBlue}{\num{0.925}}\\\textcolor{CrimsonRed}{\num{0.975}}} \\
{\scriptsize PATH\_NORM\_OVER\_MARGIN} & \shortstack{\textcolor{RoyalBlue}{\num{1.288}}\\\textcolor{CrimsonRed}{\num{1.218}}} & \shortstack{\textcolor{RoyalBlue}{\num{1.220}}\\\textcolor{CrimsonRed}{\num{1.134}}} & \shortstack{\textcolor{RoyalBlue}{\num{0.445}}\\\textcolor{CrimsonRed}{\num{0.403}}} & \shortstack{\textcolor{RoyalBlue}{\num{1.356}}\\\textcolor{CrimsonRed}{\num{1.303}}} & \shortstack{\textcolor{RoyalBlue}{\num{3.325}}\\\textcolor{CrimsonRed}{\num{3.351}}} & \shortstack{\textcolor{RoyalBlue}{\num{0.593}}\\\textcolor{CrimsonRed}{\num{0.555}}} & \shortstack{\textcolor{RoyalBlue}{\num{4.025}}\\\textcolor{CrimsonRed}{\num{4.024}}} \\
{\scriptsize SUM\_OF\_SPEC\_OVER\_MARGIN} & \shortstack{\textcolor{RoyalBlue}{\num{1.115}}\\\textcolor{CrimsonRed}{\num{1.235}}} & \shortstack{\textcolor{RoyalBlue}{\num{0.744}}\\\textcolor{CrimsonRed}{\num{0.856}}} & \shortstack{\textcolor{RoyalBlue}{\num{1.216}}\\\textcolor{CrimsonRed}{\num{1.252}}} & \shortstack{\textcolor{RoyalBlue}{\num{1.069}}\\\textcolor{CrimsonRed}{\num{1.251}}} & \shortstack{\textcolor{RoyalBlue}{\num{0.819}}\\\textcolor{CrimsonRed}{\num{0.845}}} & \shortstack{\textcolor{RoyalBlue}{\num{1.161}}\\\textcolor{CrimsonRed}{\num{1.219}}} & \shortstack{\textcolor{RoyalBlue}{\num{1.468}}\\\textcolor{CrimsonRed}{\num{1.488}}} \\
{\scriptsize SUM\_OF\_SPEC} & \shortstack{\textcolor{RoyalBlue}{\num{1.124}}\\\textcolor{CrimsonRed}{\num{1.245}}} & \shortstack{\textcolor{RoyalBlue}{\num{0.744}}\\\textcolor{CrimsonRed}{\num{0.857}}} & \shortstack{\textcolor{RoyalBlue}{\num{1.227}}\\\textcolor{CrimsonRed}{\num{1.269}}} & \shortstack{\textcolor{RoyalBlue}{\num{1.068}}\\\textcolor{CrimsonRed}{\num{1.253}}} & \shortstack{\textcolor{RoyalBlue}{\num{0.819}}\\\textcolor{CrimsonRed}{\num{0.848}}} & \shortstack{\textcolor{RoyalBlue}{\num{1.180}}\\\textcolor{CrimsonRed}{\num{1.238}}} & \shortstack{\textcolor{RoyalBlue}{\num{1.473}}\\\textcolor{CrimsonRed}{\num{1.489}}} \\
{\scriptsize PACBAYES\_INIT} & \shortstack{\textcolor{RoyalBlue}{\num{1.241}}\\\textcolor{CrimsonRed}{\num{1.254}}} & \shortstack{\textcolor{RoyalBlue}{\num{0.973}}\\\textcolor{CrimsonRed}{\num{1.012}}} & \shortstack{\textcolor{RoyalBlue}{\num{1.227}}\\\textcolor{CrimsonRed}{\num{1.257}}} & \shortstack{\textcolor{RoyalBlue}{\num{0.870}}\\\textcolor{CrimsonRed}{\num{0.917}}} & \shortstack{\textcolor{RoyalBlue}{\num{1.777}}\\\textcolor{CrimsonRed}{\num{1.803}}} & \shortstack{\textcolor{RoyalBlue}{\num{1.256}}\\\textcolor{CrimsonRed}{\num{1.251}}} & \shortstack{\textcolor{RoyalBlue}{\num{1.864}}\\\textcolor{CrimsonRed}{\num{1.849}}} \\
{\scriptsize PATH\_NORM} & \shortstack{\textcolor{RoyalBlue}{\num{1.367}}\\\textcolor{CrimsonRed}{\num{1.296}}} & \shortstack{\textcolor{RoyalBlue}{\num{1.322}}\\\textcolor{CrimsonRed}{\num{1.242}}} & \shortstack{\textcolor{RoyalBlue}{\num{0.556}}\\\textcolor{CrimsonRed}{\num{0.520}}} & \shortstack{\textcolor{RoyalBlue}{\num{1.413}}\\\textcolor{CrimsonRed}{\num{1.349}}} & \shortstack{\textcolor{RoyalBlue}{\num{3.133}}\\\textcolor{CrimsonRed}{\num{3.149}}} & \shortstack{\textcolor{RoyalBlue}{\num{0.632}}\\\textcolor{CrimsonRed}{\num{0.590}}} & \shortstack{\textcolor{RoyalBlue}{\num{3.930}}\\\textcolor{CrimsonRed}{\num{3.921}}} \\
{\scriptsize DIST\_SPEC\_INIT} & \shortstack{\textcolor{RoyalBlue}{\num{1.659}}\\\textcolor{CrimsonRed}{\num{1.708}}} & \shortstack{\textcolor{RoyalBlue}{\num{1.442}}\\\textcolor{CrimsonRed}{\num{1.561}}} & \shortstack{\textcolor{RoyalBlue}{\num{1.578}}\\\textcolor{CrimsonRed}{\num{1.605}}} & \shortstack{\textcolor{RoyalBlue}{\num{1.417}}\\\textcolor{CrimsonRed}{\num{1.549}}} & \shortstack{\textcolor{RoyalBlue}{\num{2.577}}\\\textcolor{CrimsonRed}{\num{2.656}}} & \shortstack{\textcolor{RoyalBlue}{\num{1.740}}\\\textcolor{CrimsonRed}{\num{1.810}}} & \shortstack{\textcolor{RoyalBlue}{\num{2.778}}\\\textcolor{CrimsonRed}{\num{2.777}}} \\
{\scriptsize FRO\_DIST} & \shortstack{\textcolor{RoyalBlue}{\num{1.763}}\\\textcolor{CrimsonRed}{\num{1.807}}} & \shortstack{\textcolor{RoyalBlue}{\num{1.387}}\\\textcolor{CrimsonRed}{\num{1.453}}} & \shortstack{\textcolor{RoyalBlue}{\num{1.670}}\\\textcolor{CrimsonRed}{\num{1.713}}} & \shortstack{\textcolor{RoyalBlue}{\num{1.269}}\\\textcolor{CrimsonRed}{\num{1.386}}} & \shortstack{\textcolor{RoyalBlue}{\num{2.500}}\\\textcolor{CrimsonRed}{\num{2.535}}} & \shortstack{\textcolor{RoyalBlue}{\num{1.856}}\\\textcolor{CrimsonRed}{\num{1.901}}} & \shortstack{\textcolor{RoyalBlue}{\num{2.646}}\\\textcolor{CrimsonRed}{\num{2.653}}} \\
{\scriptsize PROD\_OF\_FRO\_OVER\_MARGIN} & \shortstack{\textcolor{RoyalBlue}{\num{45.293}}\\\textcolor{CrimsonRed}{\num{50.008}}} & \shortstack{\textcolor{RoyalBlue}{\num{150.478}}\\\textcolor{CrimsonRed}{\num{183.171}}} & \shortstack{\textcolor{RoyalBlue}{\num{7.228}}\\\textcolor{CrimsonRed}{\num{7.921}}} & \shortstack{\textcolor{RoyalBlue}{\num{44.252}}\\\textcolor{CrimsonRed}{\num{53.202}}} & \shortstack{\textcolor{RoyalBlue}{\num{120.482}}\\\textcolor{CrimsonRed}{\num{131.962}}} & \shortstack{\textcolor{RoyalBlue}{\num{6.377}}\\\textcolor{CrimsonRed}{\num{7.487}}} & \shortstack{\textcolor{RoyalBlue}{\num{46.334}}\\\textcolor{CrimsonRed}{\num{46.814}}} \\
{\scriptsize PROD\_OF\_FRO} & \shortstack{\textcolor{RoyalBlue}{\num{45.538}}\\\textcolor{CrimsonRed}{\num{50.283}}} & \shortstack{\textcolor{RoyalBlue}{\num{150.676}}\\\textcolor{CrimsonRed}{\num{183.231}}} & \shortstack{\textcolor{RoyalBlue}{\num{7.314}}\\\textcolor{CrimsonRed}{\num{7.989}}} & \shortstack{\textcolor{RoyalBlue}{\num{44.385}}\\\textcolor{CrimsonRed}{\num{53.258}}} & \shortstack{\textcolor{RoyalBlue}{\num{121.211}}\\\textcolor{CrimsonRed}{\num{132.342}}} & \shortstack{\textcolor{RoyalBlue}{\num{6.503}}\\\textcolor{CrimsonRed}{\num{7.627}}} & \shortstack{\textcolor{RoyalBlue}{\num{46.690}}\\\textcolor{CrimsonRed}{\num{47.308}}} \\
{\scriptsize SPEC\_ORIG\_MAIN} & \shortstack{\textcolor{RoyalBlue}{\num{73.068}}\\\textcolor{CrimsonRed}{\num{78.684}}} & \shortstack{\textcolor{RoyalBlue}{\num{179.892}}\\\textcolor{CrimsonRed}{\num{206.998}}} & \shortstack{\textcolor{RoyalBlue}{\num{8.362}}\\\textcolor{CrimsonRed}{\num{8.650}}} & \shortstack{\textcolor{RoyalBlue}{\num{61.769}}\\\textcolor{CrimsonRed}{\num{71.864}}} & \shortstack{\textcolor{RoyalBlue}{\num{197.573}}\\\textcolor{CrimsonRed}{\num{204.030}}} & \shortstack{\textcolor{RoyalBlue}{\num{7.901}}\\\textcolor{CrimsonRed}{\num{8.281}}} & \shortstack{\textcolor{RoyalBlue}{\num{84.367}}\\\textcolor{CrimsonRed}{\num{85.503}}} \\
{\scriptsize PROD\_OF\_SPEC\_OVER\_MARGIN} & \shortstack{\textcolor{RoyalBlue}{\num{73.576}}\\\textcolor{CrimsonRed}{\num{79.426}}} & \shortstack{\textcolor{RoyalBlue}{\num{180.133}}\\\textcolor{CrimsonRed}{\num{207.233}}} & \shortstack{\textcolor{RoyalBlue}{\num{8.512}}\\\textcolor{CrimsonRed}{\num{8.765}}} & \shortstack{\textcolor{RoyalBlue}{\num{62.025}}\\\textcolor{CrimsonRed}{\num{72.539}}} & \shortstack{\textcolor{RoyalBlue}{\num{198.114}}\\\textcolor{CrimsonRed}{\num{204.537}}} & \shortstack{\textcolor{RoyalBlue}{\num{8.128}}\\\textcolor{CrimsonRed}{\num{8.533}}} & \shortstack{\textcolor{RoyalBlue}{\num{85.126}}\\\textcolor{CrimsonRed}{\num{86.313}}} \\
{\scriptsize PROD\_OF\_SPEC} & \shortstack{\textcolor{RoyalBlue}{\num{73.690}}\\\textcolor{CrimsonRed}{\num{79.493}}} & \shortstack{\textcolor{RoyalBlue}{\num{180.019}}\\\textcolor{CrimsonRed}{\num{207.366}}} & \shortstack{\textcolor{RoyalBlue}{\num{8.589}}\\\textcolor{CrimsonRed}{\num{8.882}}} & \shortstack{\textcolor{RoyalBlue}{\num{61.970}}\\\textcolor{CrimsonRed}{\num{72.646}}} & \shortstack{\textcolor{RoyalBlue}{\num{198.217}}\\\textcolor{CrimsonRed}{\num{205.168}}} & \shortstack{\textcolor{RoyalBlue}{\num{8.262}}\\\textcolor{CrimsonRed}{\num{8.664}}} & \shortstack{\textcolor{RoyalBlue}{\num{85.409}}\\\textcolor{CrimsonRed}{\num{86.340}}} \\
{\scriptsize SPEC\_INIT\_MAIN} & \shortstack{\textcolor{RoyalBlue}{\num{73.884}}\\\textcolor{CrimsonRed}{\num{79.766}}} & \shortstack{\textcolor{RoyalBlue}{\num{180.209}}\\\textcolor{CrimsonRed}{\num{207.184}}} & \shortstack{\textcolor{RoyalBlue}{\num{8.527}}\\\textcolor{CrimsonRed}{\num{8.844}}} & \shortstack{\textcolor{RoyalBlue}{\num{61.985}}\\\textcolor{CrimsonRed}{\num{72.541}}} & \shortstack{\textcolor{RoyalBlue}{\num{199.461}}\\\textcolor{CrimsonRed}{\num{206.594}}} & \shortstack{\textcolor{RoyalBlue}{\num{8.284}}\\\textcolor{CrimsonRed}{\num{8.740}}} & \shortstack{\textcolor{RoyalBlue}{\num{85.783}}\\\textcolor{CrimsonRed}{\num{86.990}}} \\
\bottomrule
\end{tabular}
\renewcommand{\arraystretch}{1.0}
\vspace{1mm}
{\raggedright\footnotesize\textit{\textcolor{RoyalBlue}{Top numbers}: CMS; \textcolor{CrimsonRed}{bottom numbers}: eCMS.}\par}
\end{table}

\paragraph{A note on scope.}
Our quantitative fragility scores target a single desideratum: \emph{when test error does
not change, a credible generalization measure should remain stable}. Scoring well on
CMS/eCMS does \emph{not} imply that a measure is “correct” or sufficient. For example,
VC dimension/parameter count is widely regarded as a poor complexity measure for modern
deep networks, yet it is not fragile in our sense (cf.\ \textsc{PARAMS} rows). Likewise,
the ML‑PACBayes bound discussed in the main text (\S\ref{sec:post-mortem_vs_mar_lik})
is not fragile largely because it is agnostic to hyperparameter changes. Stability is
necessary for utility, but it is not, by itself, a full explanation of generalization.

\section{Post-mortem vs.\ ML-PACBayes }
\label{sec:post-mortem_vs_mar_lik}

We contrast post-training (``post-mortem'') generalization measures with a function-space approach based on the marginal-likelihood PAC--Bayes bound of \citet{valleprez2020generalization}. The bound controls the test error of a hypothesis sampled from a Bayesian posterior over \emph{functions} (not parameters); the key capacity term is the marginal likelihood (Bayesian evidence) of the data under the architecture’s Gaussian‑process (GP) limit. \textbf{All figure numbers below refer to \citet{valleprez2020generalization}.}

\begin{definition}[Marginal-likelihood PAC--Bayes bound]
  Consider binary classification with data distribution $\mathcal{D}$ over $\mathcal{X}\times\{0,1\}$ and a hypothesis space $\mathcal{H}$ of functions $h:\mathcal{X}\to\{0,1\}$. Let $S\sim\mathcal{D}^n$ be a training set of size $n\ge 2$, and let $P$ be a prior on $\mathcal{H}$. Define the consistency set
  \[
    C(S)\;:=\;\{\,h\in\mathcal{H}\;:\;\hat{\varepsilon}(h,S)=0\,\},
  \]
  where $\hat{\varepsilon}(h,S)=\tfrac{1}{n}\sum_{(x,y)\in S}\mathbf{1}\{h(x)\neq y\}$ is the empirical 0--1 error. The (realizable) Bayesian posterior supported on $C(S)$ is
  \[
    Q(h)\;=\;\begin{cases}
      \dfrac{P(h)}{P(C(S))}, & h\in C(S),\\[4pt]
      0, & \text{otherwise},
    \end{cases}
  \]
  where $P(C(S))=\sum_{h\in C(S)}P(h)$ is the \emph{marginal likelihood} (Bayesian evidence) of $S$.

  For any confidence levels $\delta,\gamma\in(0,1]$, with probability at least $1-\delta$ over $S\sim\mathcal{D}^n$ and, conditional on $S$, with probability at least $1-\gamma$ over $h\sim Q$, the generalization error $\varepsilon(h)=\Pr_{(x,y)\sim\mathcal{D}}[h(x)\neq y]$ satisfies
  \begin{equation}
    -\ln\!\bigl(1-\varepsilon(h)\bigr)\;<\;
    \frac{\ln\!\tfrac{1}{P(C(S))} + \ln n + \ln\!\tfrac{1}{\delta} + \ln\!\tfrac{1}{\gamma}}{\,n-1\,}.
  \end{equation}
\end{definition}

\begin{remark}[binary vs. multiclass]
    We state the bound for binary classification for clarity; the function‑space marginal‑likelihood perspective extends to multiclass via standard reductions (e.g., one‑vs‑rest) or multiclass likelihoods.
\end{remark}

\paragraph{What our fragility checks reveal—and what the ML-PACBayes bound gets right.}
We evaluated each measure against consequential perturbations: dataset difficulty, data-size scaling, training‑pipeline changes, and temporal behavior (overtraining). Post‑mortem measures (sharpness/flatness, norm/margin proxies, compression‑style, and deterministic PAC–Bayes surrogates) often fail at least one stressor; in our runs, even strong performers such as \texttt{PACBAYES\_orig} failed key tests. By contrast, the GP‑based bound tracks the \emph{function‑level} regularities we care about:

\begin{itemize}
  \item \textbf{Data complexity.} The marginal‑likelihood bound \emph{tracks dataset difficulty}: it increases with label corruption and preserves the canonical ordering MNIST $<$ Fashion‑MNIST $<$ CIFAR‑10; see Figure~1 in \citep{valleprez2020generalization}. In our audits, several post‑mortem measures—including \texttt{PACBAYES\_orig}—either flattened under corruption or were not able to catch the ordering, even when the test error did.

  \item \textbf{Learning‑curve scaling.} The marginal‑likelihood bound \emph{tracks learning‑curve scaling}: it mirrors the empirical power‑law in $n$ and clusters exponents primarily by \emph{dataset}; see Figures~2--5 in \citep{valleprez2020generalization}. By contrast, some post‑mortem measures barely moved with $n$; others bent the wrong way or mixed optimizer effects with data effects (cf.\ \citep{nagarajan2019uniform}).

  \item \textbf{Temporal behavior (overtraining invariance).} The marginal‑likelihood bound \emph{tracks temporal invariance}: once $S$ is fixed and training has interpolated ($\hat\varepsilon=0$), the bound depends only on $P(C(S))$ and $n$—it is agnostic to how long or by which path the parameters were trained. In our temporal‑behavior experiments, several norm‑based post‑mortem measures (e.g., $\ell_2$/spectral/path norms) \emph{kept growing} during overtraining while the generalization error stayed roughly unchanged; the GP bound remained stable. (This echoes broader observations that longer training can leave test error flat or improved \citep{hoffer2017train}, and that norms can diverge under separable losses without hurting classification error \citep{soudry2018the}.)

  \item \textbf{Training‑pipeline changes.} Many post‑mortem measures swing with optimizer, batch size, explicit/implicit regularization, and early stopping—sometimes tracking curvature artifacts rather than out‑of‑sample error \citep[e.g.,][]{keskar2016on,hochreiter1997flatb,jiang2019fantastic,dziugaite2020in}. The GP marginal‑likelihood bound is, by construction, insensitive to these knobs: it depends on the architecture’s function prior and the data, not on the path taken through parameter space. We view this \emph{invariance} as a virtue for a predictor; the bound is not designed to explain \emph{differences caused by} optimizer choice and should not be judged on that criterion.
\end{itemize}

These properties are consistent with our optimizer‑swap and pixel‑permutation stress tests (Apps.~\ref{app:temporal-behavior} and \ref{sec:data-complexity}) and with our scale‑symmetry Exp++ dial, which only alters parameter scale (App.~\ref{app:exp-lr}).

\paragraph{Why does the function‑space route outperform post‑mortems?}
Two ingredients stand out. First, the bound evaluates an Occam factor in \emph{function space}: architectures that place high prior mass on data‑compatible functions earn better evidence, naturally capturing dataset ordering and learning‑curve slopes. Second, there is a credible external yardstick: infinite‑width GP predictors (NNGP/NTK) often \emph{quantitatively} predict finite‑width DNN generalization trends \citep{lee2020finite}. Together, these explain why Figures~1–5 show greater stability for the GP‑based approach than we observe for post‑hoc, parameter‑space measures.

\paragraph{So why do post‑mortems underperform?}
The live hypothesis is that they \emph{measure the wrong object}. Post‑mortem scores probe properties of a single trained parameter vector (curvature, norms, compressibility), entangling optimization details and reparameterization choices with generalization. That enterprise is important—and the community has argued forcefully that such post‑training diagnostics deserve attention \citep[e.g., compression and robustness perspectives in][]{arora2018strongerc,arora2019fine,jiang2019fantastic,dziugaite2020in}—but our evidence suggests their predictions are fragile under routine perturbations. By contrast, the GP bound targets the distribution over \emph{functions} implied by architecture and data.

\paragraph{Open question.}
Even if the GP‑based predictor/bound wins these stress tests, finite‑width networks can outperform their GP limits; yet the GP still predicts a striking fraction of performance trends (Figures~2–5). Closing this gap—by designing reparameterization‑invariant, data‑aware post‑mortem diagnostics that inherit the Occam flavour of marginal likelihood—remains open.

\section{Stress-testing generalization measures with pixel permutations}
\label{sec:data-complexity}

A useful generalization measure should pass two basic stress tests: (i) it should be
\emph{insensitive} to symmetry-preserving transformations that do not change the
intrinsic task, yet (ii) \emph{sensitive} when task-relevant information is destroyed.
Pixel permutations provide such a testbed: applying the \emph{same} permutation to the
training and test sets preserves label--input relationships (a symmetry for fully
connected networks), whereas applying \emph{independent} random permutations to train
and test destroys spatial structure and any usable signal.

We evaluate three families of measures alongside test error on MNIST with two
architectures: a fully connected network (FCN) and a ResNet-50. We vary the optimizer
(\textsc{sgd} vs.\ \textsc{adam}) and the early-stopping criterion (best cross-entropy
``CE'' vs.\ best accuracy ``Acc'') to expose hyperparameter sensitivity. Results are
summarized in Tables~\ref{tab:pixel-shuffle-fcn} and \ref{tab:pixel-shuffle-resnet}.

\paragraph{FCN (permutation symmetry holds).}
When the same pixel permutation is applied to both train and test, the FCN effectively
sees an unchanged task. As expected, test error remains essentially constant
($\approx 0.03$ across optimizers/stopping), and the PAC-Bayesian marginal likelihood
(ML) bound is flat ($0.142$ throughout). This indicates desirable \emph{robustness}
to symmetry-preserving changes. In contrast, under independent random permutations,
test error rises to $\approx 0.49$--$0.50$ (near random guessing), and the ML bound
increases to $0.532$---appropriately reflecting reduced learnability. The path norm
and the original PAC-Bayes variant (PACBAYES-ORIG) trend upward in this harder setting,
but they also exhibit large spread across optimizers/stopping; e.g., the path norm spans
from $0.105$ to $6.632$ under random shuffling, smaller than some unshuffled cases,
revealing \emph{fragility} to seemingly minor training choices.

\paragraph{ResNet-50 (permutation symmetry broken).}
Because convolutional inductive biases depend on spatial locality, even a single shared
permutation distorts the data geometry and degrades performance (test error increases
from $\approx 0.014$--$0.025$ to $\approx 0.041$--$0.060$; the ML bound rises from
$0.101$ to $0.170$). With independent random permutations, test error again moves to
$\approx 0.44$--$0.46$ and the ML bound to $\approx 0.504$. The path norm is
particularly volatile here, spanning \emph{orders of magnitude} across training choices
(e.g., $0.016$ vs.\ $3375.355$), underscoring substantial \emph{measure fragility}.

\paragraph{Takeaways.}
Under symmetry-preserving changes (FCN + same-permutation), a robust measure should not
move; under signal destruction (independent permutations), it should reflect the loss
of learnability. The PAC-Bayesian ML bound behaves in this manner in both architectures,
whereas the path norm and PACBAYES-ORIG can vary dramatically with optimizer/stopping,
masking the underlying data effect. Reporting such sensitivity is crucial when proposing
or comparing generalization measures, in line with our paper's emphasis on diagnosing and
documenting \emph{fragility}.%
\footnote{While label corruption is another standard knob for data complexity, here we focus
on pixel permutations to isolate architectural symmetry vs.\ information destruction.
}

\begin{table}[t]
\centering
\small
\begin{threeparttable}
\caption{Pixel permutations with \textbf{MNIST + FCN} (training set size $n{=}10{,}000$). 
Same-permutation preserves the FCN's permutation symmetry; independent (random) permutations destroy learnable signal.}
\label{tab:pixel-shuffle-fcn}
\setlength{\tabcolsep}{6pt}
\begin{tabular}{
    >{\raggedright\arraybackslash}p{3.2cm}
    S[table-format=1.3]
    S[table-format=1.3]
    S[table-format=1.3]
    S[table-format=2.3]
}
\toprule
\textbf{Optimizer / stopping} & \textbf{Test err.} & \textbf{PAC-Bayes ML} & \textbf{Path norm} & \textbf{PACBAYES-ORIG} \\
\midrule
\multicolumn{5}{c}{\emph{Unshuffled pixels}}\\
\midrule
SGD / CE   & 0.033 & 0.142 & 0.080 & 1.534 \\
ADAM / CE  & 0.034 & 0.142 & 0.173 & 1.208 \\
SGD / Acc  & 0.032 & 0.142 & 0.081 & 1.535 \\
ADAM / Acc & 0.031 & 0.142 & 0.379 & 1.187 \\
\midrule
\multicolumn{5}{c}{\emph{Same permutation on train \& test}}\\
\midrule
SGD / CE   & 0.032 & 0.142 & 0.080 & 1.558 \\
ADAM / CE  & 0.034 & 0.142 & 0.184 & 1.202 \\
SGD / Acc  & 0.032 & 0.142 & 0.081 & 1.534 \\
ADAM / Acc & 0.032 & 0.142 & 0.314 & 1.183 \\
\midrule
\multicolumn{5}{c}{\emph{Independent random permutations (train \& test)}}\\
\midrule
SGD / CE   & 0.497 & 0.532 & 0.124 & 2.232 \\
ADAM / CE  & 0.492 & 0.532 & 5.736 & 1.791 \\
SGD / Acc  & 0.498 & 0.532 & 0.105 & 2.400 \\
ADAM / Acc & 0.488 & 0.532 & 6.632 & 1.837 \\
\bottomrule
\end{tabular}
\begin{tablenotes}[flushleft]
\footnotesize
\item \textbf{CE}: best checkpoint by cross-entropy; \textbf{Acc}: best checkpoint by accuracy.
\end{tablenotes}
\end{threeparttable}
\end{table}

\begin{table}[t]
\centering
\small
\begin{threeparttable}
\caption{Pixel permutations with \textbf{MNIST + ResNet-50} ($n{=}10{,}000$). 
Convolutional inductive biases depend on spatial locality, so even a single shared permutation harms performance.}
\label{tab:pixel-shuffle-resnet}
\setlength{\tabcolsep}{6pt}
\begin{tabular}{
    >{\raggedright\arraybackslash}p{3.2cm}
    S[table-format=1.3]
    S[table-format=1.3]
    S[table-format=4.3]
    S[table-format=2.3]
}
\toprule
\textbf{Optimizer / stopping} & \textbf{Test err.} & \textbf{PAC-Bayes ML} & \textbf{Path norm} & \textbf{PACBAYES-ORIG} \\
\midrule
\multicolumn{5}{c}{\emph{Unshuffled pixels}}\\
\midrule
SGD / CE   & 0.022 & 0.101 & 2980.415 & 15.396 \\
ADAM / CE  & 0.016 & 0.101 & 0.173    & 4.498  \\
SGD / Acc  & 0.025 & 0.101 & 2972.887 & 15.125 \\
ADAM / Acc & 0.014 & 0.101 & 0.022    & 4.417  \\
\midrule
\multicolumn{5}{c}{\emph{Same permutation on train \& test}}\\
\midrule
SGD / CE   & 0.060 & 0.170 & 3375.355 & 13.534 \\
ADAM / CE  & 0.046 & 0.170 & 0.016    & 4.074  \\
SGD / Acc  & 0.056 & 0.170 & 3001.528 & 12.529 \\
ADAM / Acc & 0.041 & 0.170 & 0.021    & 3.909  \\
\midrule
\multicolumn{5}{c}{\emph{Independent random permutations (train \& test)}}\\
\midrule
SGD / CE   & 0.460 & 0.504 & 116.602 & 13.77  \\
ADAM / CE  & 0.437 & 0.504 & 0.081   & 4.157  \\
SGD / Acc  & 0.460 & 0.504 & 67.497  & 12.475 \\
ADAM / Acc & 0.437 & 0.504 & 0.084   & 4.119  \\
\bottomrule
\end{tabular}
\begin{tablenotes}[flushleft]
\footnotesize
\item \textbf{CE}: best checkpoint by cross-entropy; \textbf{Acc}: best checkpoint by accuracy.
\end{tablenotes}
\end{threeparttable}
\end{table}

\section{Exp++ in scale-invariant nets: protocol and results}
\label{app:exp-lr}

We instantiate the symmetry in a fully connected, \emph{scale‑invariant} network (normalization after each hidden linear layer; the final linear readout is frozen), trained on \textsc{MNIST} with SGD + momentum. We sweep a single control, the \emph{Exp++ multiplicative factor}~$\alpha$ from Theorem~\ref{theorem:exp+wd}, which makes the learning rate grow exponentially across steps, and repeat the sweep both without and with weight decay. For each run we log parameter‑space PAC–Bayes proxies (including magnitude‑aware variants), path‑norm proxies (with/without margin normalization), and test error. The stopping rule is cross‑entropy unless stated otherwise.

\textbf{Findings.} Across both regimes (no WD and with WD) the test‑error trace barely moves as $\alpha$ grows, yet the magnitude‑sensitive diagnostics explode on a log scale. In particular, the PAC–Bayes family and the path‑norm proxies rise by many orders of magnitude even though the error curve hugs a dashed 10\% reference. This is exactly what the equivalence predicts in a scale‑invariant net: Exp++ changes parameter \emph{scale}, not the predictor—exposing the fragility of magnitude‑dependent measures (notably \texttt{PACBAYES\_orig}) in this setting.

\begin{figure}[t]
  \centering
  \begin{minipage}[t]{0.49\columnwidth}
    \centering
    \includegraphics[width=\linewidth]{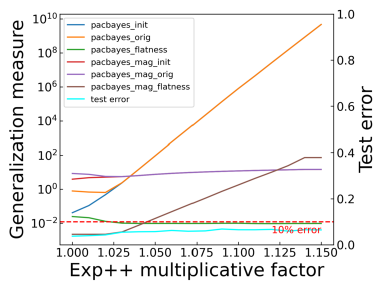}
  \end{minipage}\hfill
  \begin{minipage}[t]{0.49\columnwidth}
    \centering
    \includegraphics[width=\linewidth]{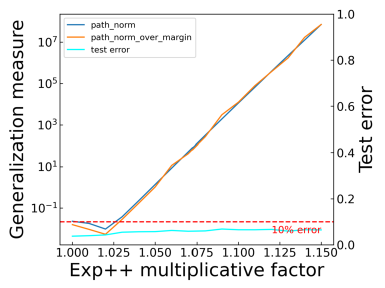}
  \end{minipage}
  \caption{\textbf{Exp++ in a scale‑invariant FCN (no WD; CE stopping).}
  As the Exp++ factor increases, PAC–Bayes proxies (left) and path‑norm proxies (right) swell by orders of magnitude, while the test‑error curve (right axis; dashed 10\% line) remains essentially flat.}
  \label{fig:exp-lr-nowd}
\end{figure}

\begin{figure}[t]
  \centering
  \begin{minipage}[t]{0.49\columnwidth}
    \centering
    \includegraphics[width=\linewidth]{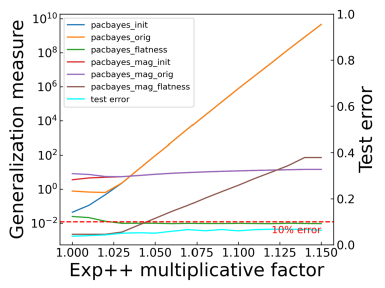}
  \end{minipage}\hfill
  \begin{minipage}[t]{0.49\columnwidth}
    \centering
    \includegraphics[width=\linewidth]{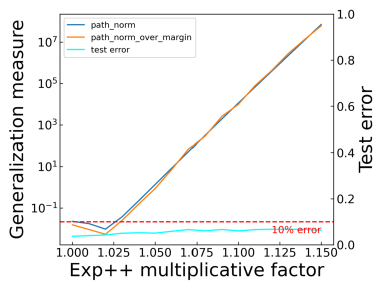}
  \end{minipage}
  \caption{\textbf{Exp++ with weight decay (CE stopping).}
  The qualitative picture persists with WD: magnitude‑sensitive PAC–Bayes and path‑norm diagnostics climb sharply with the Exp++ factor, but the predictor’s test error barely changes.}
  \label{fig:exp-lr-wd}
\end{figure}

\noindent\textbf{Reading the panels.} In both Figure~\ref{fig:exp-lr-nowd} (no WD) and Figure~\ref{fig:exp-lr-wd} (with WD), left panels aggregate PAC–Bayes proxies; right panels show path‑norm proxies. Axes are logarithmic for the measures and linear for the test error (right axis); the dashed horizontal line marks 10\% test error for visual reference. Together, the panels illustrate a clean separation between \emph{parameter scale} (which Exp++ manipulates) and \emph{predictive behavior} (which stays essentially fixed).

\section{Additional Compression Bounds Results}
\label{app:additional_compression}

In Section~\ref{sec:compression_fragility}, we analyzed the breakdown of compression bounds for CIFAR-10 (Intrinsic Dimension 3500) and MNIST. Here, we provide the corresponding omitted results for CIFAR-10 with a higher intrinsic dimension ($ID=10000$).

\begin{figure}[h]
\centering
\subfigure[CIFAR-10 (ID 10000) Bounds]{
    \includegraphics[width=0.48\linewidth]{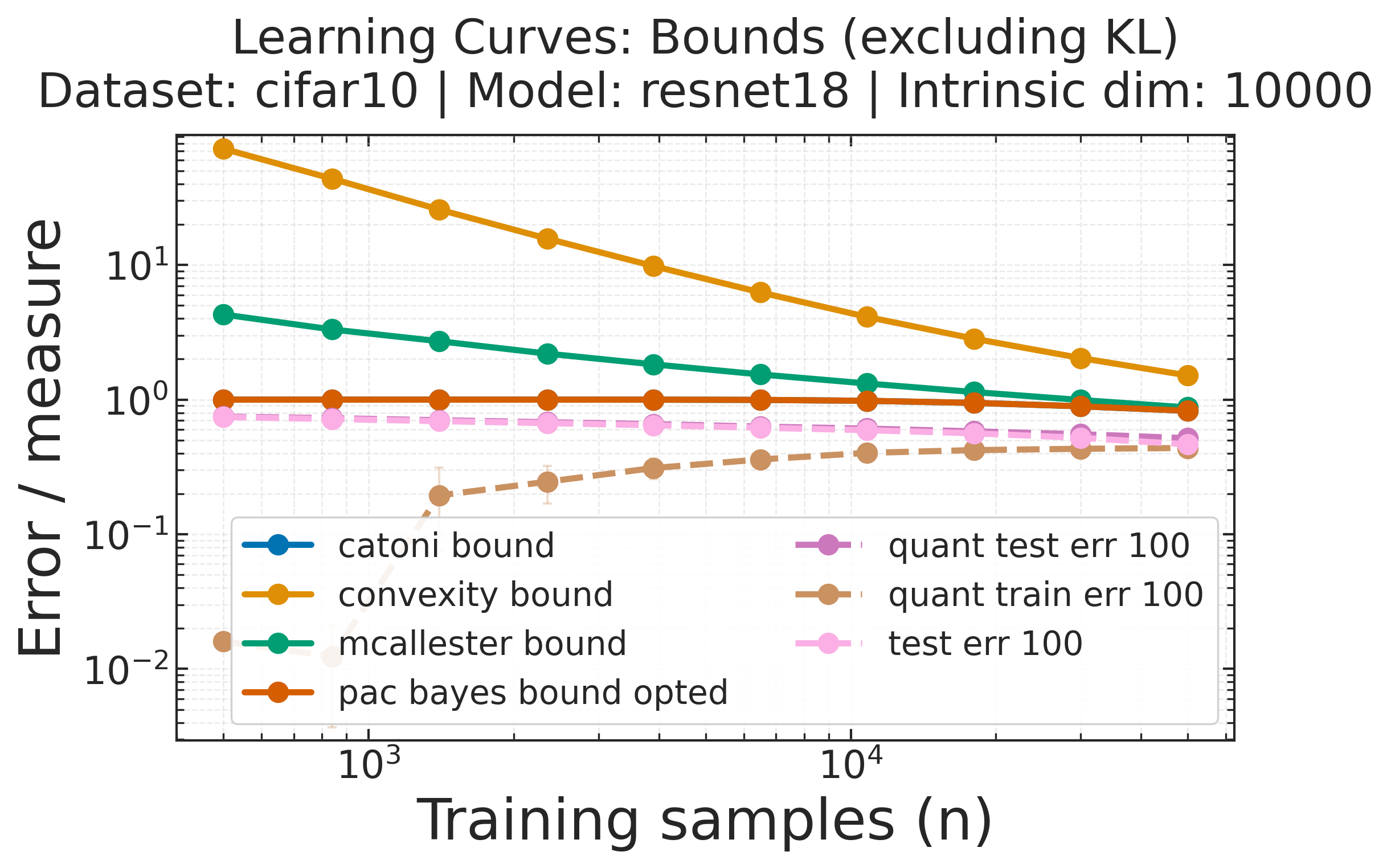}
    \label{fig:comp_c10_10000_met_app}
}
\hfill
\subfigure[CIFAR-10 (ID 10000) Raw KL]{
    \includegraphics[width=0.48\linewidth]{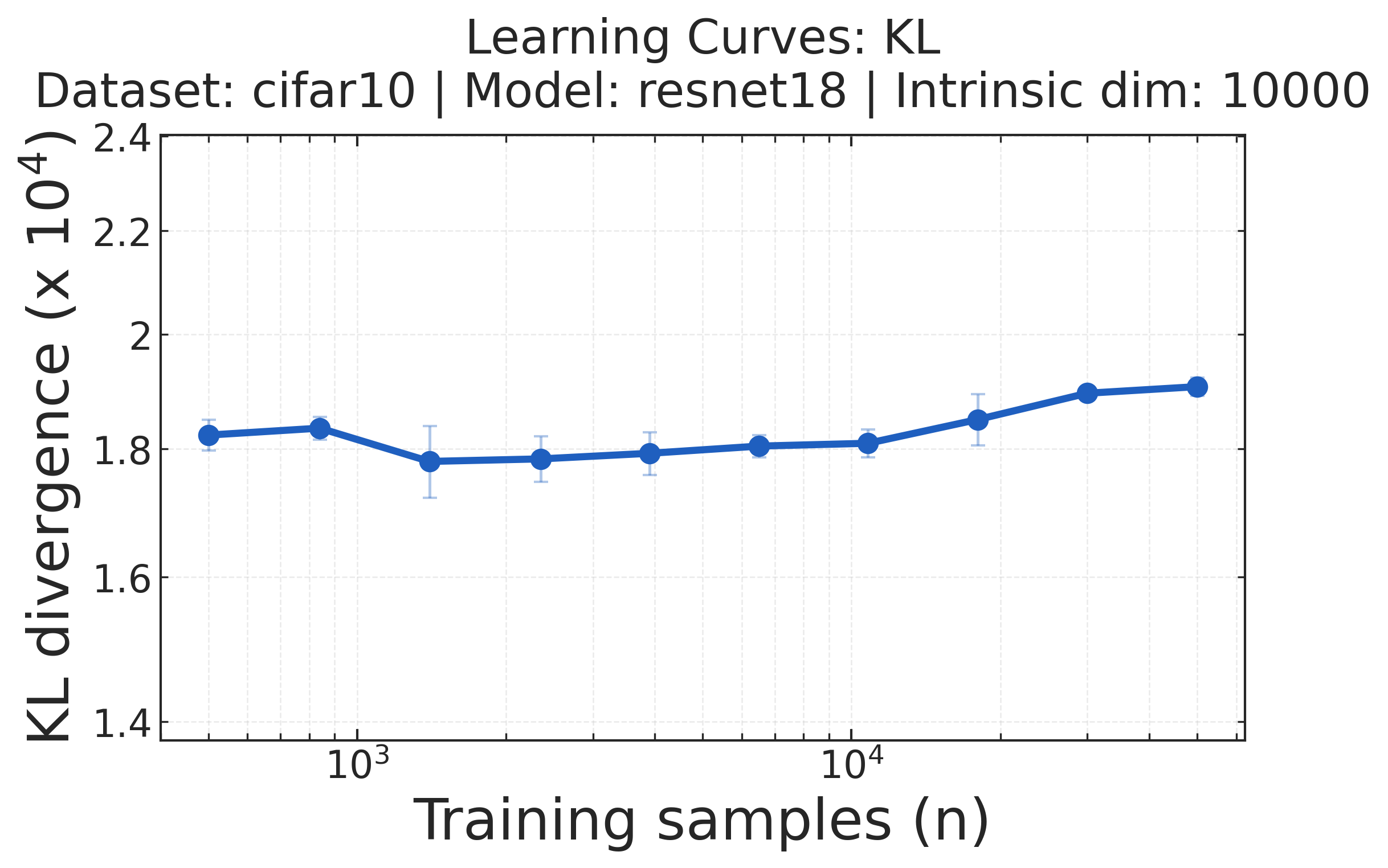}
    \label{fig:comp_c10_10000_kl_app}
}
\caption{\textbf{Compression Bounds with Higher Intrinsic Dimension (CIFAR-10).} These plots correspond to the setting where the intrinsic dimension is set to 10,000. \emph{Left:} The bound components versus sample size. \emph{Right:} The raw KL divergence. Consistent with the main text analysis, the complexity term (KL) remains flat despite the increase in sample size.}
\label{fig:compression_additional_results}
\end{figure}

\end{document}